\documentclass[11pt]{article}

\usepackage[final]{acl}

\usepackage{times}
\usepackage{latexsym}

\usepackage[T1]{fontenc}

\usepackage[utf8]{inputenc}
\usepackage[main=english,russian,ukrainian]{babel}

\usepackage{microtype}

\usepackage{inconsolata}

\usepackage{booktabs}
\usepackage{multirow}
\usepackage{hhline}
\usepackage{makecell}
\usepackage{graphicx}
\usepackage{tabularx}
\usepackage{array}
\usepackage{hyperref}
\usepackage{natbib}
\usepackage{booktabs}
\usepackage{color, colortbl}
\usepackage{enumitem}
\usepackage{caption}
\usepackage{adjustbox}
\usepackage{amssymb}
\usepackage{utfsym}
\usepackage{mathtools}
\usepackage{amsmath}
\usepackage{inconsolata}
\usepackage{subcaption}
\usepackage{enumitem}
\usepackage{pifont}
\usepackage{xcolor}
\usepackage[most]{tcolorbox}
\tcbuselibrary{breakable, skins}
\usepackage{textcomp}
\usepackage{xspace}
\usepackage{rotating}
\usepackage{longtable}
\usepackage{fontawesome5}

\usepackage{epigraph}

\newcommand{\pmath}{\textsc{PluraMath}\xspace}

\newcommand{\cmark}{\textcolor{green!60!black}{\ding{51}}}  
\definecolor{partialyellow}{HTML}{E0A800}   

\usepackage{tikz}
\newcommand{\partialmark}{%
  \begin{tikzpicture}[baseline=-0.5ex]
    \draw[partialyellow, thick] (0,0) circle (0.7ex);
    \fill[partialyellow] (0,0) -- (0,0.7ex) arc (90:-90:0.7ex) -- cycle;
  \end{tikzpicture}%
}

\definecolor{highres}{RGB}{134, 197, 134}   
\definecolor{midres} {RGB}{188, 228, 240}   
\definecolor{lowres} {RGB}{255, 221, 140}   
\definecolor{xlowres}{RGB}{255, 185, 175}   

\definecolor{slavic}   {RGB}{174, 198, 232}   
\definecolor{indoaryan}{RGB}{206, 183, 232}   
\definecolor{hellenic} {RGB}{160, 225, 215}   
\definecolor{romance}  {RGB}{250, 195, 210}   
\definecolor{turkic}   {RGB}{255, 210, 155}   
\definecolor{semitic}  {RGB}{250, 235, 145}   

\definecolor{classone}{HTML}{FBE3E3}   
\definecolor{classtwo}{HTML}{FCEFD9}   
\definecolor{classthree}{HTML}{FCF8D9} 
\definecolor{classfour}{HTML}{E3F0DC}  

\newcommand{\ClsOne}{\cellcolor{classone}1}
\newcommand{\ClsTwo}{\cellcolor{classtwo}2}
\newcommand{\ClsThree}{\cellcolor{classthree}3}
\newcommand{\ClsFour}{\cellcolor{classfour}4}

\newcommand{\Slavic}[1]   {\cellcolor{slavic}#1}
\newcommand{\IndoAryan}[1]{\cellcolor{indoaryan}#1}
\newcommand{\Hellenic}[1] {\cellcolor{hellenic}#1}
\newcommand{\Romance}[1]  {\cellcolor{romance}#1}
\newcommand{\Turkic}[1]   {\cellcolor{turkic}#1}
\newcommand{\Semitic}[1]  {\cellcolor{semitic}#1}

\newcommand{\High}[1]  {\cellcolor{highres}#1}
\newcommand{\Mid}[1]   {\cellcolor{midres}#1}
\newcommand{\Low}[1]   {\cellcolor{lowres}#1}
\newcommand{\XLow}[1]  {\cellcolor{xlowres}#1}

\newtcolorbox{promptbox}[1][]{%
  enhanced, breakable,
  colback=gray!4, colframe=gray!55!black,
  boxrule=0.4pt, arc=2pt,
  left=6pt, right=6pt, top=4pt, bottom=4pt,
  fontupper=\small\ttfamily,
  #1
}
\newtcolorbox{promptboxtitled}[2][]{%
  enhanced, breakable,
  colback=gray!4, colframe=gray!55!black,
  boxrule=0.4pt, arc=2pt,
  left=6pt, right=6pt, top=4pt, bottom=4pt,
  fontupper=\small\ttfamily,
  title=#2, fonttitle=\small\bfseries\sffamily,
  coltitle=white, colbacktitle=gray!55!black,
  attach boxed title to top left={xshift=4pt, yshift=-2pt},
  boxed title style={arc=2pt, boxrule=0pt},
  #1
}

\newcommand{\hflogo}{\raisebox{-0.15ex}{\includegraphics[height=0.95em]{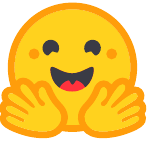}}}
\newcommand{\ghlogo}{\faGithub}
\newcommand{\weblogo}{\faGlobe}

\title{\pmath: Extending Mathematical Reasoning \\ Evaluation Beyond High-Resource Languages}

\author{
  Daryna Dementieva\textsuperscript{1,2}\thanks{\;Correspondence: \href{mailto:daryna.dementieva@tum.de}{\texttt{daryna.dementieva@tum.de}}} \quad
  Nikolay Babakov\textsuperscript{4} \quad
  Kathy Hämmerl\textsuperscript{1,2} \quad
  Ilseyar Alimova\textsuperscript{5} \\
  \textbf{Jindřich Libovický}\textsuperscript{6} \quad
  \textbf{Shu Okabe}\textsuperscript{1,2} \quad
  \textbf{Miras Baisbay}\textsuperscript{7} \quad
  \textbf{Lukas Edman}\textsuperscript{1,2} \quad
  \textbf{Abrorkhon Inomkhujaev}\textsuperscript{1} \\
  \textbf{Antonia Karamolegkou}\textsuperscript{8} \quad
  \textbf{Mateusz Lango}\textsuperscript{6} \quad
  \textbf{Volkan Özer}\textsuperscript{1,2} \quad
  \textbf{Nikola Selic}\textsuperscript{1} \quad
  \textbf{Subhankar Swain}\textsuperscript{9} \\
  \textbf{Tsedeniya Kinfe Temesgen}\textsuperscript{1,2} \quad
  \textbf{Galit Bary Weisberg}\textsuperscript{10} \quad
  \textbf{Alexander Fraser}\textsuperscript{1,2,3} \\[0.4em]
  \footnotesize{\textsuperscript{1}Technical University of Munich (TUM) \quad
  \textsuperscript{2}Munich Center for Machine Learning (MCML)} \quad \textsuperscript{3}Munich Data Science Institute (MDSI) \\
  \footnotesize{
  \textsuperscript{4}Independent Researcher \quad
  \textsuperscript{5}Applied AI Institute \quad
  \textsuperscript{6}Charles University \quad
  \textsuperscript{7}Nazarbayev University} \\
  \footnotesize{\textsuperscript{8}Inria \quad
  \textsuperscript{9}Indian Institute of Technology, Kharagpur (IIT Kharagpur) \quad
  \textsuperscript{10}German University of Digital Science}
}

\begin{document}
\maketitle
\begin{abstract}
Mathematical reasoning has become a central task for evaluating and tuning reasoning Large Language Models (LLMs), yet existing benchmarks remain heavily biased toward high-resource languages, with English and Chinese dominating both pre-training corpora and evaluation suites. The recently released PolyMath~\cite{wang2025polymath} dataset represents a significant step forward, yet its coverage is still limited to 18 \textit{only} high-resource languages. To address this gap, we introduce \pmath, an extension of PolyMath to 18 additional \textit{underrepresented} languages spanning 6 language families---ranging from mid-resource to extreme low-resource settings. We constructed the dataset through a human-curated pipeline, where native speakers thoroughly validated pre-computed translations.
Using \pmath, we then benchmark 27 reasoning LLMs across four model scales---small, mid-size, large, and closed-source ensembles---probing the multilingual mathematical reasoning capabilities of state-of-the-art models under diverse linguistic conditions. Our fine-grained analysis confirms a persistent gap in mathematical reasoning performance between high-resource and underrepresented languages, with stronger results largely associated with better instruction-following ability. We fully open-source our dataset, data acquisition pipeline, and evaluation framework, with the goal of lowering the barrier to multilingual benchmark development for underrepresented communities.

\end{abstract}

\section{Introduction}

\begin{figure}[ht!]
    \centering
    \includegraphics[width=0.6\linewidth]{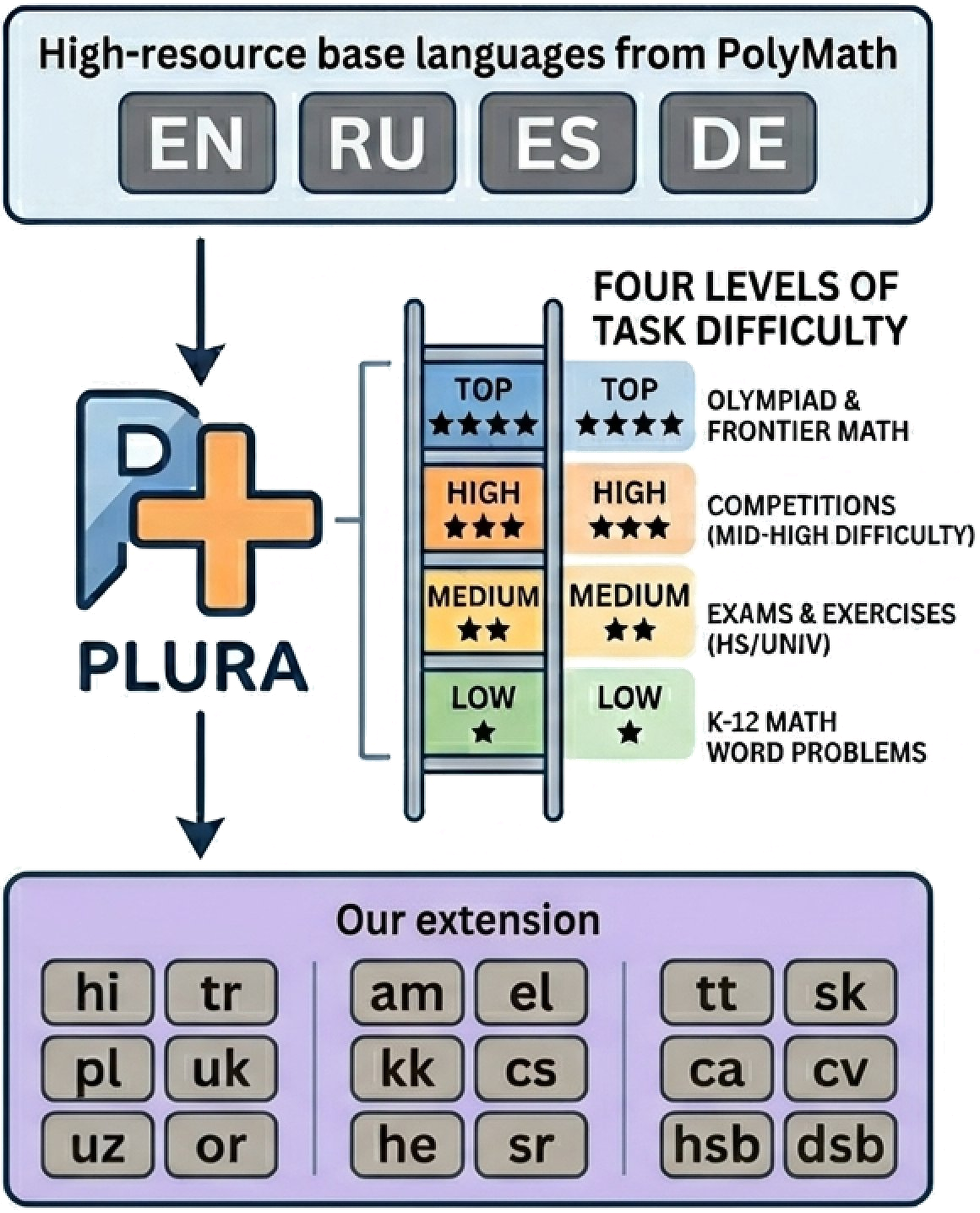}
    \caption{\textit{Plura}---from Latin,  ``more'': our extension of PolyMath~\cite{wang2025polymath} 4-level mathematical reasoning benchmark to 18 underrepresented languages through comprehensive human annotation of pre-computed translations.}
    \label{fig:pluramath}
\end{figure}

Both the pretraining and the evaluation of large language models (LLMs) remain heavily skewed towards English and set of high-resource languages \citep{joshi2020state,ghosh2025multilingualmind}. Mathematical reasoning is no exception: the popular benchmarks are either English-only---GSM8K \citep{cobbe2021gsm8k}, AIME, MathArena \citep{balunovic2025matharena}, OMEGA \citep{sun2025omega}---or Chinese-only \citep{wei2023cmath,zhong2023agieval}. Recent multilingual efforts such as PolyMath \citep{wang2025polymath}, MGSM8KInstruct \citep{chen2024breaking}, and MathNet \citep{alshammari2026mathnet} broaden coverage but still concentrate on the high-resource languages, leaving underrepresented ones untested.

At the same time, a growing amount of recent work shows that mathematical reasoning ability does not transfer easily across languages, even with translation or pivot-language prompting. For instance, models exhibit reasoning-answer misalignment that is more severe in non-Latin scripts as in Latin scripts \citep{ovalle2025beg}; latent reasoning pathways remain English-centred even across typologically diverse languages \citep{liu2026latent}. Thus, truly democratized for multiple languages LLMs development progress therefore requires evaluation benchmarks that stress-test reasoning models in genuinely diverse linguistic settings, including the long tail.

\begin{table*}[th!]
\centering
\scriptsize
\setlength{\tabcolsep}{6pt}
\renewcommand{\arraystretch}{1.15}

\begin{tabular}{
  l
  l
  r
  c
  l
  c
  l
  l
  c
  c
}
\toprule
\multirow{2}{*}{\textbf{Code}}
&
\multirow{2}{*}{\textbf{Language}}
  & \textbf{L1}
  & \textbf{Lang}
  & \multirow{2}{*}{\textbf{Language Family}}
  & \textbf{Math}
  & \textbf{Trans.}
  & \textbf{Source}
  & \textbf{\#}
  & \textbf{\#} \\
  &
  & \textbf{Spk.\ (M)}
  & Class
  &
  & \textbf{Instr.}
  & \textbf{Model}
  & \textbf{Language}
  & \textbf{Ann.}
  & \textbf{Rev.} \\
\midrule

\texttt{hi} & Hindi         & \High{600.0}  & \ClsFour  & \IndoAryan{IE -- Indo-Aryan} & \cmark & Sarvamai & English & 1 & $<$10\\
\texttt{tr} & Turkish       & \Mid{80.0}    & \ClsFour  & \Turkic{Turkic}              & \cmark & Gemini & English & 2 & 179 \\
\texttt{pl} & Polish        & \Mid{45.0}    & \ClsFour  & \Slavic{IE -- Slavic}        & \cmark & DeepL & English & 1 & 427 \\
\texttt{uk} & Ukrainian     & \Mid{40.0}    & \ClsThree & \Slavic{IE -- Slavic}        & \cmark & DeepL & Russian & 2 & 14\\
\texttt{uz} & Uzbek         & \Mid{35.0}    & \ClsThree & \Turkic{Turkic}              & \partialmark & DeepL & English & 1 & 500\\
\texttt{or} & Odia          & \Mid{35.0}    & \ClsOne   & \IndoAryan{IE -- Indo-Aryan} & \cmark & Sarvamai & English & 1 & $<$10\\
\texttt{am} & Amharic       & \Mid{32.0}    & \ClsTwo   & \Semitic{AA -- Semitic}      & \cmark & Gemini & English & 1 & 126\\
\texttt{el} & Greek         & \Mid{13.0}    & \ClsThree & \Hellenic{IE -- Hellenic}    & \cmark & DeepL & English  & 2 & 27 \\
\texttt{kk} & Kazakh        & \Mid{13.0}    & \ClsThree & \Turkic{Turkic}              & \cmark & DeepL & Russian & 2 & 500 \\
\texttt{cs} & Czech         & \Mid{10.0}    & \ClsFour  & \Slavic{IE -- Slavic}        & \cmark & DeepL & German & 1 & 173 \\
\texttt{he} & Hebrew        & \Low{9.0}     & \ClsThree & \Semitic{AA -- Semitic}      & \cmark & DeepL & English  & 2 & 34 \\
\texttt{sr} & Serbian       & \Low{8.2}     & \ClsFour  & \Slavic{IE -- Slavic}        & \cmark & DeepL & English & 1 & 370 \\
\texttt{tt} & Tatar         & \Low{5.5}     & \ClsOne   & \Turkic{Turkic}              & \cmark & DeepL & Russian & 2 & 390 \\
\texttt{sk} & Slovak        & \Low{5.0}     & \ClsThree & \Slavic{IE -- Slavic}        & \cmark & DeepL & English & 1 & 305 \\
\texttt{ca} & Catalan       & \Low{4.0}     & \ClsFour  & \Romance{IE -- Romance}      & \cmark & SalamandraTA & Spanish & 1 & 222 \\
\texttt{cv} & Chuvash       & \Low{1.0}     & \ClsOne   & \Turkic{Turkic}              & \partialmark & Gemini & Russian & 1 & 496 \\
\texttt{hsb} & Upper Sorbian & \XLow{0.013} & \ClsOne   & \Slavic{IE -- Slavic}        & \partialmark & TartuNLP & German & 1 & 379 \\
\texttt{dsb} & Lower Sorbian & \XLow{0.007} & \ClsOne   & \Slavic{IE -- Slavic}        & \partialmark & TartuNLP & German & 1 & 440 \\

\bottomrule
\end{tabular}

\bigskip

\renewcommand{\arraystretch}{1.2}
\setlength{\tabcolsep}{4pt}

\noindent
\begin{tabular}{@{}lll@{\hspace{1.5em}}ll@{\hspace{1.5em}}ll@{}}
  \multicolumn{2}{l}{\textbf{Number of L1 speakers}} & 
  & \multicolumn{2}{l}{\textbf{Language class~\cite{joshi-etal-2020-state}}}
  & \multicolumn{2}{l}{\textbf{Language family}} \\[3pt]

  \cellcolor{highres}\phantom{XX} & $> 100$M & 
  & \cellcolor{classfour}\phantom{XX}  & 4 Underdogs: much unlabeled, less labeled
  & \cellcolor{slavic}\phantom{XX}    & IE -- Slavic \\

  \cellcolor{midres}\phantom{XX}  & $10$--$100$M & 
  & \cellcolor{classthree}\phantom{XX} & 3 Rising Stars: strong web, few labels
  & \cellcolor{indoaryan}\phantom{XX} & IE -- Indo-Aryan \\

  \cellcolor{lowres}\phantom{XX} & $1$--$10$M & 
  & \cellcolor{classtwo}\phantom{XX}   & 2 Hopefuls: some labeled data, support
  & \cellcolor{hellenic}\phantom{XX}  & IE -- Hellenic \\

  \cellcolor{xlowres}\phantom{XX} & $< 1$M & 
  & \cellcolor{classone}\phantom{XX}   & 1 Scraping-Bys: little data, few labels
  & \cellcolor{romance}\phantom{XX}   & IE -- Romance \\

  & & & & & \cellcolor{turkic}\phantom{XX}  & Turkic \\
  & & & & & \cellcolor{semitic}\phantom{XX} & AA -- Semitic \\
\end{tabular}

\caption{Overview of languages in \textsc{PluraMath}. L1 speaker counts (in millions) are approximate, sourced from \href{https://www.ethnologue.com}{Ethnologue}.
\textbf{Lang Class} follows the resource taxonomy of \citet{joshi-etal-2020-state} (1--4).
\cmark~indicates the language is used for mathematics teaching in a region; \protect\partialmark~indicates if not for the university level.
\textbf{IE} = Indo-European; \textbf{AA} = Afro-Asiatic. We cover a diversity of underrepresented languages including quite low-resource ones.}
\label{tab:lang_info}
\end{table*}

We address this gap by extending PolyMath \citep{wang2025polymath} to 18 underrepresented languages from 6 language families. Crucially, rather than relying on machine-translation-only pipelines or with LLM-based post-editing,
we conduct full human rigorous check of pre-computed translations by native speakers for our benchmark.

Our main contributions are as follows:
\begin{itemize}
    \item We release \pmath, a novel multilingual mathematical reasoning benchmark that extends PolyMath to 18 underrepresented languages from 6 language families (Table~\ref{tab:lang_info}).
    \item We open-source the full data acquisition and validation pipeline, including annotators guidelines and quality-control procedures.
    \item We then evaluate 27 modern reasoning models under three prompting setups, spanning small, mid-sized, large open-weight, and closed-source systems, to assess multilingual mathematical reasoning across diverse linguistic conditions and compare performance against high-resource base languages.
    \item We conduct a comprehensive analysis of models' performance, reasoning behavior, and error patterns, identifying correlations with models' families, language categories, and translation-related capabilities, complemented by human evaluation.
\end{itemize}

Our findings confirm a persistent gap in mathematical reasoning performance between high-resource and underrepresented languages. Although reasoning behavior varies across models, including reasoning in either English or the target language, performance gains are only weakly associated with translation capability and are not substantially improved by prompting in high-resource languages. Instead, stronger results primarily correlate with general instruction-following ability, with larger models consistently achieving better performance with shorter reasoning.
The dataset, prompts, analysis code, and annotation instructions are publicly released.\footnote{\weblogo~Website: \href{https://tum-nlp.github.io/pluramath}{tum-nlp.github.io/pluramath}}$^,$\footnote{\hflogo~\pmath: \href{https://huggingface.co/datasets/tum-nlp/PluraMath}{tum-nlp/PluraMath}}$^,$\footnote{\ghlogo~Code and details: \href{https://github.com/TUM-NLP/pluramath}{TUM-NLP/pluramath}}
We hope that our obtained dataset, extension methodology, and error analysis will support further progress on multilingual mathematical reasoning.

\section{Related Work}
\label{sec:related}

\subsection{Mathematical Reasoning Benchmarks}

Early benchmarks for evaluating mathematical reasoning in language models have been overwhelmingly English-centric. GSM8K \citep{cobbe2021gsm8k} established the standard for grade-school word problems, while OpenWebMath \citep{paster2023openwebmath} curated 14.7B tokens of English mathematical web text. Beyond grade school, English-only resources have probed competition-level mathematics through AIME (the American Invitational Mathematics Examination), MathArena \citep{balunovic2025matharena}, and OMEGA \citep{sun2025omega}. Adjacent reasoning evaluations follow the same pattern: HumanEval \citep{chen2021humaneval} for English code generation and MME-Reasoning \citep{yuan2025mmereasoning} for multimodal logical reasoning are similarly English-centred. The principal exceptions have come from Chinese CMATH \citep{wei2023cmath} and AGIEval \citep{zhong2023agieval}.

Multilingual mathematical benchmarks have only recently begun to emerge, but their language coverage remains skewed towards mostly high-resource languages. PolyMath \citep{wang2025polymath} covers 18 languages across four difficulty tiers, from K-12 to Olympiad, and is among the most diverse benchmarks of its kind. Nevertheless, the great majority of its languages are high-resource and already well-served by pretraining corpora. MGSM8KInstruct and the contemporaneous MSVAMP test set \citep{chen2024breaking} translate GSM8K and SVAMP into ten languages with a similar high-resource bias. MathNet \citep{alshammari2026mathnet} is the most ambitious effort to date, providing over 30K Olympiad problems across 47 countries and 17 languages. Despite this progress, the long tail of the world's languages---particularly low-resource languages---remains absent from the reasoning evaluation~\citet{ghosh2025multilingualmind}.

\subsection{Multilingual and Cross-Lingual Reasoning}

A growing amount of work examines how reasoning quality varies across languages and whether observed cross-lingual gaps reflect reasoning competence or surface artefacts. \citet{k2021analyzing} provide an early findings by category-annotating multilingual NLI and showing that transfer performance correlates with reasoning type and language similarity. More recent diagnostic work establishes that high task accuracy can mask reasoning that fails to support the model's own conclusions: \citet{ovalle2025beg} introduce a human-validated framework over 65K traces from Global-MMLU and find that non-Latin-script outputs exhibit at least twice the reasoning-answer misalignment of Latin-script outputs, with low-resource languages frequently arriving at correct answers through incoherent traces. \citet{liu2026latent} confirm this showing that latent reasoning in large reasoning models is uneven across 11 languages and broadly aligns with an English-centred internal pathway.

Several studies already attempt to close these gaps through cross-lingual reasoning transfer. A common strategy routes reasoning through a high-resource pivot, either by translating inputs---like MathOctopus~\cite{chen2024breaking}---or by aggregating traces across languages. Thus, \citet{rajaee2026best} propose Best-of-L, a cross-lingual outcome reward model that ranks reasoning candidates across languages and improves accuracy on MGSM over single-language reward modeling. 
Some work propose edits directly on parameters: MAEC \citep{chen2024extracting} decomposes language-agnostic ability-related weights from LLMs and recombines them with language-specific weights via simple addition and subtraction, transferring mathematical and scientific competencies into low-resource languages without additional training.

\subsection{Translation as a Proxy for Multilingual Benchmarks}

Given the high cost of native data collection, translation has become the dominant approach for scaling multilingual evaluation. Recent benchmarks such as BenchMAX \citep{huang2025benchmax} and MMLU-ProX \citep{xuan2025mmluprox} combine machine translation with LLM-based selection and native-speaker post-editing. For reasoning, \citet{issaka2026translation} show that LLMs translation quality strongly correlates with downstream benchmark performance, though this relationship weakens on reasoning-intensive tasks such as MGSM. Similarly, \citet{rajaee2026unlocking} demonstrate that naively applying chain-of-thought reasoning during translation degrades quality, suggesting that translation and reasoning are distinct capabilities. Machine-translated benchmarks often lose cultural and pragmatic nuance, especially in low-resource languages underrepresented in pretraining data. In contrast, we rely on a thorough human control of pre-computer translations by native speakers for all mid- and low-resource languages.


\section{\pmath Collection}
\label{sec:pluramath}




We build upon the PolyMath~\cite{wang2025polymath} dataset, which consists of four difficulty levels---low, medium, high, and top---with 125 samples per level, for a total of 500 tasks per language. Our data acquisition pipeline consists of three stages: (i) producing a first-draft translation through automatic systems; (ii) manual verification by native speakers; and (iii) automated and manual \LaTeX{} checking together with final error analysis. All annotators were fully informed about the goals of the project and were provided with written instructions, the full text of which is reproduced in Appendix~\ref{app:annotator_instructions}.

\subsection{Language Stakeholders Selection}
For every selected language we recruited native speakers (i.e., speakers whose
mother tongue is the target language) who additionally hold a higher-education
degree in Computer Science or Mathematics---at least at the Master's level, and
in most cases at the PhD level. For languages requiring more checks,
the primary stakeholder recruited additional native speakers (Bachelor's or
Master's level in Computer Science or Mathematics) to confirm the annotations
and perform a second pass over the data.

\subsection{Translation Pre-computation}

\paragraph{Translation Models and Language Pair Selection}
We asked each language stakeholder to recommend the source language and
translation system that, in their empirical experience and after a small pilot
on a handful of PolyMath tasks, yielded the most accurate translations into
their target language. The resulting language pairs and systems are summarised
in Table~\ref{tab:lang_info}.

\paragraph{Automated Translation}
For most languages, proprietary systems such as DeepL or Gemini were selected.
For a smaller set of languages, stakeholders preferred open-weights LLMs that
are specifically tuned for a given language pair and that exhibit very few
translation artefacts in their domain---for instance,
\texttt{salamandraTA-7b-instruct} for Spanish--Catalan and \texttt{sarvam-m}
for English--Hindi and English--Odia. For closed-source systems, we report the
budget spent on obtained translations in Appendix~\ref{app:budget}.


\subsection{Manual Translation Verification}

Annotators were asked to verify three properties of every translated task:
(i) the fluency and adequacy of the natural-language content;
(ii) the correctness of the mathematical terminology with respect to both the
target language and its mathematical conventions; and
(iii) the strict equivalence of the \LaTeX{} code to that of the source.

\subsection{Post-polishing}

We performed an additional automated and manual pass to verify the \LaTeX{} code
and the completeness of every translation. The automated check verifies
(i) that the \LaTeX{} code compiles---i.e.\ that all \verb|$| delimiters are
correctly opened and closed---and (ii) that all command names and reserved
keywords remain in English and unaltered. The corresponding scripts are
released for public use.\footnote{\scriptsize{\href{https://github.com/TUM-NLP/pluramath/blob/main/check_latex.ipynb}{\ghlogo~TUM-NLP/pluramath/blob/main/check\_latex.ipynb}}}

\subsection{Final Dataset}

\subsubsection{\pmath Statistics}
Table~\ref{tab:lang_info} summarizes the \pmath data creation process with dataset examples in Appendix~\ref{app:pluramth_examples}. The extent of required revisions varied substantially across languages, largely depending on language family and the maturity of available NLP and machine translation technologies. To ensure quality, we conducted a second annotation pass for nearly half of the languages.

We also compare task lengths between high-resource PolyMath languages and our target languages in Figure~\ref{fig:length_comparison_main}, with a complete analysis provided in Appendix~\ref{app:data_length}. Tokenization was performed using the \texttt{Qwen3-4B} base model. Length differences are strongly influenced by language family: Hindi, Odia, and Amharic exhibit notably longer sequences, whereas Latin- and Cyrillic-based languages remain closer to the distributions observed in high-resource counterparts.

\subsubsection{Encountered Errors}

\begin{figure*}[ht!]
    \centering
    \includegraphics[width=\linewidth]{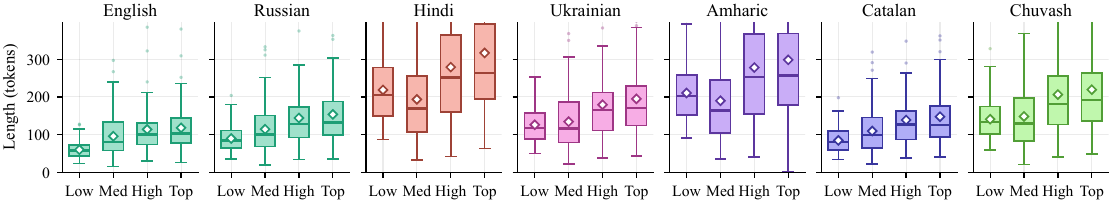}
    \caption{Comparison of math task lengths across high-resource languages from PolyMath (English and Russian) and five newly added languages in \pmath, using \texttt{Qwen3-4B} tokenization. Task lengths vary substantially across language families. A complete comparison for all languages is provided in Appendix~\ref{app:data_length}.}
    \label{fig:length_comparison_main}
\end{figure*}

\paragraph{Typical Translation Problems.}
The required corrections across languages fall into four main categories: (i) \LaTeX{} corruption, including translated command names, identifiers, and malformed \verb|$|-delimited expressions; (ii) incorrect mathematical terminology, often involving generic substitutions or cross-linguistic leakage between related languages; (iii) morphological and syntactic issues in morphologically rich languages, such as agreement, word order, and overly literal phrasing; and (iv) residual hallucinations, including untranslated tokens and incorrect entity substitutions. Correction effort varied substantially and correlated with both linguistic distance from the source language and target-language resource availability, ranging from minimal editing (e.g., Ukrainian, Greek, Hebrew) to extensive revision (e.g., Slovak, Tatar, Kazakh). In several cases, specialized language-pair models, including \texttt{TartuNLP}, \texttt{sarvamai}, and \texttt{salamandraTA}, outperformed closed-source systems due to the languages targeted adaptation.

\paragraph{Problems in the Original PolyMath}
During annotation and subsequent evaluation, we identified several inconsistencies in the original PolyMath dataset, including translation errors in non-English subsets and incorrect answers in the English data. We summarize the main issues in Appendix~\ref{app:polymath_problems} and will report them to the PolyMath HuggingFace repository via a pull request.

\section{Reasoning LLMs Benchmarking}
\label{sec:benchmarking}

To evaluate a broad range of reasoning capabilities comparing $4$ high-resource and our $18$ languages performance, we benchmark 27 state-of-the-art reasoning-oriented LLMs spanning multiple model families, training paradigms, parameter scales from sub-billion open-weight models to frontier proprietary APIs. The selection covers diverse architectural designs, post-training strategies, and levels of model accessibility. Models details, direct links to all checkpoints, and their corresponding licenses are provided in Appendix~\ref{sec:app_resources_info}.

\subsection{Prompt Design}
\label{sec:prompts}

We evaluate each model under three prompting settings: \textbf{Base} (problem and solution in the target language), \textbf{Base+EN-CoT} (target-language problem with reasoning instructed in English), and \textbf{Backtranslated} (problem translated back with NLLB model~\cite{nllb-24} into the original high-resource language with instructions in that language). The Base setting follows the original PolyMath setup~\citep{wang2025polymath}, while the latter two probe whether shifting reasoning to a high-resource language reduces performance gaps. The Base prompt is shown below; all prompt templates, language-specific instructions, and full examples for the Ukrainian case are provided in Appendix~\ref{sec:app_prompts}.


\begin{promptboxtitled}{Base Prompt Template}
\textnormal{\{Problem in the target language\}}\\[2pt]
\textnormal{\{Closing instruction in the target language\}}
\textnormal{(\textit{Original: Note: Please put the final answer in the \$$\backslash$boxed\{\}\$.})}
\end{promptboxtitled}


\subsection{Evaluation Metric}
\label{sec:evaluation_metric}

For per-level scores, we used exact match accuracy of the text in the
\verb|\boxed{}|. For the aggregated score per language overall, we
used the difficulty-weighted accuracy across all levels as defined in the
original PolyMath benchmark~\cite{wang2025polymath}. Specifically, let
$\{a_i\}_{i=1}^{4}$ denote the per-level accuracies for the four difficulty
levels (\emph{low}, \emph{medium}, \emph{high}, \emph{top}), and let the
weights be defined as $w_1 = 1$ and $w_i = 2$, $w_{i-1}$ for $i = 2, 3, 4$. The
aggregated benchmark score per language is then given by:
\begin{equation}
    \text{DW-ACC}
    \;=\; \frac{\sum_{i=1}^{4} w_i\, a_i}{\sum_{i=1}^{4} w_i}
    \;=\; \sum_{i=1}^{4} \left( \frac{2^{\,i-1}}{15}\, a_i \right).
\end{equation}

\subsection{Hyperparameters Selection}
\label{sec:hyperparameters}

\definecolor{emptyc}{RGB}{244,242,238}
\newcommand{\dcell}[2]{\renewcommand{\arraystretch}{0.72}\begin{tabular}{@{}c@{}}{\scriptsize #1}\\[-0.5pt]{\fontsize{4.4}{5}\selectfont #2}\end{tabular}}

\begin{sidewaystable*}[p]
\centering\scriptsize
\setlength{\tabcolsep}{1.6pt}\renewcommand{\arraystretch}{1.15}
\begin{tabular}{l|cccc|ccc|cccccccccccccccccc|ccc}
\toprule
 & \multicolumn{4}{c|}{\textbf{High-resource}} & \multicolumn{3}{c|}{\textbf{Summary (HR)}} & \multicolumn{18}{c|}{\textbf{Our target languages}} & \multicolumn{3}{c}{\textbf{Summary (target)}} \\
\cmidrule(lr){2-5}\cmidrule(lr){6-8}\cmidrule(lr){9-26}\cmidrule(lr){27-29}
\textbf{Model} & \rotatebox{70}{\texttt{en}} & \rotatebox{70}{\texttt{de}} & \rotatebox{70}{\texttt{ru}} & \rotatebox{70}{\texttt{es}} & \rotatebox{70}{\textbf{Avg.}} & \rotatebox{70}{Len} & \rotatebox{70}{Lang} & \rotatebox{70}{\texttt{hi}} & \rotatebox{70}{\texttt{tr}} & \rotatebox{70}{\texttt{pl}} & \rotatebox{70}{\texttt{uk}} & \rotatebox{70}{\texttt{uz}} & \rotatebox{70}{\texttt{or}} & \rotatebox{70}{\texttt{am}} & \rotatebox{70}{\texttt{el}} & \rotatebox{70}{\texttt{kk}} & \rotatebox{70}{\texttt{cs}} & \rotatebox{70}{\texttt{he}} & \rotatebox{70}{\texttt{sr}} & \rotatebox{70}{\texttt{tt}} & \rotatebox{70}{\texttt{sk}} & \rotatebox{70}{\texttt{ca}} & \rotatebox{70}{\texttt{cv}} & \rotatebox{70}{\texttt{hsb}} & \rotatebox{70}{\texttt{dsb}} & \rotatebox{70}{\textbf{Avg.}} & \rotatebox{70}{Len} & \rotatebox{70}{Lang} \\
\midrule
\multicolumn{29}{c}{\textit{Open-weight --- small ($\leq$4B)}}\\
\bottomrule
\end{tabular}
\caption{\texttt{Base} prompting aggregated across levels results on our \pmath languages vs high-resource ones from PolyMath. Each cell shows difficulty-weighted accuracy (DW-Acc, \%; large) over the answer-format compliance rate (\%; small). Cell shading encodes DW-Acc (pale\,$\rightarrow$\,deep teal, $0\rightarrow40^{+}$); a numeral is printed in \textbf{black} on light (low-score) cells and in \textbf{white} on dark (high-score) cells solely for legibility---the text colour carries no extra meaning. \underline{\textbf{Best}} and \textit{second-best} per column are highlighted. 
After each language block we report the macro-average DW-Acc, the mean$\pm$std generation length (tokens, reasoning+answer), and the dominant answer language coded as \textsc{en} when the model answers predominantly in English or \textsc{tl} when it answers in the requested target language.
}
\label{tab:base-rich}
\end{sidewaystable*}

Before running the full evaluation, we conducted a hyperparameter search on the high-resource PolyMath languages using a subset of large reasoning models (\texttt{gpt-oss-120b}, \texttt{Qwen3-235B}, and \texttt{Qwen3-30B}). We explored different \texttt{reasoning\_effort} and temperature settings, ultimately selecting \texttt{reasoning\_effort=medium} \texttt{temperature=0.1}, and \texttt{max\_completion\_tokens=2k} as the best overall configuration across languages. This setup was used for all subsequent experiments. Full hyperparameters search details and results are provided in Appendix~\ref{app:hp-search}.




\begin{figure*}[ht!]
    \centering
    \includegraphics[width=0.8\linewidth]{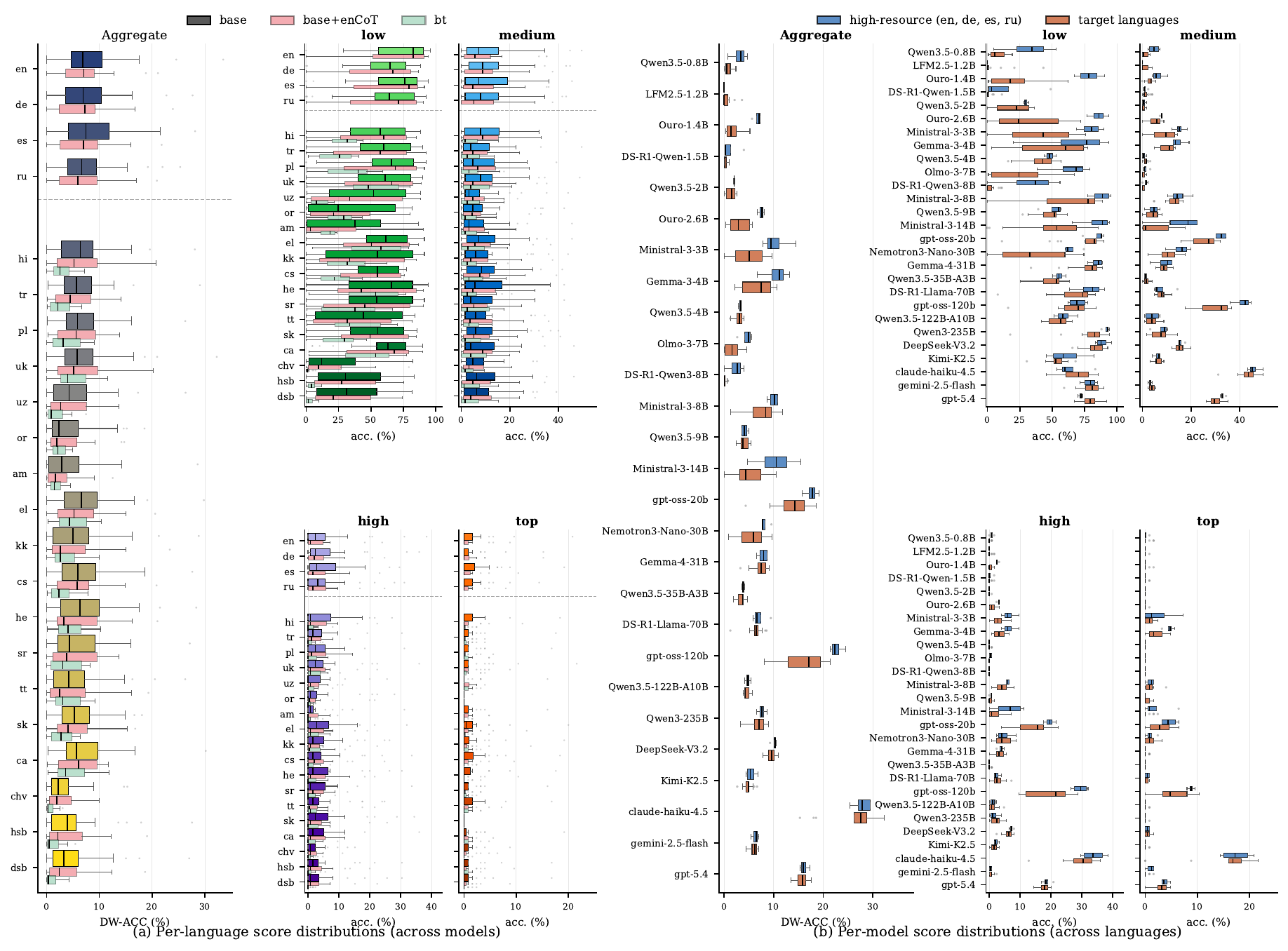}
    \caption{Math answer correctness distributions per language (a) and per model (b). We compare high-resource and target \pmath languages across all prompting settings; the same for models, we analyze performance stability for the \texttt{base} prompting and differences between languages groups. Extended versions: Figures~\ref{fig:lang_scores_distributions},~\ref{fig:models_stability_distributions} in Appendix.}
    \label{fig:combined_results_box_plots}
\end{figure*}



\subsection{\pmath Benchmarking Results}
Aggregated \texttt{DW-ACC} scores across all four difficulty levels are reported in Table~\ref{tab:base-rich}, with per-level results in Appendix~\ref{app:results_per_level}. In addition to task accuracy, we report the proportion of responses adhering to the required \verb|\boxed{}| format, aggregated reasoning length and the dominant language used in models' outputs for both high-resource and our target languages. We also report fully API costs and budget spent on our experiments in Appendix~\ref{app:budget}.

\paragraph{Math Answers Correctness} The distribution of results across languages (Figure~\ref{fig:combined_results_box_plots}) and the comparison between high-resource and \pmath languages confirm a persistent gap in mathematical reasoning performance across language categories. We compute the Spearman correlation between languages' class (Table~\ref{tab:lang_info}) and math reasoning benchmark rankings, obtaining $\rho=0.646$ with $p=0.0038$. This result indicates that \textbf{the level of language support in current language technologies strongly correlates with downstream mathematical reasoning performance}.

The magnitude of the gap varies substantially across languages. Greek and Polish show the smallest differences relative to high-resource languages ($+0.67$), whereas the average gap is $+2.15$, reaching $+4.86$ for Chuvash and Ahmaric. Smaller models often exhibit unstable performance across language groups, while larger recent systems, and especially proprietary models such as \texttt{GPT-5.4} and \texttt{Claude-Haiku-4.5}, remain considerably more stable across all evaluated languages.


\paragraph{Reasoning Length Comparison} A comparison of model reasoning lengths is shown in Appendix~\ref{app:models_reasoning_length}. The gap between language groups is \textbf{relatively modest}, with models often generating shorter outputs for \pmath languages due to failure to reach correct solutions. Notably, the best-performing models, \texttt{Claude-Haiku-4.5} and \texttt{GPT-5.4}, often produce correct answers with substantially shorter reasoning traces, indicating that longer reasoning does not necessarily correspond to better performance.

\begin{figure*}[th!]
  \centering
  \begin{minipage}[c]{0.52\textwidth}
    \centering
    \includegraphics[width=\linewidth]{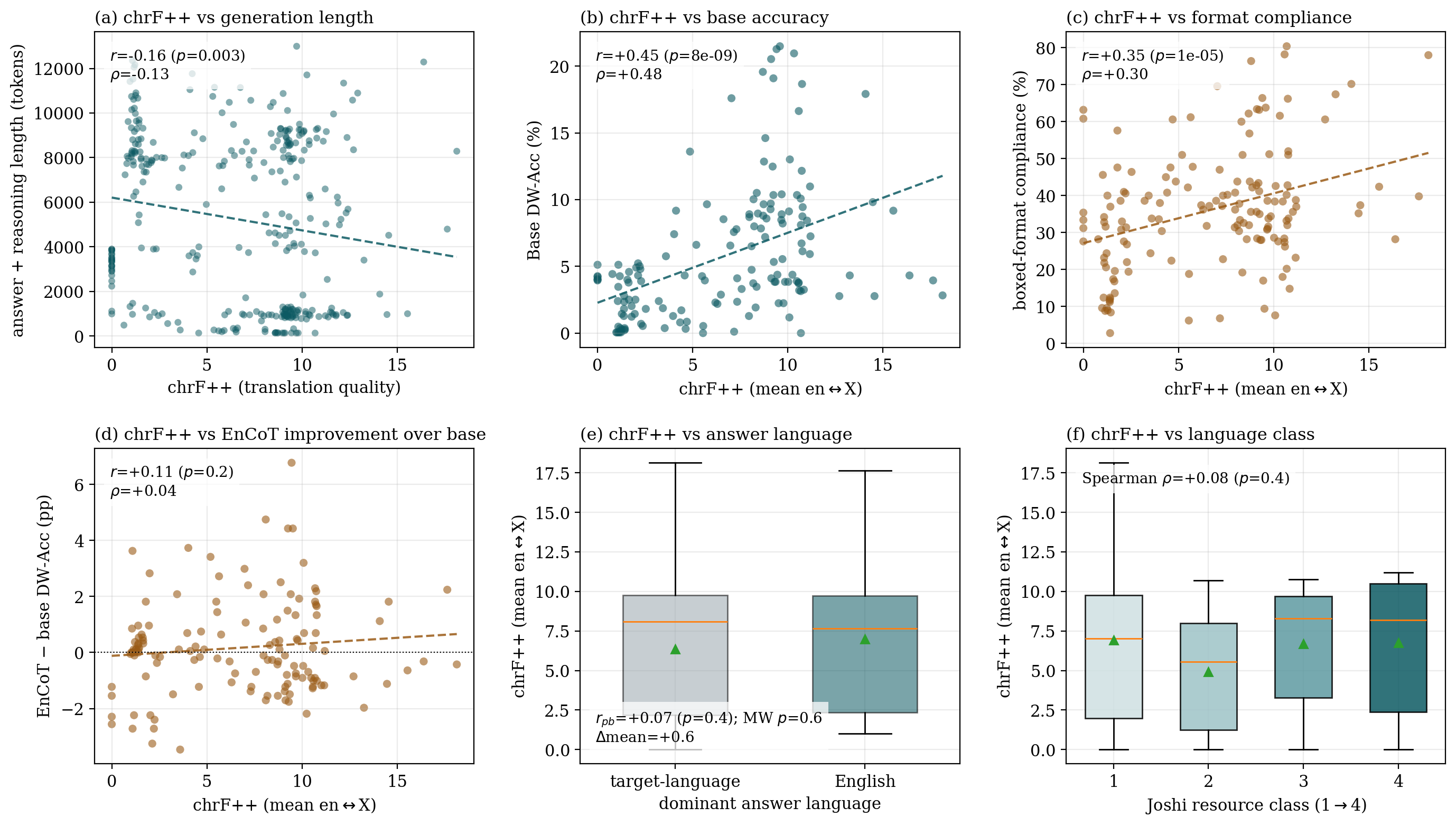}
  \end{minipage}
  \hfill
  \begin{minipage}[c]{0.46\textwidth}
    \centering
    \scriptsize
    \begin{tabular}{lcccccc}
    \toprule
    Model & Q1 & Q2 & Q3 & Q4 & Q5 & Q6 \\
    \midrule
    \multicolumn{7}{c}{\textit{High-resource (EN, RU)}} \\
    \cmidrule(lr){1-7}
    gemma-3-4B       & 0 & 99 & 94 & 77 & 45 & 77 \\
    Ministral-3-8B   & 3 & 23 & 100 & 100 & 0 & 31 \\
    Nemotron3\_30B   & 0 & 42 & 45 & 100 & 3 & 27 \\
    gpt-oss-120b     & 21 & 7 & 93 & 100 & 18 & 69 \\
    DeepSeek-V3.2    & 3 & 100 & 100 & 100 & 6 & 33 \\
    \midrule
    \multicolumn{7}{c}{\textit{\pmath (11 underrepresented languages)}} \\
    \cmidrule(lr){1-7}
    gemma-3-4B       & 2 & 89 & 49 & 66 & 6 & 73 \\
    Ministral-3-8B   & 3 & 25 & 65 & 63 & 5 & 57 \\
    Nemotron3\_30B   & 1 & 9 & 26 & 47 & 2 & 56 \\
    gpt-oss-120b     & 16 & 11 & 71 & 83 & 12 & 64 \\
    DeepSeek-V3.2    & 3 & 35 & 56 & 64 & 11 & 30 \\
    \bottomrule
    \end{tabular}
  \end{minipage}
  \caption{(A): Correlations with translations capabilities analysis. (B): Human reasoning assessment results.}
  \label{fig:correlations_humaneval}
\end{figure*}

\paragraph{Prompts Choice Comparison} Figure~\ref{fig:combined_results_box_plots} also compares the effects of different prompting strategies across languages; full results are in Appendix~\ref{app:results_ablation_prompts}. Overall, alternative prompting designs yield \textbf{limited improvements} for most models. 

\subsection{Correlation with Translation Capabilities}
To examine whether gains in reasoning are related to underlying translation capabilities, we conduct a case study on a subset of models using both reasoning and non-thinking modes for translation tasks on the FLORES+~\cite{nllb-24} \texttt{dev} and Sorbian shared-task~\citep{okabe-etal-2025-findings} splits. Correlation analyses are shown in Figure~\ref{fig:correlations_humaneval} (a), with detailed translation results in Appendix~\ref{app:reasoning_translations_details}.

Overall, we observe a \textbf{negative correlation} between \textsc{chrF}\texttt{++} scores and output length ($r=-0.16$, $p<0.01$), further confirming that longer reasoning traces do not necessarily improve translation quality. Translation quality is moderately correlated with both math task accuracy ($r=+0.45$, $p<10^{-8}$) and instruction-following performance ($r=+0.35$, $p<10^{-4}$), suggesting that some models are generally \textbf{more reliable at adhering to task requirements}. In contrast, gains from \textsc{En-CoT} prompting ($r=+0.11$, $p=0.21$), the tendency to answer in English versus the target language ($r_{pb}=+0.07$, $p=0.36$), and language class ($\rho=+0.08$, $p=0.36$) show \textbf{little correlation} with translation capability, indicating that other cross-lingual reasoning mechanisms play a larger role.


\subsection{Human Assessment of Reasoning}


Finally, we conduct a human evaluation of selected models using 12 instances per difficulty level across two high-resource and (N) target languages. Annotators assessed whether correct answers appeared despite extraction failures (Q1), reasoning language compliance with the target (Q2), coherence and step-by-step validity of reasoning (Q3), consistency between reasoning and final answers (Q4), if a correct answer appeared at some point in reasoning (Q5), and if the model finished reasoning at all (Q6). Aggregated results are reported in Figure~\ref{fig:correlations_humaneval} (b), with details provided in Appendix~\ref{app:humaneval_extended}.

\textbf{Reasoning quality declines for underrepresented languages}, with models less consistently reasoning in the target language and producing less coherent explanations. With the exception of \texttt{gemma}, models frequently switch to English during reasoning. Moreover, all models---particularly \texttt{Nemotron}---often generate disfluent or incoherent reasoning and responses, including cases when the final answer was never produced during the reasoning. Across all language groups, models also frequently fail to complete their reasoning within the \texttt{2k} token limit of our setup. These findings indicate that substantial improvements in multilingual reasoning capabilities are still required.


\section{Conclusion}
We introduced \pmath, an extension of a four-level mathematical reasoning benchmark to 18 underrepresented languages spanning six language families. We presented a multilingual benchmark construction pipeline combining automatic translation for initial drafts with rigorous human evaluation and correction. 
%
Using this benchmark, we conducted a large-scale evaluation of 27 modern reasoning LLMs. The performance gap persists between high-resource and our languages. We further compared standard prompting with EN-CoT and backtranslation strategies, finding that these interventions rarely yield consistent gains. Our analyses also indicate that stronger reasoning performance is not substantially correlated with translation capability, but is more closely associated with general instruction-following ability. Human evaluation further reveals that many models generate repetitive, incomplete, or non-fluent reasoning traces, whereas top-performing proprietary systems achieve substantially better results with shorter and concise reasoning. We hope that \pmath and its accompanying analyses will support future research on multilingual and cross-lingual reasoning.

\section*{Limitations}

Our work has several limitations. First, despite covering 18 underrepresented languages, the benchmark remains far from exhaustive. Nevertheless, we hope the presented pipeline can serve as a practical framework for extending reasoning benchmarks to additional digitally represented languages. Second, our approach depends on the availability of at least minimal machine translation resources, such as WMT-supported systems. For many extremely low-resource languages, benchmark construction would require fully manual translation and annotation, substantially increasing the cost and complexity of data creation.

Third, advanced mathematical concepts covered by higher difficulty levels may not be commonly taught in all target-language educational contexts, raising questions about the practical relevance of such evaluations for some languages. At the same time, multilingual mathematical benchmarks may help future LLM systems support access to advanced education in underrepresented languages. Fourth, our evaluation uses a single-generation pass@1 setup rather than multiple generations strategies and such scores aggregation as pass@k, which may underestimate the capabilities of some models. For some high-resource languages and difficulty levels, our experiments yielded lower results than those reported in PolyMath~\cite{wang2025polymath}, likely due to the only pass@1 over the benchmark. However, the primary objective of this work is to highlight the performance gap between high-resource and underrepresented languages and demonstrate the diversity of model behaviors across different model families and scales.

Finally, we do not investigate continued pretraining or domain adaptation on multilingual mathematical corpora. Future work could explore how to collect and balance mathematical training data for underrepresented languages in order to improve multilingual reasoning performance.

\section*{Ethics Statement}

\paragraph{Annotator Well-being}
We took explicit measures to support annotator well-being, both in workload management and recognition of contributions. All annotation was conducted on a non-profit basis by academic researchers and independent contributors motivated to support their native languages. Most contributors are acknowledged through co-authorship. To reduce annotation burden, we first generated high-quality machine translations for each language pair, allowing annotators to focus on verification and correction rather than translation from scratch. Depending on translation quality, annotators were given two to four weeks to complete their assignments with flexible scheduling. This process resulted in a manageable workload and a positive collaborative environment.

\paragraph{LLM Usage} Our evaluation involved extensive testing of dozens of LLMs across 22 languages, which likely incurred substantial computational and environmental costs. To maximize the long-term value of these experiments, we plan to release the generated reasoning traces and model outputs, subject to license compliance. We hope these resources will support future research, including knowledge distillation into smaller and more accessible models that may better serve underrepresented language communities and have a smaller environmental cost.

\paragraph{Benchmark Data Contamination} Finally, we do not systematically investigate potential benchmark contamination in large-scale models. Since PolyMath~\cite{wang2025polymath} was released in mid 2025, some newer flagship LLMs may already have been exposed to portions of the benchmark during training. In addition, certain tasks originate from widely used mathematics textbooks and well-known olympiad problem sets, making incidental memorization possible across multiple languages. A deeper analysis of contamination effects remains an important direction for future work, and we hope that the release of our multilingual benchmark will facilitate such studies beyond high-resource languages.

\section*{Acknowledgments}
We express an enormous gratitude for all annotators and supporters of the project. Firstly, we are grateful for our collaboration with the WITAJ-Sprachzentrum and thank Anita Hendrichowa, Marko Měškank, and Kryštof Peršín, in particular, for their annotations of the Upper Sorbian and Lower Sorbian splits. Secondly, the translation to Catalan has been promoted by the Aina Project. We are also grateful to Šimon Kapusta for the help in Slovak translations check. The work of authors on Czech and Slovak splits was supported by the project CZ.02.01.01/00/23\_020/0008518 of the Czech Ministry of Education, Youth and Sports. Finally, we warmly thank Alexander Antonov for the annotation of the Chuvash split.

This work was co-funded by the European Union (ERC, EPICAL, 101141712 and ERC, NG-NLG, 101039303). Views and opinions expressed are however those of the author(s) only and do not necessarily reflect those of the European Union or the European Research Council. Neither the European Union nor the granting authority can be held responsible for them.

\bibliography{custom}

@article{cobbe2021gsm8k,
  author       = {Karl Cobbe and
                  Vineet Kosaraju and
                  Mohammad Bavarian and
                  Mark Chen and
                  Heewoo Jun and
                  Lukasz Kaiser and
                  Matthias Plappert and
                  Jerry Tworek and
                  Jacob Hilton and
                  Reiichiro Nakano and
                  Christopher Hesse and
                  John Schulman},
  title        = {Training Verifiers to Solve Math Word Problems},
  journal      = {CoRR},
  volume       = {abs/2110.14168},
  year         = {2021},
  url          = {https://arxiv.org/abs/2110.14168},
  eprinttype   = {arXiv},
  eprint       = {2110.14168},
  bibsource    = {dblp computer science bibliography, https://dblp.org}
}

@inproceedings{paster2023openwebmath,
  author       = {Keiran Paster and
                  Marco Dos Santos and
                  Zhangir Azerbayev and
                  Jimmy Ba},
  title        = {OpenWebMath: An Open Dataset of High-Quality Mathematical Web Text},
  booktitle    = {The Twelfth International Conference on Learning Representations,
                  {ICLR} 2024, Vienna, Austria, May 7-11, 2024},
  publisher    = {OpenReview.net},
  year         = {2024},
  url          = {https://openreview.net/forum?id=jKHmjlpViu},
  bibsource    = {dblp computer science bibliography, https://dblp.org}
}

@article{sun2025omega,
  author       = {Yiyou Sun and
                  Shawn Hu and
                  Georgia Zhou and
                  Ken Zheng and
                  Hannaneh Hajishirzi and
                  Nouha Dziri and
                  Dawn Song},
  title        = {{OMEGA:} Can LLMs Reason Outside the Box in Math? Evaluating Exploratory,
                  Compositional, and Transformative Generalization},
  journal      = {CoRR},
  volume       = {abs/2506.18880},
  year         = {2025},
  url          = {https://doi.org/10.48550/arXiv.2506.18880},
  doi          = {10.48550/ARXIV.2506.18880},
  eprinttype   = {arXiv},
  eprint       = {2506.18880},
  bibsource    = {dblp computer science bibliography, https://dblp.org}
}

@article{yuan2025mmereasoning,
  author       = {Jiakang Yuan and
                  Tianshuo Peng and
                  Yilei Jiang and
                  Yiting Lu and
                  Renrui Zhang and
                  Kaituo Feng and
                  Chaoyou Fu and
                  Tao Chen and
                  Lei Bai and
                  Bo Zhang and
                  Xiangyu Yue},
  title        = {MME-Reasoning: {A} Comprehensive Benchmark for Logical Reasoning in
                  MLLMs},
  journal      = {CoRR},
  volume       = {abs/2505.21327},
  year         = {2025},
  url          = {https://doi.org/10.48550/arXiv.2505.21327},
  doi          = {10.48550/ARXIV.2505.21327},
  eprinttype   = {arXiv},
  eprint       = {2505.21327},
  bibsource    = {dblp computer science bibliography, https://dblp.org}
}

@article{wei2023cmath,
  author       = {Tianwen Wei and
                  Jian Luan and
                  Wei Liu and
                  Shuang Dong and
                  Bin Wang},
  title        = {{CMATH:} Can Your Language Model Pass Chinese Elementary School Math
                  Test?},
  journal      = {CoRR},
  volume       = {abs/2306.16636},
  year         = {2023},
  url          = {https://doi.org/10.48550/arXiv.2306.16636},
  doi          = {10.48550/ARXIV.2306.16636},
  eprinttype   = {arXiv},
  eprint       = {2306.16636},
  bibsource    = {dblp computer science bibliography, https://dblp.org}
}

@inproceedings{zhong2023agieval,
  author       = {Wanjun Zhong and
                  Ruixiang Cui and
                  Yiduo Guo and
                  Yaobo Liang and
                  Shuai Lu and
                  Yanlin Wang and
                  Amin Saied and
                  Weizhu Chen and
                  Nan Duan},
  editor       = {Kevin Duh and
                  Helena G{\'{o}}mez{-}Adorno and
                  Steven Bethard},
  title        = {AGIEval: {A} Human-Centric Benchmark for Evaluating Foundation Models},
  booktitle    = {Findings of the Association for Computational Linguistics: {NAACL}
                  2024, Mexico City, Mexico, June 16-21, 2024},
  series       = {Findings of {ACL}},
  pages        = {2299--2314},
  publisher    = {Association for Computational Linguistics},
  year         = {2024},
  url          = {https://doi.org/10.18653/v1/2024.findings-naacl.149},
  doi          = {10.18653/V1/2024.FINDINGS-NAACL.149},
  bibsource    = {dblp computer science bibliography, https://dblp.org}
}

@article{chen2021humaneval,
  author       = {Mark Chen and
                  Jerry Tworek and
                  Heewoo Jun and
                  Qiming Yuan and
                  Henrique Pond{\'{e}} de Oliveira Pinto and
                  Jared Kaplan and
                  Harri Edwards and
                  Yuri Burda and
                  Nicholas Joseph and
                  Greg Brockman and
                  Alex Ray and
                  Raul Puri and
                  Gretchen Krueger and
                  Michael Petrov and
                  Heidy Khlaaf and
                  Girish Sastry and
                  Pamela Mishkin and
                  Brooke Chan and
                  Scott Gray and
                  Nick Ryder and
                  Mikhail Pavlov and
                  Alethea Power and
                  Lukasz Kaiser and
                  Mohammad Bavarian and
                  Clemens Winter and
                  Philippe Tillet and
                  Felipe Petroski Such and
                  Dave Cummings and
                  Matthias Plappert and
                  Fotios Chantzis and
                  Elizabeth Barnes and
                  Ariel Herbert{-}Voss and
                  William Hebgen Guss and
                  Alex Nichol and
                  Alex Paino and
                  Nikolas Tezak and
                  Jie Tang and
                  Igor Babuschkin and
                  Suchir Balaji and
                  Shantanu Jain and
                  William Saunders and
                  Christopher Hesse and
                  Andrew N. Carr and
                  Jan Leike and
                  Joshua Achiam and
                  Vedant Misra and
                  Evan Morikawa and
                  Alec Radford and
                  Matthew Knight and
                  Miles Brundage and
                  Mira Murati and
                  Katie Mayer and
                  Peter Welinder and
                  Bob McGrew and
                  Dario Amodei and
                  Sam McCandlish and
                  Ilya Sutskever and
                  Wojciech Zaremba},
  title        = {Evaluating Large Language Models Trained on Code},
  journal      = {CoRR},
  volume       = {abs/2107.03374},
  year         = {2021},
  url          = {https://arxiv.org/abs/2107.03374},
  eprinttype   = {arXiv},
  eprint       = {2107.03374},
  bibsource    = {dblp computer science bibliography, https://dblp.org}
}

@article{balunovic2025matharena,
  author       = {Mislav Balunovic and
                  Jasper Dekoninck and
                  Ivo Petrov and
                  Nikola Jovanovic and
                  Martin T. Vechev},
  title        = {MathArena: Evaluating LLMs on Uncontaminated Math Competitions},
  journal      = {CoRR},
  volume       = {abs/2505.23281},
  year         = {2025},
  url          = {https://doi.org/10.48550/arXiv.2505.23281},
  doi          = {10.48550/ARXIV.2505.23281},
  eprinttype   = {arXiv},
  eprint       = {2505.23281},
  bibsource    = {dblp computer science bibliography, https://dblp.org}
}

@article{wang2025polymath,
  author       = {Yiming Wang and
                  Pei Zhang and
                  Jialong Tang and
                  Haoran Wei and
                  Baosong Yang and
                  Rui Wang and
                  Chenshu Sun and
                  Feitong Sun and
                  Jiran Zhang and
                  Junxuan Wu and
                  Qiqian Cang and
                  Yichang Zhang and
                  Fei Huang and
                  Junyang Lin and
                  Fei Huang and
                  Jingren Zhou},
  title        = {PolyMath: Evaluating Mathematical Reasoning in Multilingual Contexts},
  journal      = {CoRR},
  volume       = {abs/2504.18428},
  year         = {2025},
  url          = {https://doi.org/10.48550/arXiv.2504.18428},
  doi          = {10.48550/ARXIV.2504.18428},
  eprinttype   = {arXiv},
  eprint       = {2504.18428},
  bibsource    = {dblp computer science bibliography, https://dblp.org}
}

@inproceedings{chen2024breaking,
  author       = {Nuo Chen and
                  Zinan Zheng and
                  Ning Wu and
                  Ming Gong and
                  Dongmei Zhang and
                  Jia Li},
  editor       = {Yaser Al{-}Onaizan and
                  Mohit Bansal and
                  Yun{-}Nung Chen},
  title        = {Breaking Language Barriers in Multilingual Mathematical Reasoning:
                  Insights and Observations},
  booktitle    = {Findings of the Association for Computational Linguistics: {EMNLP}
                  2024, Miami, Florida, USA, November 12-16, 2024},
  series       = {Findings of {ACL}},
  pages        = {7001--7016},
  publisher    = {Association for Computational Linguistics},
  year         = {2024},
  url          = {https://doi.org/10.18653/v1/2024.findings-emnlp.411},
  doi          = {10.18653/V1/2024.FINDINGS-EMNLP.411},
  bibsource    = {dblp computer science bibliography, https://dblp.org}
}

@inproceedings{alshammari2026mathnet,
  title     = {{MathNet}: A Global Multimodal Benchmark for Mathematical Reasoning and Retrieval},
  author    = {Alshammari, Shaden and Wen, Kevin and Zainal, Abrar and
               Hamilton, Mark and Safaei, Navid and Albarakati, Sultan and
               Freeman, William T. and Torralba, Antonio},
  booktitle = {International Conference on Learning Representations (ICLR)},
  year      = {2026},
  note      = {arXiv:2604.18584},
  url       = {https://arxiv.org/abs/2604.18584}
}

@inproceedings{huang2025benchmax,
  author       = {Xu Huang and
                  Wenhao Zhu and
                  Hanxu Hu and
                  Conghui He and
                  Lei Li and
                  Shujian Huang and
                  Fei Yuan},
  editor       = {Christos Christodoulopoulos and
                  Tanmoy Chakraborty and
                  Carolyn Rose and
                  Violet Peng},
  title        = {BenchMAX: {A} Comprehensive Multilingual Evaluation Suite for Large
                  Language Models},
  booktitle    = {Findings of the Association for Computational Linguistics: {EMNLP}
                  2025, Suzhou, China, November 4-9, 2025},
  pages        = {16751--16774},
  publisher    = {Association for Computational Linguistics},
  year         = {2025},
  url          = {https://aclanthology.org/2025.findings-emnlp.909/},
  bibsource    = {dblp computer science bibliography, https://dblp.org}
}

@inproceedings{xuan2025mmluprox,
  author       = {Weihao Xuan and
                  Rui Yang and
                  Heli Qi and
                  Qingcheng Zeng and
                  Yunze Xiao and
                  Aosong Feng and
                  Dairui Liu and
                  Yun Xing and
                  Junjue Wang and
                  Fan Gao and
                  Jinghui Lu and
                  Yuang Jiang and
                  Huitao Li and
                  Xin Li and
                  Kunyu Yu and
                  Ruihai Dong and
                  Shangding Gu and
                  Yuekang Li and
                  Xiaofei Xie and
                  Felix Juefei{-}Xu and
                  Foutse Khomh and
                  Osamu Yoshie and
                  Qingyu Chen and
                  Douglas Teodoro and
                  Nan Liu and
                  Randy Goebel and
                  Lei Ma and
                  Edison Marrese{-}Taylor and
                  Shijian Lu and
                  Yusuke Iwasawa and
                  Yutaka Matsuo and
                  Irene Li},
  editor       = {Christos Christodoulopoulos and
                  Tanmoy Chakraborty and
                  Carolyn Rose and
                  Violet Peng},
  title        = {MMLU-ProX: {A} Multilingual Benchmark for Advanced Large Language
                  Model Evaluation},
  booktitle    = {Proceedings of the 2025 Conference on Empirical Methods in Natural
                  Language Processing, {EMNLP} 2025, Suzhou, China, November 4-9, 2025},
  pages        = {1513--1532},
  publisher    = {Association for Computational Linguistics},
  year         = {2025},
  url          = {https://doi.org/10.18653/v1/2025.emnlp-main.79},
  doi          = {10.18653/V1/2025.EMNLP-MAIN.79},
  bibsource    = {dblp computer science bibliography, https://dblp.org}
}

@inproceedings{joshi2020state,
  author       = {Pratik Joshi and
                  Sebastin Santy and
                  Amar Budhiraja and
                  Kalika Bali and
                  Monojit Choudhury},
  editor       = {Dan Jurafsky and
                  Joyce Chai and
                  Natalie Schluter and
                  Joel R. Tetreault},
  title        = {The State and Fate of Linguistic Diversity and Inclusion in the {NLP}
                  World},
  booktitle    = {Proceedings of the 58th Annual Meeting of the Association for Computational
                  Linguistics, {ACL} 2020, Online, July 5-10, 2020},
  pages        = {6282--6293},
  publisher    = {Association for Computational Linguistics},
  year         = {2020},
  url          = {https://doi.org/10.18653/v1/2020.acl-main.560},
  doi          = {10.18653/V1/2020.ACL-MAIN.560},
  bibsource    = {dblp computer science bibliography, https://dblp.org}
}

@inproceedings{ghosh2025multilingualmind,
  author       = {Akash Ghosh and
                  Debayan Datta and
                  Sriparna Saha and
                  Chirag Agarwal},
  editor       = {Christos Christodoulopoulos and
                  Tanmoy Chakraborty and
                  Carolyn Rose and
                  Violet Peng},
  title        = {A Survey of Multilingual Reasoning in Language Models},
  booktitle    = {Findings of the Association for Computational Linguistics: {EMNLP}
                  2025, Suzhou, China, November 4-9, 2025},
  pages        = {8920--8936},
  publisher    = {Association for Computational Linguistics},
  year         = {2025},
  url          = {https://aclanthology.org/2025.findings-emnlp.474/},
  bibsource    = {dblp computer science bibliography, https://dblp.org}
}

@inproceedings{k2021analyzing,
    title = "Analyzing the Effects of Reasoning Types on Cross-Lingual Transfer Performance",
    author = "K, Karthikeyan  and
      Sathe, Aalok  and
      Aditya, Somak  and
      Choudhury, Monojit",
    editor = "Ataman, Duygu  and
      Birch, Alexandra  and
      Conneau, Alexis  and
      Firat, Orhan  and
      Ruder, Sebastian  and
      Sahin, Gozde Gul",
    booktitle = "Proceedings of the 1st Workshop on Multilingual Representation Learning",
    month = nov,
    year = "2021",
    address = "Punta Cana, Dominican Republic",
    publisher = "Association for Computational Linguistics",
    url = "https://aclanthology.org/2021.mrl-1.8/",
    doi = "10.18653/v1/2021.mrl-1.8",
    pages = "86--95"
}

@article{ovalle2025beg,
  author       = {Anaelia Ovalle and
                  Candace Ross and
                  Sebastian Ruder and
                  Adina Williams and
                  Karen Ullrich and
                  Mark Ibrahim and
                  Levent Sagun},
  title        = {Beg to Differ: Understanding Reasoning-Answer Misalignment Across
                  Languages},
  journal      = {CoRR},
  volume       = {abs/2512.22712},
  year         = {2025},
  url          = {https://doi.org/10.48550/arXiv.2512.22712},
  doi          = {10.48550/ARXIV.2512.22712},
  eprinttype   = {arXiv},
  eprint       = {2512.22712},
  bibsource    = {dblp computer science bibliography, https://dblp.org}
}

@article{liu2026latent,
  author       = {Yihong Liu and
                  Raoyuan Zhao and
                  Hinrich Sch{\"{u}}tze and
                  Michael A. Hedderich},
  title        = {Large Reasoning Models Are (Not Yet) Multilingual Latent Reasoners},
  journal      = {CoRR},
  volume       = {abs/2601.02996},
  year         = {2026},
  url          = {https://doi.org/10.48550/arXiv.2601.02996},
  doi          = {10.48550/ARXIV.2601.02996},
  eprinttype   = {arXiv},
  eprint       = {2601.02996},
  bibsource    = {dblp computer science bibliography, https://dblp.org}
}

@inproceedings{rajaee2026best,
  author       = {Sara Rajaee and
                  Rochelle Choenni and
                  Ekaterina Shutova and
                  Christof Monz},
  editor       = {Vera Demberg and
                  Kentaro Inui and
                  Llu{\'{\i}}s Marquez},
  title        = {Best-of-L: Cross-Lingual Reward Modeling for Mathematical Reasoning},
  booktitle    = {Findings of the Association for Computational Linguistics: {EACL}
                  2026, Rabat, Morocco, March 24-29, 2026},
  series       = {Findings of {ACL}},
  pages        = {1930--1939},
  publisher    = {Association for Computational Linguistics},
  year         = {2026},
  url          = {https://aclanthology.org/2026.findings-eacl.99/},
  bibsource    = {dblp computer science bibliography, https://dblp.org}
}

@inproceedings{chen2024extracting,
  author       = {Zhipeng Chen and
                  Kun Zhou and
                  Liang Song and
                  Wayne Xin Zhao and
                  Bingning Wang and
                  Weipeng Chen and
                  Ji{-}Rong Wen},
  editor       = {Christos Christodoulopoulos and
                  Tanmoy Chakraborty and
                  Carolyn Rose and
                  Violet Peng},
  title        = {Extracting and Combining Abilities For Building Multi-lingual Ability-enhanced
                  Large Language Models},
  booktitle    = {Proceedings of the 2025 Conference on Empirical Methods in Natural
                  Language Processing, {EMNLP} 2025, Suzhou, China, November 4-9, 2025},
  pages        = {17563--17580},
  publisher    = {Association for Computational Linguistics},
  year         = {2025},
  url          = {https://doi.org/10.18653/v1/2025.emnlp-main.887},
  doi          = {10.18653/V1/2025.EMNLP-MAIN.887},
  bibsource    = {dblp computer science bibliography, https://dblp.org}
}

@article{issaka2026translation,
  author       = {Sheriff Issaka and
                  Erick Rosas Gonzalez and
                  Lieqi Liu and
                  Evans Kofi Agyei and
                  Lucas Bandarkar and
                  Nanyun Peng and
                  David Ifeoluwa Adelani and
                  Francisco Guzm{\'{a}}n and
                  Saadia Gabriel},
  title        = {Translation as a Scalable Proxy for Multilingual Evaluation},
  journal      = {CoRR},
  volume       = {abs/2601.11778},
  year         = {2026},
  url          = {https://doi.org/10.48550/arXiv.2601.11778},
  doi          = {10.48550/ARXIV.2601.11778},
  eprinttype   = {arXiv},
  eprint       = {2601.11778},
  bibsource    = {dblp computer science bibliography, https://dblp.org}
}

@article{rajaee2026unlocking,
  author       = {Sara Rajaee and
                  Sebastian Vincent and
                  Alexandre Berard and
                  Marzieh Fadaee and
                  Kelly Marchisio and
                  Tom Kocmi},
  title        = {Unlocking Reasoning Capability on Machine Translation in Large Language
                  Models},
  journal      = {CoRR},
  volume       = {abs/2602.14763},
  year         = {2026},
  url          = {https://doi.org/10.48550/arXiv.2602.14763},
  doi          = {10.48550/ARXIV.2602.14763},
  eprinttype   = {arXiv},
  eprint       = {2602.14763},
  bibsource    = {dblp computer science bibliography, https://dblp.org}
}

@article{qwen3,
  author       = {Qwen Team},
  title        = {Qwen3 Technical Report},
  journal      = {CoRR},
  volume       = {abs/2505.09388},
  year         = {2025},
  url          = {https://doi.org/10.48550/arXiv.2505.09388},
  doi          = {10.48550/ARXIV.2505.09388},
  eprinttype   = {arXiv},
  eprint       = {2505.09388},
  bibsource    = {dblp computer science bibliography, https://dblp.org}
}

@article{lfm,
  author       = {Alexander Amini and
                  Anna Banaszak and
                  Harold Benoit and
                  Arthur B{\"{o}}{\"{o}}k and
                  Tarek Dakhran and
                  Song Duong and
                  Alfred Eng and
                  Fernando Fernandes and
                  Marc H{\"{a}}rk{\"{o}}nen and
                  Anne Harrington and
                  Ramin M. Hasani and
                  Saniya Karwa and
                  Yuri Khrustalev and
                  Maxime Labonne and
                  Mathias Lechner and
                  Valentine Lechner and
                  Simon Lee and
                  Zetian Li and
                  Noel Loo and
                  Jacob Marks and
                  Edoardo Mosca and
                  Samuel J. Paech and
                  Paul Pak and
                  Rom N. Parnichkun and
                  Alex Quach and
                  Ryan Rogers and
                  Daniela Rus and
                  Nayan Saxena and
                  Bettina Schlager and
                  Tim Seyde and
                  Jimmy T. H. Smith and
                  Aditya Tadimeti and
                  Neehal Tumma},
  title        = {{LFM2} Technical Report},
  journal      = {CoRR},
  volume       = {abs/2511.23404},
  year         = {2025},
  url          = {https://doi.org/10.48550/arXiv.2511.23404},
  doi          = {10.48550/ARXIV.2511.23404},
  eprinttype   = {arXiv},
  eprint       = {2511.23404},
  bibsource    = {dblp computer science bibliography, https://dblp.org}
}

@article{ouro,
  author       = {Rui{-}Jie Zhu and
                  Zixuan Wang and
                  Kai Hua and
                  Tianyu Zhang and
                  Ziniu Li and
                  Haoran Que and
                  Boyi Wei and
                  Zixin Wen and
                  Fan Yin and
                  He Xing and
                  Lu Li and
                  Jiajun Shi and
                  Kaijing Ma and
                  Shanda Li and
                  Taylor Kergan and
                  Andrew Smith and
                  Xingwei Qu and
                  Mude Hui and
                  Bohong Wu and
                  Qiyang Min and
                  Hongzhi Huang and
                  Xun Zhou and
                  Wei Ye and
                  Jiaheng Liu and
                  Jian Yang and
                  Yunfeng Shi and
                  Chenghua Lin and
                  Enduo Zhao and
                  Tianle Cai and
                  Ge Zhang and
                  Wenhao Huang and
                  Yoshua Bengio and
                  Jason Eshraghian},
  title        = {Scaling Latent Reasoning via Looped Language Models},
  journal      = {CoRR},
  volume       = {abs/2510.25741},
  year         = {2025},
  url          = {https://doi.org/10.48550/arXiv.2510.25741},
  doi          = {10.48550/ARXIV.2510.25741},
  eprinttype   = {arXiv},
  eprint       = {2510.25741},
  bibsource    = {dblp computer science bibliography, https://dblp.org}
}

@article{deepseekr1,
  author       = {DeepSeek{-}AI},
  title        = {DeepSeek-R1: Incentivizing Reasoning Capability in LLMs via Reinforcement
                  Learning},
  journal      = {CoRR},
  volume       = {abs/2501.12948},
  year         = {2025},
  url          = {https://doi.org/10.48550/arXiv.2501.12948},
  doi          = {10.48550/ARXIV.2501.12948},
  eprinttype   = {arXiv},
  eprint       = {2501.12948},
  bibsource    = {dblp computer science bibliography, https://dblp.org}
}

@article{mistral,
  author       = {Albert Q. Jiang and
                  Alexandre Sablayrolles and
                  Arthur Mensch and
                  Chris Bamford and
                  Devendra Singh Chaplot and
                  Diego de Las Casas and
                  Florian Bressand and
                  Gianna Lengyel and
                  Guillaume Lample and
                  Lucile Saulnier and
                  L{\'{e}}lio Renard Lavaud and
                  Marie{-}Anne Lachaux and
                  Pierre Stock and
                  Teven Le Scao and
                  Thibaut Lavril and
                  Thomas Wang and
                  Timoth{\'{e}}e Lacroix and
                  William El Sayed},
  title        = {Mistral 7B},
  journal      = {CoRR},
  volume       = {abs/2310.06825},
  year         = {2023},
  url          = {https://doi.org/10.48550/arXiv.2310.06825},
  doi          = {10.48550/ARXIV.2310.06825},
  eprinttype   = {arXiv},
  eprint       = {2310.06825},
  bibsource    = {dblp computer science bibliography, https://dblp.org}
}

@article{ministral,
  author       = {Mistral AI},
  title        = {Ministral 3},
  journal      = {CoRR},
  volume       = {abs/2601.08584},
  year         = {2026},
  url          = {https://doi.org/10.48550/arXiv.2601.08584},
  doi          = {10.48550/ARXIV.2601.08584},
  eprinttype   = {arXiv},
  eprint       = {2601.08584},
  bibsource    = {dblp computer science bibliography, https://dblp.org}
}

@article{gemma,
  author       = {Gemma Team},
  title        = {Gemma 3 Technical Report},
  journal      = {CoRR},
  volume       = {abs/2503.19786},
  year         = {2025},
  url          = {https://doi.org/10.48550/arXiv.2503.19786},
  doi          = {10.48550/ARXIV.2503.19786},
  eprinttype   = {arXiv},
  eprint       = {2503.19786},
  bibsource    = {dblp computer science bibliography, https://dblp.org}
}

@article{olmo,
  author       = {Team OLMo and
                  Pete Walsh and
                  Luca Soldaini and
                  Dirk Groeneveld and
                  Kyle Lo and
                  Shane Arora and
                  Akshita Bhagia and
                  Yuling Gu and
                  Shengyi Huang and
                  Matt Jordan and
                  Nathan Lambert and
                  Dustin Schwenk and
                  Oyvind Tafjord and
                  Taira Anderson and
                  David Atkinson and
                  Faeze Brahman and
                  Christopher Clark and
                  Pradeep Dasigi and
                  Nouha Dziri and
                  Michal Guerquin and
                  Hamish Ivison and
                  Pang Wei Koh and
                  Jiacheng Liu and
                  Saumya Malik and
                  William Merrill and
                  Lester James V. Miranda and
                  Jacob Morrison and
                  Tyler Murray and
                  Crystal Nam and
                  Valentina Pyatkin and
                  Aman Rangapur and
                  Michael Schmitz and
                  Sam Skjonsberg and
                  David Wadden and
                  Christopher Wilhelm and
                  Michael Wilson and
                  Luke Zettlemoyer and
                  Ali Farhadi and
                  Noah A. Smith and
                  Hannaneh Hajishirzi},
  title        = {2 OLMo 2 Furious},
  journal      = {CoRR},
  volume       = {abs/2501.00656},
  year         = {2025},
  url          = {https://doi.org/10.48550/arXiv.2501.00656},
  doi          = {10.48550/ARXIV.2501.00656},
  eprinttype   = {arXiv},
  eprint       = {2501.00656},
  bibsource    = {dblp computer science bibliography, https://dblp.org}
}

@article{gptoss,
  author       = {OpenAI},
  title        = {gpt-oss-120b {\&} gpt-oss-20b Model Card},
  journal      = {CoRR},
  volume       = {abs/2508.10925},
  year         = {2025},
  url          = {https://doi.org/10.48550/arXiv.2508.10925},
  doi          = {10.48550/ARXIV.2508.10925},
  eprinttype   = {arXiv},
  eprint       = {2508.10925},
  bibsource    = {dblp computer science bibliography, https://dblp.org}
}

@article{kimik2,
  author       = {Kimi Team},
  title        = {Kimi {K2:} Open Agentic Intelligence},
  journal      = {CoRR},
  volume       = {abs/2507.20534},
  year         = {2025},
  url          = {https://doi.org/10.48550/arXiv.2507.20534},
  doi          = {10.48550/ARXIV.2507.20534},
  eprinttype   = {arXiv},
  eprint       = {2507.20534},
  bibsource    = {dblp computer science bibliography, https://dblp.org}
}

@techreport{claude4,
  title       = {System {C}ard: Claude {O}pus 4 and {C}laude {S}onnet 4},
  author      = {{Anthropic}},
  institution = {Anthropic},
  year        = {2025},
  url         = {https://www-cdn.anthropic.com/6d8a8055020700718b0c49369f60816ba2a7c285.pdf},
  note        = {Accessed: May 2026}
}

@article{gpt5,
  author       = {OpenAI},
  title        = {OpenAI {GPT-5} System Card},
  journal      = {CoRR},
  volume       = {abs/2601.03267},
  year         = {2026},
  url          = {https://doi.org/10.48550/arXiv.2601.03267},
  doi          = {10.48550/ARXIV.2601.03267},
  eprinttype   = {arXiv},
  eprint       = {2601.03267},
  bibsource    = {dblp computer science bibliography, https://dblp.org}
}

@article{gemini,
  author       = {Gemini Team},
  title        = {Gemini: {A} Family of Highly Capable Multimodal Models},
  journal      = {CoRR},
  volume       = {abs/2312.11805},
  year         = {2023},
  url          = {https://doi.org/10.48550/arXiv.2312.11805},
  doi          = {10.48550/ARXIV.2312.11805},
  eprinttype   = {arXiv},
  eprint       = {2312.11805},
  bibsource    = {dblp computer science bibliography, https://dblp.org}
}

@article{nllb-24,
    author="{NLLB Team} and Costa-juss{\`a}, Marta R. and Cross, James and {\c{C}}elebi, Onur and Elbayad, Maha and Heafield, Kenneth and Heffernan, Kevin and Kalbassi, Elahe and Lam, Janice and Licht, Daniel and Maillard, Jean and Sun, Anna and Wang, Skyler and Wenzek, Guillaume and Youngblood, Al and Akula, Bapi and Barrault, Loic and Gonzalez, Gabriel Mejia and Hansanti, Prangthip and Hoffman, John and Jarrett, Semarley and Sadagopan, Kaushik Ram and Rowe, Dirk and Spruit, Shannon and Tran, Chau and Andrews, Pierre and Ayan, Necip Fazil and Bhosale, Shruti and Edunov, Sergey and Fan, Angela and Gao, Cynthia and Goswami, Vedanuj and Guzm{\'a}n, Francisco and Koehn, Philipp and Mourachko, Alexandre and Ropers, Christophe and Saleem, Safiyyah and Schwenk, Holger and Wang, Jeff",
    title="Scaling neural machine translation to 200 languages",
    journal="Nature",
    year="2024",
    volume="630",
    number="8018",
    pages="841--846",
    issn="1476-4687",
    doi="10.1038/s41586-024-07335-x",
    url="https://doi.org/10.1038/s41586-024-07335-x"
}

@article{DBLP:journals/corr/abs-2604-24954,
  author       = {NVIDIA},
  title        = {Nemotron 3 Nano Omni: Efficient and Open Multimodal Intelligence},
  journal      = {CoRR},
  volume       = {abs/2604.24954},
  year         = {2026},
  url          = {https://doi.org/10.48550/arXiv.2604.24954},
  doi          = {10.48550/ARXIV.2604.24954},
  eprinttype   = {arXiv},
  eprint       = {2604.24954},
  bibsource    = {dblp computer science bibliography, https://dblp.org}
}

@inproceedings{joshi-etal-2020-state,
    title = "The State and Fate of Linguistic Diversity and Inclusion in the {NLP} World",
    author = "Joshi, Pratik  and
      Santy, Sebastin  and
      Budhiraja, Amar  and
      Bali, Kalika  and
      Choudhury, Monojit",
    editor = "Jurafsky, Dan  and
      Chai, Joyce  and
      Schluter, Natalie  and
      Tetreault, Joel",
    booktitle = "Proceedings of the 58th Annual Meeting of the Association for Computational Linguistics",
    month = jul,
    year = "2020",
    address = "Online",
    publisher = "Association for Computational Linguistics",
    url = "https://aclanthology.org/2020.acl-main.560/",
    doi = "10.18653/v1/2020.acl-main.560",
    pages = "6282--6293"
}

@inproceedings{okabe-etal-2025-findings,
    title = "Findings of the {WMT} 2025 Shared Task {LLM}s with Limited Resources for {S}lavic Languages: {MT} and {QA}",
    author = "Okabe, Shu  and
      Dementieva, Daryna  and
      Di Marco, Marion  and
      Edman, Lukas  and
      Haemmerl, Katharina  and
      M{\v{e}}{\v{s}}kank, Marko  and
      Hendrichowa, Anita  and
      Fraser, Alexander",
    editor = "Haddow, Barry  and
      Kocmi, Tom  and
      Koehn, Philipp  and
      Monz, Christof",
    booktitle = "Proceedings of the Tenth Conference on Machine Translation",
    month = nov,
    year = "2025",
    address = "Suzhou, China",
    publisher = "Association for Computational Linguistics",
    url = "https://aclanthology.org/2025.wmt-1.27/",
    doi = "10.18653/v1/2025.wmt-1.27",
    pages = "503--519",
    ISBN = "979-8-89176-341-8"
}

\onecolumn
\appendix

\section{Details and Licensing of Resources}
\label{sec:app_resources_info}

In this section, we provide details about used LLMs for the experiments as well as all details, direct links, and licenses of all used resources.

\subsection{Short Summary of LLMs Families Used for the Experiments}
\label{app:reasoning_llms}

\paragraph{Qwen} The Qwen series from Alibaba~\citep{qwen3} has become one of the most widely adopted open-weight backbones for reasoning research, owing to its strong multilingual coverage, competitive math and code performance, and an actively maintained range of dense and Mixture-of-Experts variants. We include Qwen3.5 checkpoints at 0.8B, 2B, 4B, 9B, and the 35B-A3B, 122B-A10B MoE configurations as well as the older version of Qwen3 with the biggest 235B-A22B MoE variant.

\paragraph{LiquidAI LFM} The Liquid Foundation Models~\citep{lfm} depart from the standard Transformer template by combining attention with structured state-space and convolutional blocks, yielding favourable memory and latency characteristics at small scales. The LFM2.5-1.2B-Thinking variant is interesting as a low-resource reasoning baseline.

\paragraph{ByteDance Ouro} The Ouro family~\citep{ouro} introduces looped language models, in which a shared block of parameters is iterated multiple times during inference to amortise reasoning depth without increasing the parameter count. We test the 1.4B and 2.6B Thinking checkpoints to measure how this recurrence-style compute trades off against conventional depth-scaling.

\paragraph{DeepSeek} DeepSeek-R1~\citep{deepseekr1} pioneered large-scale reinforcement learning with verifiable rewards (RLVR) for open-weight reasoning, and its distilled checkpoints have become standard reference points. We evaluate the R1-Distill-Qwen-1.5B, R1-0528-Qwen3-8B, and R1-Distill-LLama-70B distillates as well as full V3.2 MoE version, providing both small-scale distillation evidence and a frontier open MoE comparison.

\paragraph{Ministral / Mistral} Mistral's Ministral line~\citep{mistral} targets efficient on-device and edge inference while inheriting the reasoning post-training recipe from the larger Mistral models. We include the recent 3B, 8B, and 14B Reasoning-2512 releases.

\paragraph{Google Gemma} Gemma~\citep{gemma} is Google DeepMind's open-weight family derived from the same research stack as Gemini. The Gemma-3 4B and Gemma-4 31B variants allow us to compare a Gemini-aligned training recipe against other lineages at matched parameter budgets.

\texttt{Nemotron 3 Nano Omni} is a multimodal large language model developed by NVIDIA~\cite{DBLP:journals/corr/abs-2604-24954} that jointly processes video, audio, images, and text for tasks such as question answering, summarization, transcription, OCR, and document understanding. We use the Reasoning variant of the Nemotron series Nemotron-3-Nano-Omni-30B-A3B-Reasoning.

\paragraph{AllenAI OLMo} OLMo~\citep{olmo} is, to our knowledge, the most fully open frontier-style series, releasing not only weights but also data, training code, and intermediate checkpoints. The OLMo-3-7B-Think allow examining whether full transparency comes at a measurable reasoning cost.

\paragraph{OpenAI gpt-oss} The gpt-oss series~\citep{gptoss} are OpenAI's open-weight Mixture-of-Experts re
asoning models, released under a permissive license and explicitly tuned for chain-of-thought workloads. The 20B and 120B checkpoints anchor the open-weight side of the comparison at the upper end.

\paragraph{Moonshot Kimi} Kimi-K2.5-Thinking~\citep{kimik2} is Moonshot AI's frontier MoE reasoning model, notable for its very long native context window and aggressive RL post-training.


\paragraph{Closed-source frontier models} To upper-bound the comparison, we include API-only models from the three major frontier labs: Anthropic's \textbf{Claude}-Haiku-4.5~\citep{claude4}, OpenAI's \textbf{GPT}-5.4~\citep{gpt5}, and Google's \textbf{Gemini}-2.5-Flash~\citep{gemini}.

\subsection{Direct Links and Licensing Information for All Resources Used in This Work}
\label{sec:app_licenses}

Below is an overview of the licenses and direct repositories of every resource used in this work (Table~\ref{tab:overview-license}). The licenses associated with the models and datasets are consistent with the intended use of conducting academic research on multilingual mathematical reasoning for positive societal impact. Closed-source models are accessed exclusively through their respective official APIs in compliance with each provider's terms of service.

\begin{table*}[h!]
\centering
\scriptsize
\begin{tabular}{p{4.5cm}p{3.2cm}p{7.3cm}}
\toprule
Resource & License & Resource Link \\
\midrule
\multicolumn{3}{c}{\textit{Datasets}} \\
\midrule
\pmath & Apache-2.0 & \href{https://hf.co/datasets/tum-nlp/PluraMath}{hf.co/datasets/tum-nlp/PluraMath} \\
PolyMath & Apache-2.0 & \href{https://hf.co/datasets/Qwen/PolyMath}{hf.co/datasets/Qwen/PolyMath}
\newline \href{https://github.com/QwenLM/PolyMath}{github.com/QwenLM/PolyMath} \\
FLORES+ & CC-BY-SA-4.0 & \href{https://hf.co/datasets/openlanguagedata/flores_plus}{hf.co/datasets/openlanguagedata/flores\_plus} \\
Sorbian MT Dev Set & CC-BY-NC-SA-4.0 & \href{https://github.com/TUM-NLP/llms-limited-resources2025/tree/main/Sorbian}{github.com/TUM-NLP/llms-limited-resources2025/tree/main/Sorbian} \\
\midrule
\multicolumn{3}{c}{\textit{Open-weight models — small ($\leq$4B)}} \\
\midrule
Qwen3.5-0.8B                  & Apache-2.0       & \href{https://hf.co/Qwen/Qwen3.5-0.8B}{hf.co/Qwen/Qwen3.5-0.8B} \\
LFM2.5-1.2B-Thinking          & LFM Open License & \href{https://hf.co/LiquidAI/LFM2.5-1.2B-Thinking}{hf.co/LiquidAI/LFM2.5-1.2B-Thinking} \\
Ouro-1.4B-Thinking            & Apache-2.0       & \href{https://hf.co/ByteDance/Ouro-1.4B-Thinking}{hf.co/ByteDance/Ouro-1.4B-Thinking} \\
R1-Distill-Qwen-1.5B & MIT              & \href{https://hf.co/deepseek-ai/DeepSeek-R1-Distill-Qwen-1.5B}{hf.co/deepseek-ai/DeepSeek-R1-Distill-Qwen-1.5B} \\
Qwen3.5-2B                    & Apache-2.0       & \href{https://hf.co/Qwen/Qwen3.5-2B}{hf.co/Qwen/Qwen3.5-2B} \\
Ouro-2.6B            & Apache-2.0       & \href{https://hf.co/ByteDance/Ouro-2.6B-Thinking}{hf.co/ByteDance/Ouro-2.6B-Thinking} \\
Ministral-3-3B & Mistral Research License & \href{https://hf.co/mistralai/Ministral-3-3B-Reasoning-2512}{hf.co/mistralai/Ministral-3-3B-Reasoning-2512} \\
Gemma-3-E4B                   & Gemma License    & \href{https://hf.co/google/gemma-3-4b-it}{hf.co/google/gemma-3-4b-it} \\
Qwe3-4B base & Apache 2.0 & \href{https://hf.co/Qwen/Qwen3.5-4B}{hf.co/Qwen/Qwen3.5-4B} \\
Qwen3.5-4B                    & Apache-2.0       & \href{https://hf.co/Qwen/Qwen3.5-4B}{hf.co/Qwen/Qwen3.5-4B} \\
\midrule
\multicolumn{3}{c}{\textit{Open-weight models — mid (7--35B)}} \\
\midrule
OLMo-3-7B-Think               & Apache-2.0       & \href{https://hf.co/allenai/Olmo-3-7B-Think}{hf.co/allenai/Olmo-3-7B-Think} \\
R1-0528-Qwen3-8B     & MIT              & \href{https://hf.co/deepseek-ai/DeepSeek-R1-0528-Qwen3-8B}{hf.co/deepseek-ai/DeepSeek-R1-0528-Qwen3-8B} \\
Ministral-3-8B & Apache 2.0 & \href{https://hf.co/mistralai/Ministral-3-8B-Reasoning-2512}{hf.co/mistralai/Ministral-3-8B-Reasoning-2512} \\
Qwen3.5-9B & Apache 2.0 & \href{https://hf.co/Qwen/Qwen3.5-9B}{hf.co/Qwen/Qwen3.5-9B} \\
Ministral-3-14B & Apache 2.0 & \href{https://hf.co/mistralai/Ministral-3-14B-Reasoning-2512}{hf.co/mistralai/Ministral-3-14B-Reasoning-2512} \\
gpt-oss-20b                   & Apache-2.0       & \href{https://hf.co/openai/gpt-oss-20b}{hf.co/openai/gpt-oss-20b} \\
Nemotron3-Nano-30B & nvidia-open-model-agreement & \href{https://hf.co/nvidia/Nemotron-3-Nano-Omni-30B-A3B-Reasoning-BF16}{hf.co/nvidia/Nemotron-3-Nano-Omni-30B-A3B-Reasoning-BF16} \\
Gemma-4-31B                   & Gemma License    & \href{https://hf.co/google/gemma-4-31B}{hf.co/google/gemma-4-31B} \\
Qwen3.5-35B-A3B               & Apache-2.0       & \href{https://hf.co/Qwen/Qwen3.5-35B-A3B}{hf.co/Qwen/Qwen3.5-35B-A3B} \\
\midrule
\multicolumn{3}{c}{\textit{Open-weight models — large (API-served)}} \\
\midrule
R1-Distill-LLama-70B & MIT & \href{https://hf.co/deepseek-ai/DeepSeek-R1-Distill-Llama-70B}{hf.co/deepseek-ai/DeepSeek-R1-Distill-Llama-70B} \\
gpt-oss-120b                  & Apache-2.0       & \href{https://hf.co/openai/gpt-oss-120b}{hf.co/openai/gpt-oss-120b} \\
Qwen3.5-122B-A10B & Apache 2.0 & \href{https://hf.co/Qwen/Qwen3.5-122B-A10B}{hf.co/Qwen/Qwen3.5-122B-A10B} \\
Qwen3-235B-A22B & Apache 2.0 & \href{https://hf.co/Qwen/Qwen3-235B-A22B}{hf.co/Qwen/Qwen3-235B-A22B} \\
DeepSeek-V3.2                 & MIT              & \href{https://hf.co/deepseek-ai/DeepSeek-V3.2}{hf.co/deepseek-ai/DeepSeek-V3.2} \\
Kimi-K2.5              & Modified MIT     & \href{https://hf.co/moonshotai/Kimi-K2.5}{hf.co/moonshotai/Kimi-K2.5} \\
\midrule
\multicolumn{3}{c}{\textit{Closed-source models (API only)}} \\
\midrule
Claude-Haiku-4.5                & Proprietary      & \href{https://www.anthropic.com/claude/haiku}{anthropic.com/claude/haiku} \\
Gemini-2.5-Flash                        & Proprietary      & \href{https://ai.google.dev/gemini-api/docs/models/gemini-2.5-flash}{ai.google.dev/gemini-api/docs/models/gemini-2.5-flash} \\
GPT-5.4                       & Proprietary      & \href{https://openai.com/index/introducing-gpt-5-4/}{openai.com/index/introducing-gpt-5-4} \\
\midrule
\multicolumn{3}{c}{\textit{Additional Models Used for Translation}} \\
\midrule
tartuNLP/Qwen2.5-3B-Instruct-hsb-dsb & CC-BY-4.0 & \href{https://hf.co/tartuNLP/Qwen2.5-3B-Instruct-hsb-dsb}{hf.co/tartuNLP/Qwen2.5-3B-Instruct-hsb-dsb} \\
BSC-LT/salamandraTA-7b-instruct & Apache-2.0 & \href{https://hf.co/BSC-LT/salamandraTA-7b-instruct}{hf.co/BSC-LT/salamandraTA-7b-instruct} \\
Sarvamai                      & Apache-2.0 & \href{https://hf.co/sarvamai/collections}{hf.co/sarvamai/collections} \\
DeepL                         & Proprietary      & \href{https://www.deepl.com/en/products/api}{https://www.deepl.com/en/products/api} \\
NLLB & CC-BY-NC-4.0 & \href{https://hf.co/facebook/nllb-200-distilled-600M}{hf.co/facebook/nllb-200-distilled-600M} \\

\bottomrule
\end{tabular}
\caption{Overview of the licenses and direct links of the resources utilized in this work.}
\label{tab:overview-license}
\end{table*}

\newpage

\section{Translations Annotator Instructions}
\label{app:annotator_instructions}

Below we reproduce, the written instructions that were shared with every language stakeholder at the start of the project.

\begin{tcolorbox}[
    enhanced, breakable,
    colback=gray!3, colframe=black!70,
    boxrule=0.5pt, arc=2pt,
    left=10pt, right=10pt, top=8pt, bottom=8pt,
    title={\textbf{Instructions for Language Stakeholders --- \pmath{} Project}},
    fonttitle=\bfseries\small,
    coltitle=white, colbacktitle=black!70,
    attach boxed title to top left={yshift=-2mm,xshift=4mm},
    boxed title style={colback=black!70, sharp corners, boxrule=0pt},
]

\textbf{Motivation.}\quad
We extend the \textbf{PolyMath} mathematical-reasoning benchmark
(\url{https://huggingface.co/datasets/Qwen/PolyMath}) beyond resource-rich
languages and evaluate the performance of current large language models
(LLMs) of various families and sizes on these languages. Language
stakeholders contribute to the project on a research basis and are
rewarded with co-authorship of the resulting paper.

\medskip
\textbf{Tasks for Language Stakeholders.}

\begin{enumerate}[leftmargin=1.4em, itemsep=4pt, topsep=4pt]

\item \textbf{Translation preparation.}
\begin{itemize}[leftmargin=1.2em, itemsep=2pt, topsep=2pt]
    \item Recommend the most suitable automatic translation system, so we
          can generate the first version of the benchmark.
          We are able to use paid versions of DeepL, Gemini, or
          Google Translate.
    \item Apart from proprietary models, or in case a very specific translation system is chosen, it is very helpful if the stakeholder can run the translation themselves.
\end{itemize}

\item \textbf{Translation check.}
\begin{itemize}[leftmargin=1.2em, itemsep=2pt, topsep=2pt]
    \item Only the question texts need to be checked; answers are typically
          numeric or LaTeX expressions and are universal.
    \item The check should be carried out by native speakers:
          \emph{at least one}, preferably two.
    \item Translation guidelines:
        \begin{itemize}[leftmargin=1.2em, itemsep=1pt, topsep=1pt]
            \item Text outside LaTeX code must be fully and correctly
                  translated, using the appropriate target-language
                  terminology.
            \item Text inside LaTeX code must remain in---or be reverted
                  to---the original English. Please use the helper
                  notebook to identify lines with an unbalanced number of
                  \texttt{\$} delimiters or otherwise broken LaTeX,
                  then fix them so that the LaTeX is identical to the
                  source.
        \end{itemize}
    \item \textbf{Please record the number of instances that required correction.}
\end{itemize}

\item \textbf{Reasoning proof-of-concept study.}
\begin{itemize}[leftmargin=1.2em, itemsep=2pt, topsep=2pt]
    \item Take only the first task of each difficulty level.
    \item Run it through several popular models, e.g.,
          ChatGPT (GPT-5), Gemini, Claude, and Qwen3-4B / Qwen3-30B-Thinking
          (the latter via the HuggingFace interface).
    \item Goal: assess whether the models can comprehend the reasoning
          task at all in the target language.
    \item Compile a short report of your findings.
\end{itemize}

\end{enumerate}
\end{tcolorbox}

\newpage

\section{\pmath Tasks Examples}
\label{app:pluramth_examples}
\begin{figure}[ht!]
    \centering
    \includegraphics[width=1\linewidth]{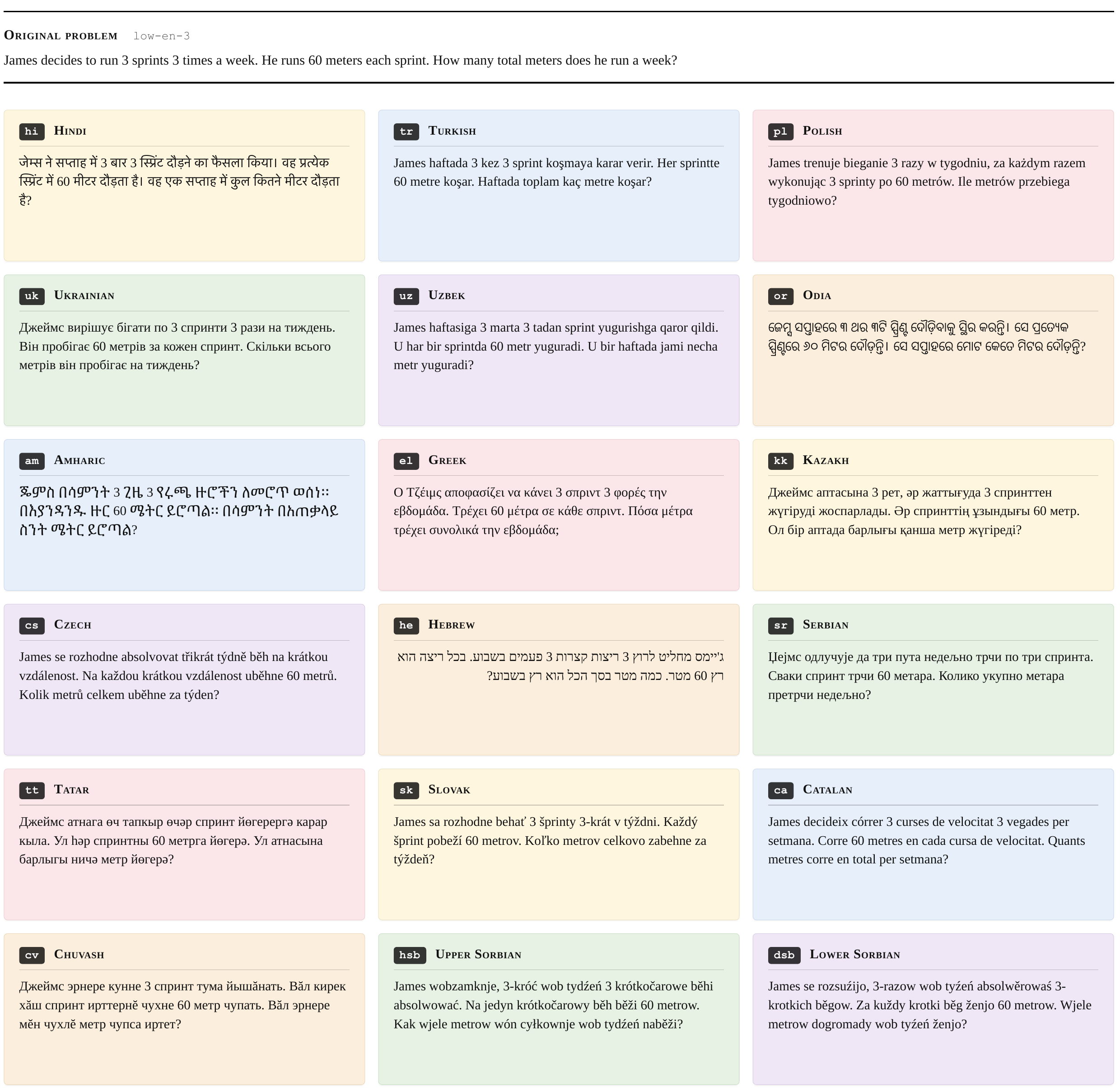}
    \caption{Examples of a translated math problem from \texttt{low} level to our \pmath languages.}
    \label{fig:pluramath_examples_low}
\end{figure}

\begin{figure}[ht!]
    \centering
    \includegraphics[width=1\linewidth]{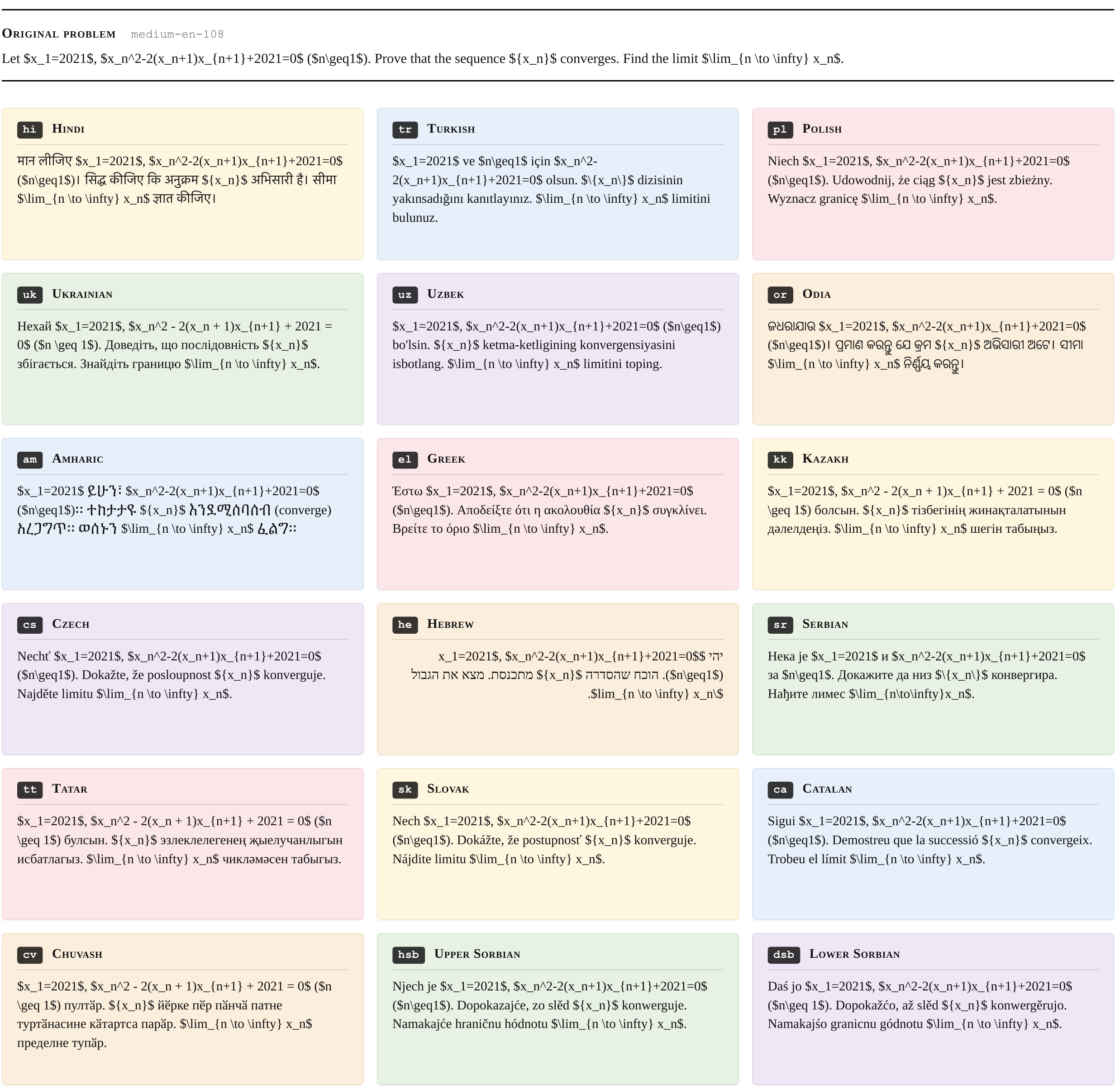}
    \caption{Examples of a translated math problem from \texttt{medium} level to our \pmath languages.}
    \label{fig:pluramath_examples_medium}
\end{figure}

\begin{figure}[ht!]
    \centering
    \includegraphics[width=1\linewidth]{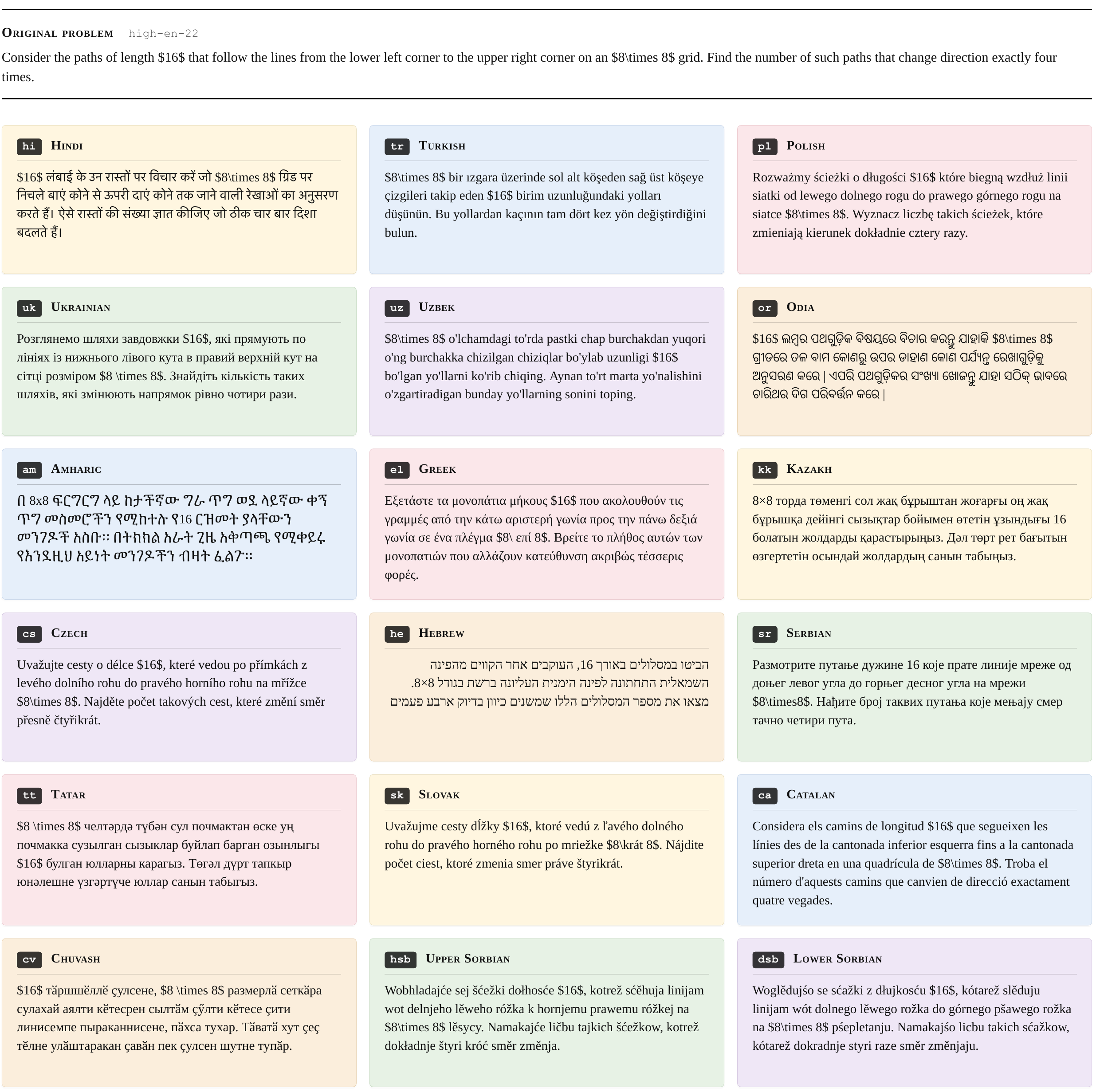}
    \caption{Examples of a translated math problem from \texttt{high} level to our \pmath languages.}
    \label{fig:pluramath_examples_high}
\end{figure}

\begin{figure}[ht!]
    \centering
    \includegraphics[width=1\linewidth]{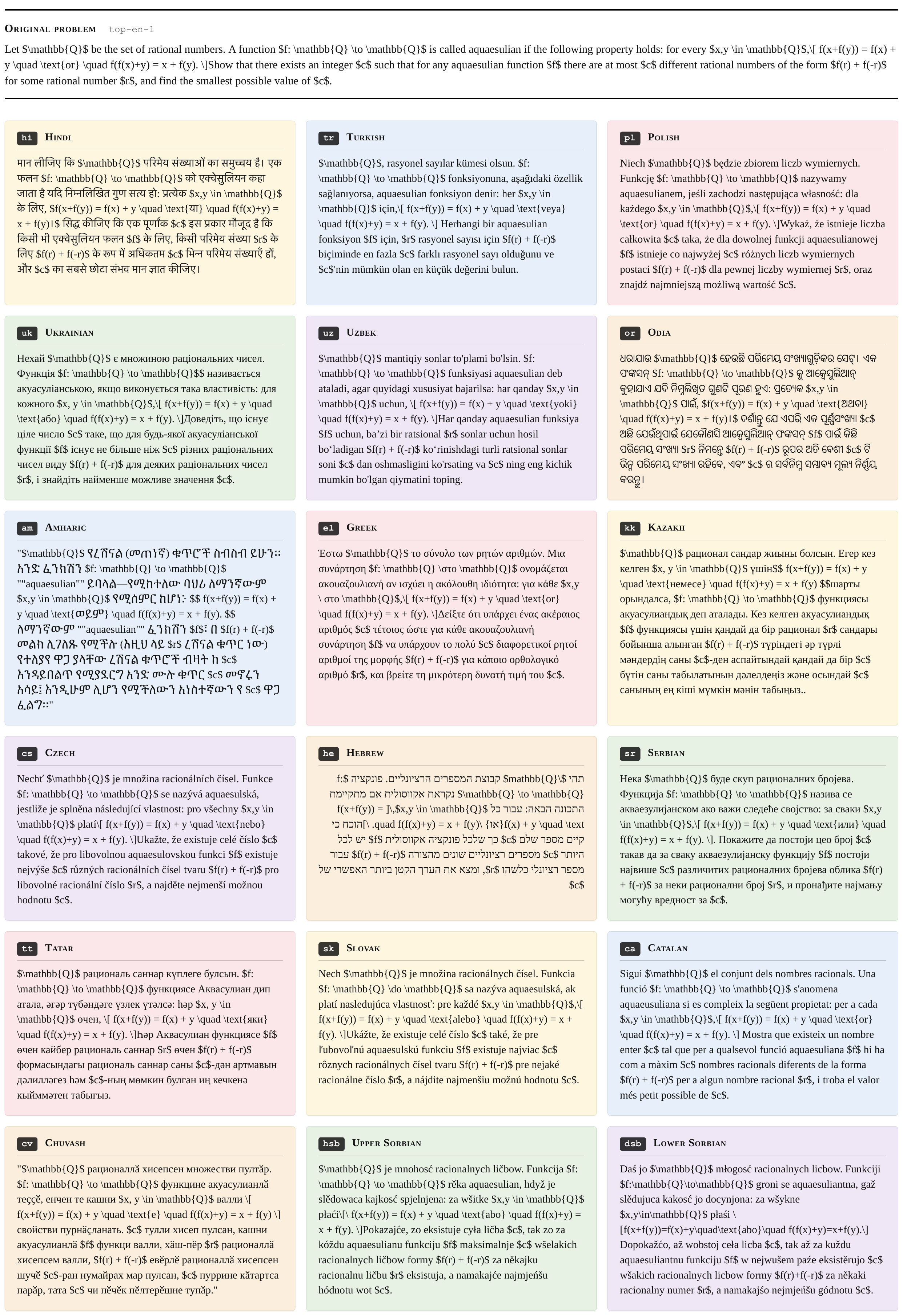}
    \caption{Examples of a translated math problem from \texttt{top} level to our \pmath languages.}
    \label{fig:pluramath_examples_top}
\end{figure}

We provide example instances from \pmath for each difficulty level: low~\ref{fig:pluramath_examples_low}, medium~\ref{fig:pluramath_examples_medium}, high~\ref{fig:pluramath_examples_high}, and top~\ref{fig:pluramath_examples_top}, alongside their corresponding original English tasks from PolyMath.

\clearpage
\newpage

\section{\pmath and PolyMath Math Tasks Length Comparison}
\label{app:data_length}

\begin{figure}[th!]
    \centering
    \includegraphics[width=0.9\linewidth]{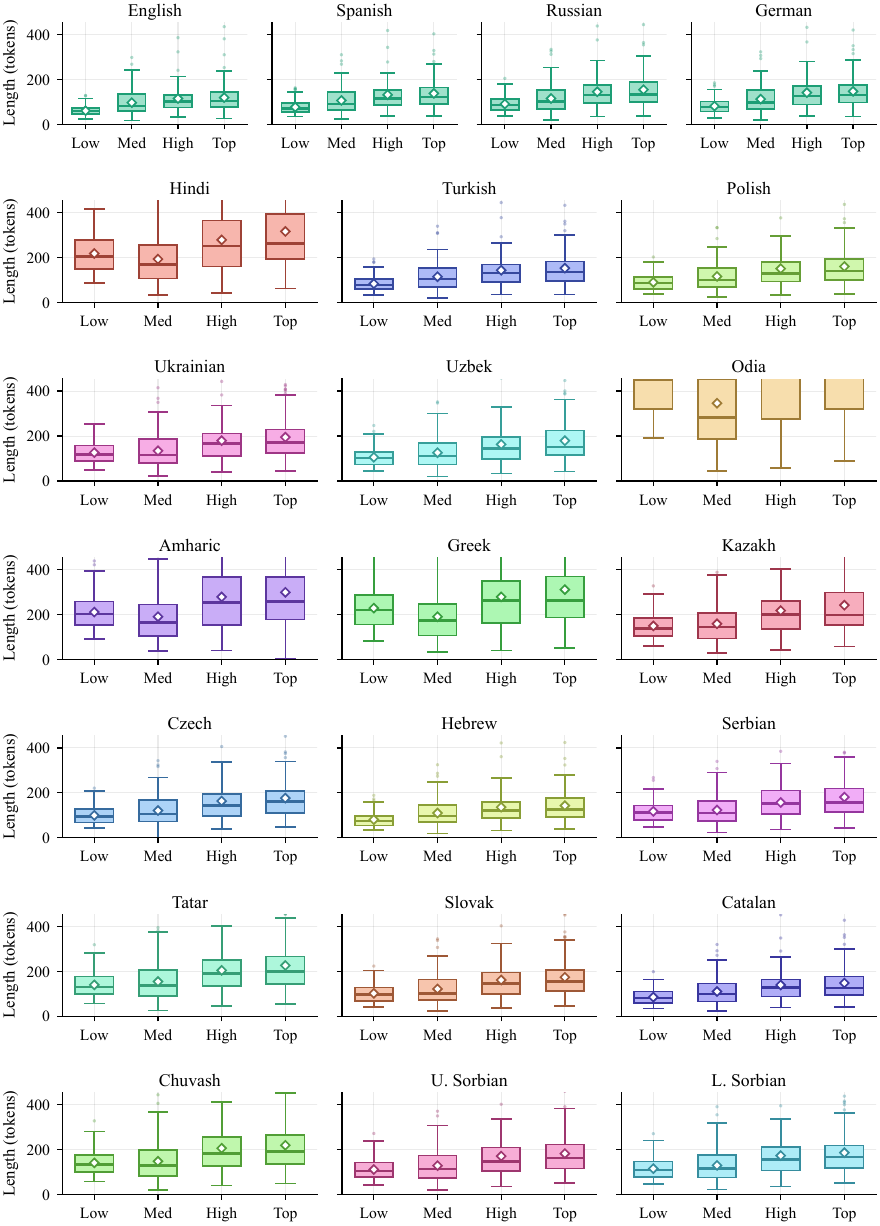}
    \caption{Fuul box-plot comparison of math task sample lengths between four high-resource languages from PolyMath and the 18 newly introduced languages in \pmath.}
    \label{fig:lengths_box_plot}
\end{figure}

We compare math task lengths between high-resource PolyMath languages—English, Spanish, Russian, and German—and the newly introduced languages in \pmath. Tokenization was performed with \texttt{Qwen3-4B} model. Mean and standard deviation statistics are reported in Tables~\ref{tab:lengths_hr},\ref{tab:lengths_ours_a},\ref{tab:lengths_ours_b},\ref{tab:lengths_ours_c}, with box plots shown in Figure~\ref{fig:lengths_box_plot}. Task lengths vary substantially across language families: Slavic languages exhibit relatively similar distributions, whereas Indo-Aryan languages such as Hindi and especially Odia, as well as the Semitic language Amharic and Turkic languages, have considerably longer inputs. These length differences may introduce additional challenges for task comprehension and long-context reasoning.

\begin{table}[th!]\centering\footnotesize
\begin{tabular}{lcccc}
\toprule
\textbf{Level} & \textbf{English} & \textbf{Spanish} & \textbf{Russian} & \textbf{German} \\
\midrule
Low & $60\,{\scriptstyle\pm22}$ & $76\,{\scriptstyle\pm29}$ & $90\,{\scriptstyle\pm34}$ & $81\,{\scriptstyle\pm32}$ \\
Medium & $96\,{\scriptstyle\pm53}$ & $106\,{\scriptstyle\pm56}$ & $115\,{\scriptstyle\pm63}$ & $112\,{\scriptstyle\pm61}$ \\
High & $114\,{\scriptstyle\pm78}$ & $131\,{\scriptstyle\pm83}$ & $144\,{\scriptstyle\pm89}$ & $140\,{\scriptstyle\pm86}$ \\
Top & $118\,{\scriptstyle\pm62}$ & $137\,{\scriptstyle\pm72}$ & $154\,{\scriptstyle\pm80}$ & $146\,{\scriptstyle\pm75}$ \\
\bottomrule
\end{tabular}
\caption{Length (tokens) per problem (mean\,$\pm$\,std) for the four high-resource languages from PolyMath, by difficulty level.}
\label{tab:lengths_hr}
\end{table}

\begin{table*}[h!]\centering\footnotesize
\begin{tabular}{lcccccc}
\toprule
\textbf{Level} & \textbf{Hindi} & \textbf{Turkish} & \textbf{Polish} & \textbf{Ukrainian} & \textbf{Uzbek} & \textbf{Odia} \\
\midrule
Low & $219\,{\scriptstyle\pm86}$ & $84\,{\scriptstyle\pm32}$ & $90\,{\scriptstyle\pm35}$ & $126\,{\scriptstyle\pm47}$ & $106\,{\scriptstyle\pm41}$ & $471\,{\scriptstyle\pm187}$ \\
Medium & $194\,{\scriptstyle\pm119}$ & $115\,{\scriptstyle\pm62}$ & $117\,{\scriptstyle\pm64}$ & $135\,{\scriptstyle\pm76}$ & $126\,{\scriptstyle\pm69}$ & $346\,{\scriptstyle\pm236}$ \\
High & $279\,{\scriptstyle\pm159}$ & $144\,{\scriptstyle\pm88}$ & $151\,{\scriptstyle\pm92}$ & $179\,{\scriptstyle\pm104}$ & $163\,{\scriptstyle\pm99}$ & $536\,{\scriptstyle\pm308}$ \\
Top & $317\,{\scriptstyle\pm181}$ & $153\,{\scriptstyle\pm78}$ & $161\,{\scriptstyle\pm82}$ & $195\,{\scriptstyle\pm103}$ & $180\,{\scriptstyle\pm91}$ & $604\,{\scriptstyle\pm385}$ \\
\bottomrule
\end{tabular}
\caption{Length (tokens) per problem (mean\,$\pm$\,std) for our \pmath target languages (1/3), by difficulty level.}
\label{tab:lengths_ours_a}
\end{table*}

\begin{table*}[h!]\centering\footnotesize
\begin{tabular}{lcccccc}
\toprule
\textbf{Level} & \textbf{Amharic} & \textbf{Greek} & \textbf{Kazakh} & \textbf{Czech} & \textbf{Hebrew} & \textbf{Serbian} \\
\midrule
Low & $211\,{\scriptstyle\pm76}$ & $228\,{\scriptstyle\pm93}$ & $149\,{\scriptstyle\pm56}$ & $100\,{\scriptstyle\pm39}$ & $81\,{\scriptstyle\pm32}$ & $118\,{\scriptstyle\pm45}$ \\
Medium & $190\,{\scriptstyle\pm115}$ & $190\,{\scriptstyle\pm116}$ & $158\,{\scriptstyle\pm88}$ & $121\,{\scriptstyle\pm68}$ & $110\,{\scriptstyle\pm59}$ & $123\,{\scriptstyle\pm67}$ \\
High & $278\,{\scriptstyle\pm161}$ & $278\,{\scriptstyle\pm161}$ & $217\,{\scriptstyle\pm120}$ & $163\,{\scriptstyle\pm97}$ & $136\,{\scriptstyle\pm86}$ & $157\,{\scriptstyle\pm69}$ \\
Top & $298\,{\scriptstyle\pm169}$ & $311\,{\scriptstyle\pm187}$ & $241\,{\scriptstyle\pm129}$ & $177\,{\scriptstyle\pm91}$ & $143\,{\scriptstyle\pm73}$ & $181\,{\scriptstyle\pm95}$ \\
\bottomrule
\end{tabular}
\caption{Length (tokens) per problem (mean\,$\pm$\,std) for our \pmath target languages (2/3), by difficulty level.}
\label{tab:lengths_ours_b}
\end{table*}

\begin{table*}[h!]\centering\footnotesize
\begin{tabular}{lcccccc}
\toprule
\textbf{Level} & \textbf{Tatar} & \textbf{Slovak} & \textbf{Catalan} & \textbf{Chuvash} & \textbf{U. Sorbian} & \textbf{L. Sorbian} \\
\midrule
Low & $139\,{\scriptstyle\pm53}$ & $102\,{\scriptstyle\pm41}$ & $85\,{\scriptstyle\pm33}$ & $141\,{\scriptstyle\pm54}$ & $112\,{\scriptstyle\pm45}$ & $115\,{\scriptstyle\pm47}$ \\
Medium & $154\,{\scriptstyle\pm86}$ & $121\,{\scriptstyle\pm68}$ & $110\,{\scriptstyle\pm59}$ & $148\,{\scriptstyle\pm85}$ & $129\,{\scriptstyle\pm71}$ & $130\,{\scriptstyle\pm73}$ \\
High & $204\,{\scriptstyle\pm114}$ & $161\,{\scriptstyle\pm96}$ & $139\,{\scriptstyle\pm88}$ & $206\,{\scriptstyle\pm119}$ & $171\,{\scriptstyle\pm100}$ & $173\,{\scriptstyle\pm101}$ \\
Top & $226\,{\scriptstyle\pm119}$ & $173\,{\scriptstyle\pm91}$ & $148\,{\scriptstyle\pm77}$ & $219\,{\scriptstyle\pm118}$ & $182\,{\scriptstyle\pm95}$ & $187\,{\scriptstyle\pm100}$ \\
\bottomrule
\end{tabular}
\caption{Length (tokens) per problem (mean\,$\pm$\,std) for our \pmath target languages (3/3), by difficulty level.}
\label{tab:lengths_ours_c}
\end{table*}

\clearpage
\newpage

\section{Prompt Templates and Examples}
\label{sec:app_prompts}

This section gives the full text of all prompts used in the experiments together with a specific illustrative examples in Ukrainian. Throughout, ``HR'' denotes the high-resource pivot language (English by default; for non-English HR baselines such as German, Russian, or Spanish we substitute the language name in the directive).

\subsection{Templates}
\label{sec:app_prompt_templates}

\paragraph{1. Base} Problem and instruction in the target language. The model is free to reason in any language but is asked to place the final answer in \verb|$\boxed{}$|.
\begin{promptboxtitled}{Base}
\textnormal{\{problem\textsubscript{tgt}\}}\\[2pt]
\textnormal{\{closing\_instruction\textsubscript{tgt}\}}
\end{promptboxtitled}

\paragraph{2. Base+EN-CoT} Problem in the target language; an additional system-level directive instructs the model to perform its chain-of-thought in English while still placing the final answer in a \verb|$\boxed{}$|. 

\begin{promptboxtitled}{Base+EN-CoT}
\textnormal{\textbf{[System]} You are solving a mathematical problem. Reason step by step in English, then write the final answer in \$$\backslash$boxed\{\}\$.}\\[3pt]
\textnormal{\textbf{[User]} \{problem\textsubscript{tgt}\}}\\[2pt]
\end{promptboxtitled}

\paragraph{3. Backtranslated} The problem is first machine-translated from the target language back to the high-resource language using X model, then presented to the model entirely in the high-resource language.

\begin{promptboxtitled}{Backtranslated}
\textnormal{\{problem\textsubscript{HR}\}}\\[2pt]
\textnormal{\{closing\_instruction\textsubscript{HR}\}}
\end{promptboxtitled}



\subsection{Prompts Example in Ukrainian}
\label{sec:app_prompt_example}

We illustrate every prompt template on a concrete problem from the low-difficulty level in Ukrainian.


\begin{promptboxtitled}{Native (Ukrainian)}
\foreignlanguage{ukrainian}{Качки Дженет несуть по 16 яєць на день. Вона їсть три на сніданок щоранку і пече кекси для своїх друзів щодня, використовуючи чотири. Решту вона щодня продає на фермерському ринку по 2~\$ за свіже качине яйце. Скільки в доларах вона заробляє на фермерському ринку щодня?}\\[2pt]
\foreignlanguage{ukrainian}{Примітка: Будь ласка, вставте остаточну відповідь в \$$\backslash$boxed\{\}\$.}
\end{promptboxtitled}

\begin{promptboxtitled}{\textit{English reference for illustration purpose only}}
Janet's ducks lay 16 eggs per day. She eats three for breakfast every morning and bakes muffins for her friends every day with four. She sells the remainder at the farmers' market daily for \$2 per fresh duck egg. How much in dollars does she make every day at the farmers' market?\\[2pt]
Note: Please put the final answer in \$$\backslash$boxed\{\}\$.
\end{promptboxtitled}


\begin{promptboxtitled}{Native+EN-CoT (Ukrainian problem\, English CoT)}
\textbf{[System]} You are solving a mathematical problem. Reason step by step in English, then write the final answer in $\backslash$boxed\{\}.\\[3pt]
\textbf{[User]} \foreignlanguage{ukrainian}{Качки Дженет несуть по 16 яєць на день. Вона їсть три на сніданок щоранку і пече кекси для своїх друзів щодня, використовуючи чотири. Решту вона щодня продає на фермерському ринку по 2~\$ за свіже качине яйце. Скільки в доларах вона заробляє на фермерському ринку щодня?}\\
\end{promptboxtitled}


\begin{promptboxtitled}{Backtranslated (Ukrainian $\rightarrow$ English via MT)}
Janet's ducks lay 16 eggs per day. She eats three for breakfast every morning and bakes muffins for her friends every day with four. She sells the remainder at the farmers' market daily for \$2 per fresh duck egg. How much in dollars does she make every day at the farmers' market?\\[2pt]
Note: Please put the final answer in $\backslash$boxed\{\}.
\end{promptboxtitled}



\subsection{Per-Language Closing Instructions}
\label{sec:app_boxing}

\begin{figure}[ht!]
    \centering
    \includegraphics[width=1\linewidth]{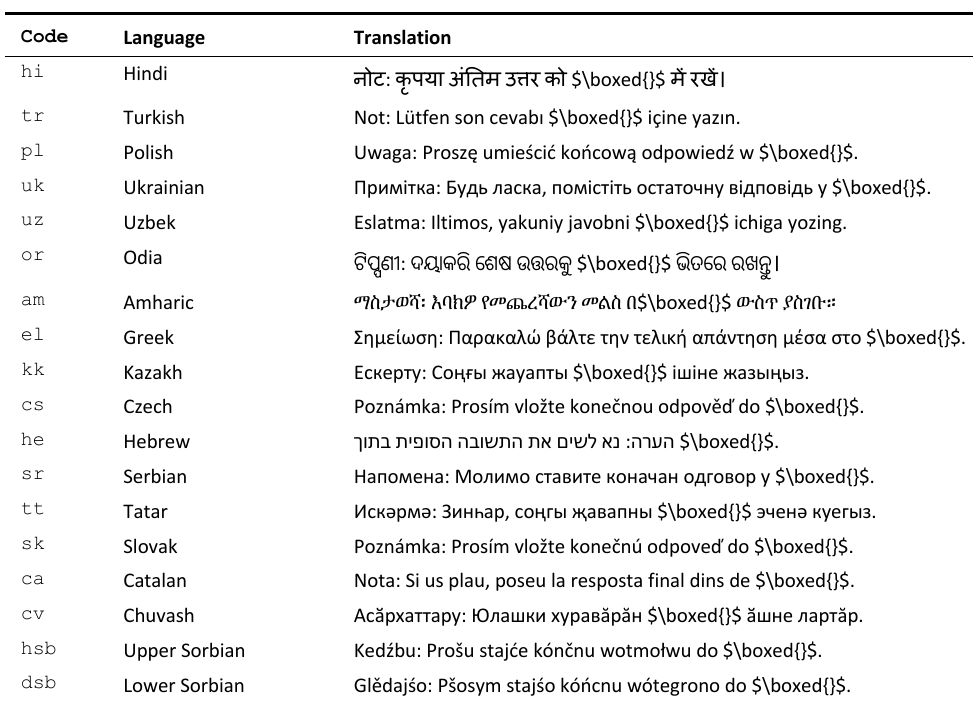}
    \caption{The translated per language final instructions which were added in the end of main tasks. English original: \textit{Note: Please put the final answer in the \$$\backslash$boxed\{\}\$.}}
    \label{fig:prompt_engings}
\end{figure}

To keep the input fully monolingual under in \textbf{Base} prompt, we translated the closing instructions recommended in the original PolyMat~\cite{wang2025polymath} to keep the final output structured. Native speakers verified each translation. We list all 18 target-language directives in Table~\ref{fig:prompt_engings}.

\clearpage
\newpage


\section{Hyperparameter Search}
\label{app:hp-search}

We performed a two-stage hyperparameter search before running the main
multilingual evaluation. All scores in this appendix are reported with the
metrics described in Section~\ref{sec:evaluation_metric}: per-level scores
are exact-match accuracies of the content of the \verb|$\boxed{}$| expression,
and the rightmost column of each table reports PolyMath's
difficulty-weighted accuracy (DW-ACC) aggregated over all four difficulty
levels with weights $\{1, 2, 4, 8\}$.

\paragraph{Stage 1 -- choosing the reasoning effort (English only).}
We first searched for an appropriate \emph{reasoning effort} setting using
only the English split of PolyMath. For each of the three reasoning models
considered, we ran the supported reasoning-effort levels (\emph{low},
\emph{medium}, \emph{high}) at a fixed sampling temperature.
Results are reported in Table~\ref{tab:hp-english-reasoning}.

\paragraph{Stage 2 -- temperature sweep on high-resource languages.}
Using the reasoning-effort settings selected in Stage~1, we then ran a
narrower temperature sweep on three high-resource languages (English,
German, Russian). The resulting configurations and per-language scores are
shown in Table~\ref{tab:hp-multilingual}. The \emph{low}+temperature$=0.5$
configuration was run on English only as a sanity check on temperature
sensitivity and was not extended to the other two languages.

\begin{table}[h!]
\centering
\small
\begin{tabular}{llrrrrr}
\toprule
\textbf{Reasoning effort} & \textbf{Model} & \textbf{Low} & \textbf{Medium} & \textbf{High} & \textbf{Top} & \textbf{DW-ACC} \\
\midrule
\multirow{3}{*}{Low} & \texttt{gpt-oss-120b} & 79.2 & 39.2 & 28.0 & 13.6 & 25.2 \\
 & \texttt{Qwen3-235B} & 93.6 & 4.0 & 0.0 & 0.0 & 6.8 \\
 & \texttt{Qwen3-30B} & 92.8 & 7.2 & 3.2 & 0.8 & 8.4 \\
\midrule
\multirow{3}{*}{Medium} & \texttt{gpt-oss-120b} & 77.6 & 44.8 & 31.2 & 9.6 & 24.6 \\
 & \texttt{Qwen3-235B} & 92.8 & 3.2 & 0.0 & 0.0 & 6.6 \\
 & \texttt{Qwen3-30B} & 93.6 & 7.2 & 3.2 & 0.8 & 8.5 \\
\midrule
\multirow{3}{*}{High} & \texttt{gpt-oss-120b} & 83.2 & 32.0 & 13.6 & 4.0 & 15.6 \\
 & \texttt{Qwen3-235B} & 93.6 & 4.8 & 0.0 & 0.0 & 6.9 \\
 & \texttt{Qwen3-30B} & 89.6 & 7.2 & 2.4 & 0.8 & 8.0 \\
\bottomrule
\end{tabular}
\caption{Effect of reasoning effort on English-only PolyMath performance.
Per-level scores are exact-match accuracies (\%) of the content of
\texttt{\textbackslash boxed\{\}}; the DW-ACC column reports the
difficulty-weighted accuracy aggregated over all four levels with
weights $\{1,2,4,8\}$.}
\label{tab:hp-english-reasoning}
\end{table}

\begin{table*}[h!]
\centering
\scriptsize
\setlength{\tabcolsep}{4pt}
\begin{tabular}{ll l rrrrr rrrrr rrrrr}
\toprule
& & & \multicolumn{5}{c}{\textbf{English}} & \multicolumn{5}{c}{\textbf{German}} & \multicolumn{5}{c}{\textbf{Russian}} \\
\cmidrule(lr){4-8} \cmidrule(lr){9-13} \cmidrule(lr){14-18}
\textbf{Effort} & \textbf{Temp.} & \textbf{Model}
& Low & Med & High & Top & DW
& Low & Med & High & Top & DW
& Low & Med & High & Top & DW \\
\midrule
\multirow{3}{*}{Low} & \multirow{3}{*}{0.1} & \texttt{gpt-oss-120b} & 79.2 & 39.2 & 28.0 & 13.6 & 25.2 & 61.6 & 32.8 & 29.6 & 11.2 & 22.3 & 66.4 & 36.8 & 27.2 & 12.8 & 23.4 \\
 &  & \texttt{Qwen3-235B} & 93.6 & 4.0 & 0.0 & 0.0 & 6.8 & 87.2 & 9.6 & 1.6 & 0.0 & 7.5 & 91.2 & 11.2 & 4.0 & 0.0 & 8.6 \\
 &  & \texttt{Qwen3-30B} & 92.8 & 7.2 & 3.2 & 0.8 & 8.4 & 87.2 & 11.2 & 4.0 & 0.8 & 8.8 & 89.6 & 8.8 & 4.8 & 0.8 & 8.9 \\
\midrule
\multirow{3}{*}{Low} & \multirow{3}{*}{0.3} & \texttt{gpt-oss-120b} & 79.2 & 40.0 & 27.2 & 8.8 & 22.6 & 63.2 & 36.8 & 31.2 & 8.8 & 22.1 & 69.6 & 41.6 & 28.8 & 16.0 & 26.4 \\
 &  & \texttt{Qwen3-235B} & 92.8 & 3.2 & 0.0 & 0.0 & 6.6 & 87.2 & 8.8 & 1.6 & 0.0 & 7.4 & 92.0 & 11.2 & 4.8 & 0.0 & 8.9 \\
 &  & \texttt{Qwen3-30B} & 91.2 & 7.2 & 1.6 & 0.8 & 7.9 & 87.2 & 11.2 & 4.0 & 0.8 & 8.8 & 88.0 & 12.0 & 4.0 & 0.8 & 9.0 \\
\midrule
\multirow{3}{*}{Low} & \multirow{3}{*}{0.5} & \texttt{gpt-oss-120b} & 80.0 & 38.4 & 28.8 & 10.4 & 23.7 & -- & -- & -- & -- & -- & -- & -- & -- & -- & -- \\
 &  & \texttt{Qwen3-235B} & 93.6 & 3.2 & 0.0 & 0.0 & 6.7 & -- & -- & -- & -- & -- & -- & -- & -- & -- & -- \\
 &  & \texttt{Qwen3-30B} & 93.6 & 5.6 & 2.4 & 0.8 & 8.1 & -- & -- & -- & -- & -- & -- & -- & -- & -- & -- \\
\midrule
\multirow{3}{*}{Medium} & \multirow{3}{*}{0.1} & \texttt{gpt-oss-120b} & 77.6 & 44.8 & 31.2 & 9.6 & 24.6 & 62.4 & 43.2 & 32.0 & 8.0 & 22.7 & 63.2 & 41.6 & 28.0 & 8.8 & 21.9 \\
 &  & \texttt{Qwen3-235B} & 92.8 & 3.2 & 0.0 & 0.0 & 6.6 & 88.0 & 9.6 & 1.6 & 0.0 & 7.6 & 92.8 & 10.4 & 4.0 & 0.0 & 8.6 \\
 &  & \texttt{Qwen3-30B} & 93.6 & 7.2 & 3.2 & 0.8 & 8.5 & 87.2 & 10.4 & 5.6 & 0.8 & 9.1 & 88.8 & 10.4 & 3.2 & 0.8 & 8.6 \\
\midrule
\multirow{3}{*}{Medium} & \multirow{3}{*}{0.3} & \texttt{gpt-oss-120b} & 79.2 & 41.6 & 28.8 & 9.6 & 23.6 & 65.6 & 43.2 & 28.8 & 9.6 & 22.9 & 58.4 & 42.4 & 27.2 & 8.8 & 21.5 \\
 &  & \texttt{Qwen3-235B} & 93.6 & 3.2 & 0.0 & 0.0 & 6.7 & 85.6 & 7.2 & 0.8 & 0.0 & 6.9 & 91.2 & 8.8 & 4.8 & 0.0 & 8.5 \\
 &  & \texttt{Qwen3-30B} & 91.2 & 6.4 & 3.2 & 0.8 & 8.2 & 86.4 & 12.0 & 4.0 & 0.8 & 8.9 & 89.6 & 12.8 & 4.0 & 0.8 & 9.2 \\
\bottomrule
\end{tabular}
\caption{Hyperparameter search on high-resource languages
(English, German, Russian) across all four PolyMath difficulty levels.
Each configuration combines a reasoning-effort setting (\emph{low} or
\emph{medium}) with a sampling temperature.
Per-level scores are exact-match accuracies (\%); the DW column reports the
difficulty-weighted accuracy aggregated over all four levels with weights
$\{1,2,4,8\}$ (denominator~$15$).
A dash (--) marks configurations not run for the given language.}
\label{tab:hp-multilingual}
\end{table*}

\clearpage
\newpage

\section{Closed Models Inference Budget}
\label{app:budget}

First, we utilized commercial closed-source APIs for precomputing translations. For 10 \pmath languages, we used the DeepL API at a total cost of ($\$63.82$), while Gemini API usage across three models amounted to approximately ($\$11.60$).

We estimate the cost of the benchmark from output-token pricing, which
dominates spend for reasoning models. The dataset contains
$L = 4$ difficulty levels with $T = 125$ tasks each, giving
$N = L \cdot T = 500$ problems. We cap every reasoning trace at
$B = 2{,}000$ output tokens, so a single model produces
$N \cdot B = 5\!\times\!10^{2} \cdot 2\!\times\!10^{3} = 10^{6}$
output tokens per language, i.e.\ exactly one million tokens. The
benchmark covers $\ell = 22$ languages and $p = 3$ prompt designs.

Let $c_m$ be the price (USD) per one million output tokens for model
$m$. Because each model emits exactly $10^{6}$ tokens per language, the
cost of evaluating $m$ under a \emph{single} prompt design across all
languages reduces to
\[
  C^{1}_{m} \;=\; c_m \cdot \frac{N \cdot B \cdot \ell}{10^{6}}
            \;=\; c_m \cdot \ell \;=\; 22\,c_m,
\]
and a \emph{full} three-design sweep costs
$C^{3}_{m} = p \cdot \ell \cdot c_m = 66\,c_m$.

\paragraph{Evaluation protocol.}
To keep the budget tractable we run the full three-design sweep only for
the open-weight mid-sized models and the two cheapest large open-weight
models, where the marginal cost is negligible. The remaining large
models---including all proprietary endpoints---are evaluated under the
single base prompt design. Table~\ref{tab:budget} reports the per-model
breakdown. Under this protocol the adopted budget is
$\$236.94$ (full thee-prompts design group for open-source models) $+\;\$724.90$ (base-only group for bigger and closed models)
$=\;\mathbf{\$961.84}$. 

\begin{table*}[h!]
\centering
\small
\begin{tabular}{llcrr}
\toprule
\textbf{Model} & \textbf{Access} & \textbf{\$/1M out} & \textbf{1 design ($\times 22$)} & \textbf{3 designs ($\times 66$)} \\
\midrule
\multicolumn{5}{l}{\textit{Full three-prompts design}} \\
\addlinespace[1pt]
GPT-OSS-20B            & Open & 0.14 & 3.08  & 9.24  \\
Gemma-4-31B-it         & Open & 0.38 & 8.36  & 25.08 \\
Qwen3.5-35B-A3B        & Open & 1.00 & 22.00 & 66.00 \\
DeepSeek-R1-Distill-70B & Open & 0.80 & 17.60 & 52.80 \\
Nemotron-3-Nano-30B-A3B & Open & 0.80 & 17.60 & 52.80 \\
GPT-OSS-120B           & Open & 0.19 & 4.18  & 12.54 \\
DeepSeek-V3.2          & Open & 0.28 & 6.16  & 18.48 \\
\addlinespace[1pt]
\multicolumn{4}{l}{\textit{Subtotal (full sweep)}} & \textbf{236.94} \\
\midrule
\multicolumn{5}{l}{\textit{Base prompt design only}} \\
\addlinespace[1pt]
Qwen3.5-122B-A10B      & Open   & 2.40  & 52.80  & --- \\
Kimi-K2.5              & Open   & 2.25  & 49.50  & --- \\
Qwen3-235B-A22B-Thinking & Open & 2.30  & 50.60  & --- \\
Claude-Haiku-4.5        & Closed & 16.00 & 352.00 & --- \\
Gemini-2.5-Flash       & Closed & 2.50  & 55.00  & --- \\
GPT-5.4$^{\dagger}$    & Closed & 15.00 & 165.00 & --- \\
\addlinespace[1pt]
\multicolumn{4}{l}{\textit{Subtotal (base only)}} & \textbf{724.90} \\
\midrule
\multicolumn{4}{l}{\textbf{Adopted budget}} & \textbf{961.84} \\
\bottomrule
\end{tabular}
\caption{Estimated inference budget (USD). Costs are derived from
output-token pricing: each model emits $N\!\cdot\!B = 10^{6}$ tokens per
language ($N=500$ problems, $B=2{,}000$ tokens per trace), so the cost of
one prompt design across $\ell=22$ languages is $22\,c_m$ and a full
three-design sweep is $66\,c_m$. Large and proprietary models are
evaluated under a single base prompt design only.
$^{\dagger}$Batched decoding halves the GPT-5.4 line item from \$330.00 to \$165.00.}
\label{tab:budget}
\end{table*}

\clearpage
\newpage

\section{Detailed Per-Difficulty-Level and Per-Prompting-Ablations Results}
\label{app:results_ablations}

Here, we present details results per levels and prompts designs as well as additional results visualizations.

Per difficulty levels results: Appendix~\ref{app:results_per_level}, models' reasoning length comparison: Appendix~\ref{app:models_reasoning_length}, per prompting types results: Appendix~\ref{app:results_ablation_prompts}.

\subsection{Per-Difficulty-Level Results}
\label{app:results_per_level}

\begin{figure}[h!]
    \centering
    \includegraphics[width=\linewidth]{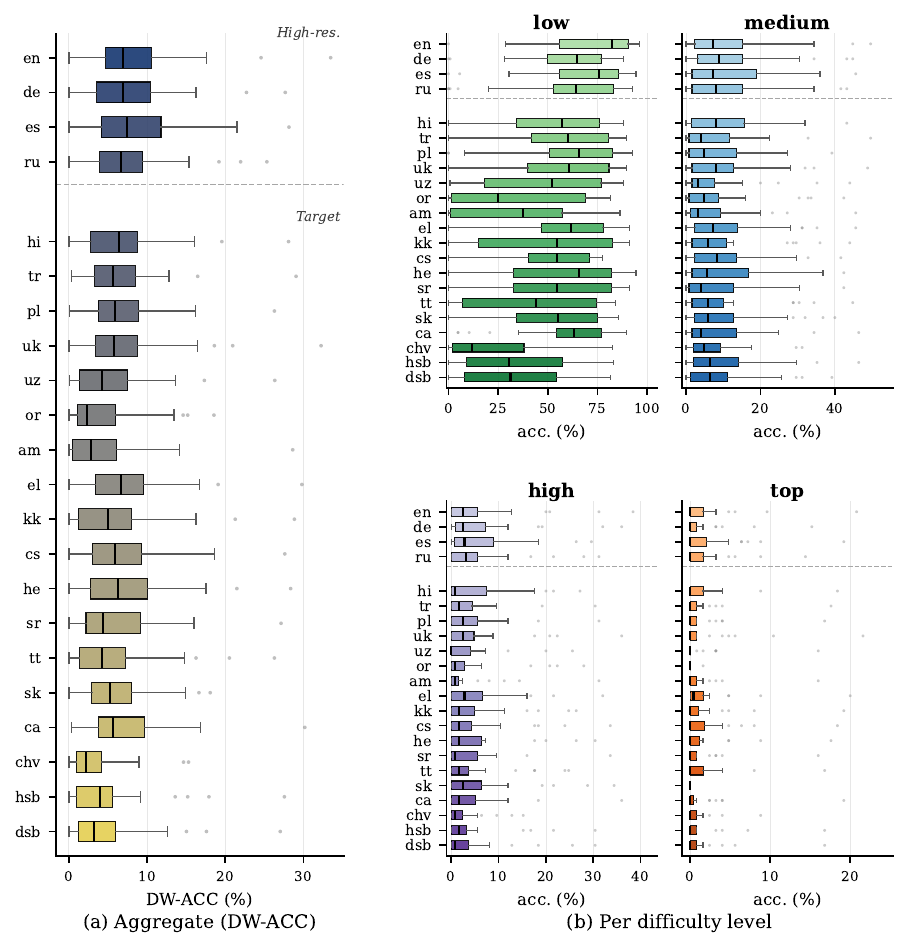}
    \caption{Aggregated \texttt{DW=ACC} \texttt{base} prompting scores distributions per all languages across all models and all levels.We observe a clear trend in mean scores that correlates with language resource rankings, with severely underrepresented languages exhibiting low performance even on low-difficulty tasks.}
    \label{fig:lang_scores_distributions}
\end{figure}

\begin{figure}[h!]
    \centering
    \includegraphics[width=\linewidth]{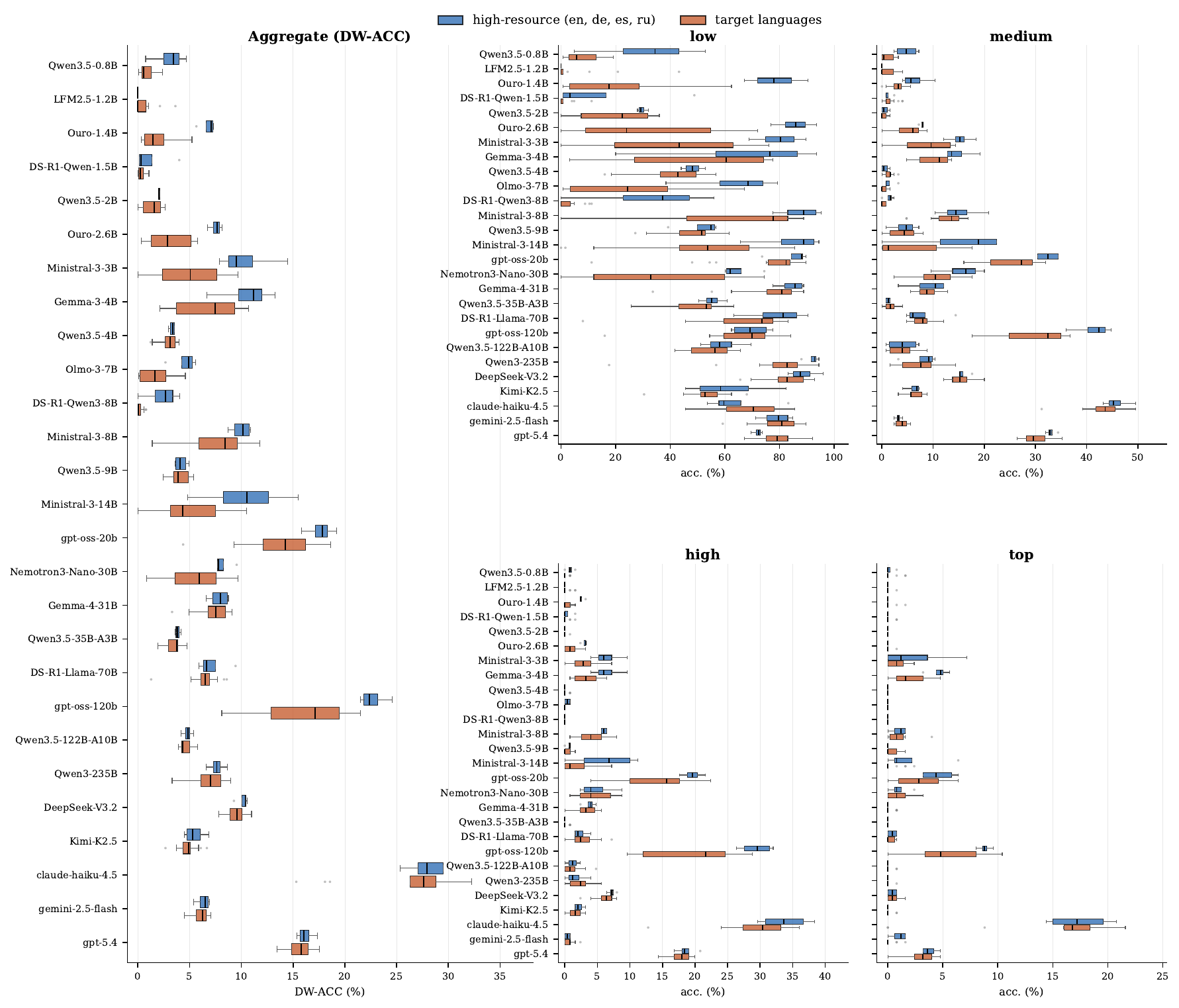}
    \caption{Answers correctness scores distributions for \texttt{base} prompting per languages types across all models per each level. Smaller models exhibit substantial performance degradation on underrepresented languages, whereas larger and proprietary models remain comparatively stable across language groups.}
    \label{fig:models_stability_distributions}
\end{figure}

Results for each difficulty level in terms of math answer correctness: low (Table~\ref{tab:base-low}), medium (Table~\ref{tab:base-medium}), high (Table~\ref{tab:base-high}), and top (Table~\ref{tab:base-top}). Additionally, we put here box-plots results comparison per language (Figure~\ref{fig:lang_scores_distributions}) and per model (Figure~\ref{fig:models_stability_distributions}). 

Overall, most models achieve relatively strong performance on low-level tasks, followed by a substantial decline on higher difficulty levels. Even at the lowest level, Odia, Amharic, and Chuvash remain the most challenging underrepresented languages. For higher difficulty levels, only the \texttt{gpt-oss} variants and the proprietary \texttt{GPT-5.4} and \texttt{Claude-Haiku-4.5} models maintain consistently non-zero performance.

\begin{sidewaystable*}[p]
\centering\scriptsize
\setlength{\tabcolsep}{1.6pt}\renewcommand{\arraystretch}{1.15}
\begin{tabular}{l|cccc|ccc|cccccccccccccccccc|ccc}
\toprule
 & \multicolumn{4}{c|}{\textbf{High-resource}} & \multicolumn{3}{c|}{\textbf{Summary (HR)}} & \multicolumn{18}{c|}{\textbf{Our target languages}} & \multicolumn{3}{c}{\textbf{Summary (target)}} \\
\cmidrule(lr){2-5}\cmidrule(lr){6-8}\cmidrule(lr){9-26}\cmidrule(lr){27-29}
\textbf{Model} & \rotatebox{70}{\texttt{en}} & \rotatebox{70}{\texttt{de}} & \rotatebox{70}{\texttt{ru}} & \rotatebox{70}{\texttt{es}} & \rotatebox{70}{\textbf{Avg.}} & \rotatebox{70}{Len} & \rotatebox{70}{Lang} & \rotatebox{70}{\texttt{hi}} & \rotatebox{70}{\texttt{tr}} & \rotatebox{70}{\texttt{pl}} & \rotatebox{70}{\texttt{uk}} & \rotatebox{70}{\texttt{uz}} & \rotatebox{70}{\texttt{or}} & \rotatebox{70}{\texttt{am}} & \rotatebox{70}{\texttt{el}} & \rotatebox{70}{\texttt{kk}} & \rotatebox{70}{\texttt{cs}} & \rotatebox{70}{\texttt{he}} & \rotatebox{70}{\texttt{sr}} & \rotatebox{70}{\texttt{tt}} & \rotatebox{70}{\texttt{sk}} & \rotatebox{70}{\texttt{ca}} & \rotatebox{70}{\texttt{cv}} & \rotatebox{70}{\texttt{hsb}} & \rotatebox{70}{\texttt{dsb}} & \rotatebox{70}{\textbf{Avg.}} & \rotatebox{70}{Len} & \rotatebox{70}{Lang} \\
\midrule
\multicolumn{29}{c}{\textit{Open-weight --- small ($\leq$4B)}}\\
\bottomrule
\end{tabular}
\caption{\textbf{Low-difficulty} base-prompting results on our \pmath languages vs high-resource ones from PolyMath. Each cell shows answer accuracy (\%; large) over the answer-format compliance rate (\%; small). Cell shading encodes accuracy on a tier-specific scale (pale\,$\rightarrow$\,deep teal, $0\rightarrow100\%$); a numeral is printed in \textbf{black} on light (low-score) cells and in \textbf{white} on dark (high-score) cells solely for legibility---the text colour carries no extra meaning. \underline{\textbf{Best}} and \textit{second-best} per column are highlighted. After each language block we report the macro-average accuracy, the mean$\pm$std generation length (tokens, reasoning+answer), and the dominant answer language coded \textsc{EN} (predominantly English) or \textsc{TL} (requested target language).}
\label{tab:base-low}
\end{sidewaystable*}

\begin{sidewaystable*}[p]
\centering\scriptsize
\setlength{\tabcolsep}{1.6pt}\renewcommand{\arraystretch}{1.15}
\begin{tabular}{l|cccc|ccc|cccccccccccccccccc|ccc}
\toprule
 & \multicolumn{4}{c|}{\textbf{High-resource}} & \multicolumn{3}{c|}{\textbf{Summary (HR)}} & \multicolumn{18}{c|}{\textbf{Our target languages}} & \multicolumn{3}{c}{\textbf{Summary (target)}} \\
\cmidrule(lr){2-5}\cmidrule(lr){6-8}\cmidrule(lr){9-26}\cmidrule(lr){27-29}
\textbf{Model} & \rotatebox{70}{\texttt{en}} & \rotatebox{70}{\texttt{de}} & \rotatebox{70}{\texttt{ru}} & \rotatebox{70}{\texttt{es}} & \rotatebox{70}{\textbf{Avg.}} & \rotatebox{70}{Len} & \rotatebox{70}{Lang} & \rotatebox{70}{\texttt{hi}} & \rotatebox{70}{\texttt{tr}} & \rotatebox{70}{\texttt{pl}} & \rotatebox{70}{\texttt{uk}} & \rotatebox{70}{\texttt{uz}} & \rotatebox{70}{\texttt{or}} & \rotatebox{70}{\texttt{am}} & \rotatebox{70}{\texttt{el}} & \rotatebox{70}{\texttt{kk}} & \rotatebox{70}{\texttt{cs}} & \rotatebox{70}{\texttt{he}} & \rotatebox{70}{\texttt{sr}} & \rotatebox{70}{\texttt{tt}} & \rotatebox{70}{\texttt{sk}} & \rotatebox{70}{\texttt{ca}} & \rotatebox{70}{\texttt{cv}} & \rotatebox{70}{\texttt{hsb}} & \rotatebox{70}{\texttt{dsb}} & \rotatebox{70}{\textbf{Avg.}} & \rotatebox{70}{Len} & \rotatebox{70}{Lang} \\
\midrule
\multicolumn{29}{c}{\textit{Open-weight --- small ($\leq$4B)}}\\
\bottomrule
\end{tabular}
\caption{\textbf{Medium-difficulty} base-prompting results on our \pmath languages vs high-resource ones from PolyMath. Each cell shows answer accuracy (\%; large) over the answer-format compliance rate (\%; small). Cell shading encodes accuracy on a tier-specific scale (pale\,$\rightarrow$\,deep teal, $0\rightarrow50\%$); a numeral is printed in \textbf{black} on light (low-score) cells and in \textbf{white} on dark (high-score) cells solely for legibility---the text colour carries no extra meaning. \underline{\textbf{Best}} and \textit{second-best} per column are highlighted. After each language block we report the macro-average accuracy, the mean$\pm$std generation length (tokens, reasoning+answer), and the dominant answer language coded \textsc{EN} (predominantly English) or \textsc{TL} (requested target language).}
\label{tab:base-medium}
\end{sidewaystable*}

\begin{sidewaystable*}[p]
\centering\scriptsize
\setlength{\tabcolsep}{1.6pt}\renewcommand{\arraystretch}{1.15}
\begin{tabular}{l|cccc|ccc|cccccccccccccccccc|ccc}
\toprule
 & \multicolumn{4}{c|}{\textbf{High-resource}} & \multicolumn{3}{c|}{\textbf{Summary (HR)}} & \multicolumn{18}{c|}{\textbf{Our target languages}} & \multicolumn{3}{c}{\textbf{Summary (target)}} \\
\cmidrule(lr){2-5}\cmidrule(lr){6-8}\cmidrule(lr){9-26}\cmidrule(lr){27-29}
\textbf{Model} & \rotatebox{70}{\texttt{en}} & \rotatebox{70}{\texttt{de}} & \rotatebox{70}{\texttt{ru}} & \rotatebox{70}{\texttt{es}} & \rotatebox{70}{\textbf{Avg.}} & \rotatebox{70}{Len} & \rotatebox{70}{Lang} & \rotatebox{70}{\texttt{hi}} & \rotatebox{70}{\texttt{tr}} & \rotatebox{70}{\texttt{pl}} & \rotatebox{70}{\texttt{uk}} & \rotatebox{70}{\texttt{uz}} & \rotatebox{70}{\texttt{or}} & \rotatebox{70}{\texttt{am}} & \rotatebox{70}{\texttt{el}} & \rotatebox{70}{\texttt{kk}} & \rotatebox{70}{\texttt{cs}} & \rotatebox{70}{\texttt{he}} & \rotatebox{70}{\texttt{sr}} & \rotatebox{70}{\texttt{tt}} & \rotatebox{70}{\texttt{sk}} & \rotatebox{70}{\texttt{ca}} & \rotatebox{70}{\texttt{cv}} & \rotatebox{70}{\texttt{hsb}} & \rotatebox{70}{\texttt{dsb}} & \rotatebox{70}{\textbf{Avg.}} & \rotatebox{70}{Len} & \rotatebox{70}{Lang} \\
\midrule
\multicolumn{29}{c}{\textit{Open-weight --- small ($\leq$4B)}}\\
\bottomrule
\end{tabular}
\caption{\textbf{High-difficulty} base-prompting results on our \pmath languages vs high-resource ones from PolyMath. Each cell shows answer accuracy (\%; large) over the answer-format compliance rate (\%; small). Cell shading encodes accuracy on a tier-specific scale (pale\,$\rightarrow$\,deep teal, $0\rightarrow40\%$); a numeral is printed in \textbf{black} on light (low-score) cells and in \textbf{white} on dark (high-score) cells solely for legibility---the text colour carries no extra meaning. \underline{\textbf{Best}} and \textit{second-best} per column are highlighted. After each language block we report the macro-average accuracy, the mean$\pm$std generation length (tokens, reasoning+answer), and the dominant answer language coded \textsc{EN} (predominantly English) or \textsc{TL} (requested target language).}
\label{tab:base-high}
\end{sidewaystable*}

\begin{sidewaystable*}[p]
\centering\scriptsize
\setlength{\tabcolsep}{1.6pt}\renewcommand{\arraystretch}{1.15}
\begin{tabular}{l|cccc|ccc|cccccccccccccccccc|ccc}
\toprule
 & \multicolumn{4}{c|}{\textbf{High-resource}} & \multicolumn{3}{c|}{\textbf{Summary (HR)}} & \multicolumn{18}{c|}{\textbf{Our target languages}} & \multicolumn{3}{c}{\textbf{Summary (target)}} \\
\cmidrule(lr){2-5}\cmidrule(lr){6-8}\cmidrule(lr){9-26}\cmidrule(lr){27-29}
\textbf{Model} & \rotatebox{70}{\texttt{en}} & \rotatebox{70}{\texttt{de}} & \rotatebox{70}{\texttt{ru}} & \rotatebox{70}{\texttt{es}} & \rotatebox{70}{\textbf{Avg.}} & \rotatebox{70}{Len} & \rotatebox{70}{Lang} & \rotatebox{70}{\texttt{hi}} & \rotatebox{70}{\texttt{tr}} & \rotatebox{70}{\texttt{pl}} & \rotatebox{70}{\texttt{uk}} & \rotatebox{70}{\texttt{uz}} & \rotatebox{70}{\texttt{or}} & \rotatebox{70}{\texttt{am}} & \rotatebox{70}{\texttt{el}} & \rotatebox{70}{\texttt{kk}} & \rotatebox{70}{\texttt{cs}} & \rotatebox{70}{\texttt{he}} & \rotatebox{70}{\texttt{sr}} & \rotatebox{70}{\texttt{tt}} & \rotatebox{70}{\texttt{sk}} & \rotatebox{70}{\texttt{ca}} & \rotatebox{70}{\texttt{cv}} & \rotatebox{70}{\texttt{hsb}} & \rotatebox{70}{\texttt{dsb}} & \rotatebox{70}{\textbf{Avg.}} & \rotatebox{70}{Len} & \rotatebox{70}{Lang} \\
\midrule
\multicolumn{29}{c}{\textit{Open-weight --- small ($\leq$4B)}}\\
\bottomrule
\end{tabular}
\caption{\textbf{Top-difficulty} base-prompting results on our \pmath languages vs high-resource ones from PolyMath. Each cell shows answer accuracy (\%; large) over the answer-format compliance rate (\%; small). Cell shading encodes accuracy on a tier-specific scale (pale\,$\rightarrow$\,deep teal, $0\rightarrow25\%$); a numeral is printed in \textbf{black} on light (low-score) cells and in \textbf{white} on dark (high-score) cells solely for legibility---the text colour carries no extra meaning. \underline{\textbf{Best}} and \textit{second-best} per column are highlighted. After each language block we report the macro-average accuracy, the mean$\pm$std generation length (tokens, reasoning+answer), and the dominant answer language coded \textsc{EN} (predominantly English), or \textsc{TL} (requested target language), or \textsc{oth} is a model generated non-fluent answers full of special characters.}
\label{tab:base-top}
\end{sidewaystable*}

\clearpage
\newpage

\subsection{Models' Reasoning Length Comparison Visualization}
\label{app:models_reasoning_length}

\begin{figure}[h!]
    \centering
    \includegraphics[width=\linewidth]{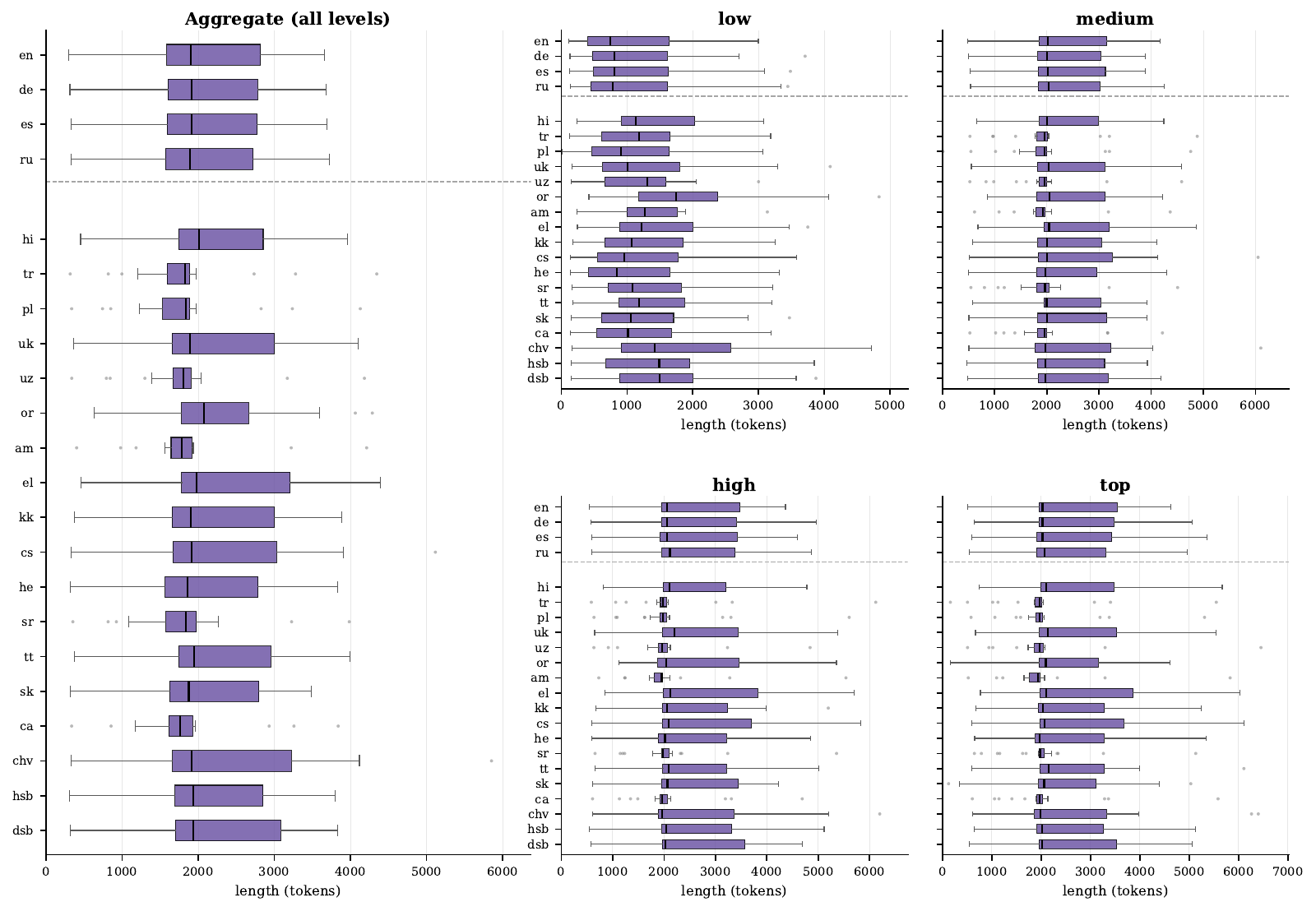}
    \caption{Distribution of reasoning and answers length for \texttt{base} prompting per language.}
    \label{fig:langs_lengths_distribution}
\end{figure}

\begin{figure}[h!]
    \centering
    \includegraphics[width=\linewidth]{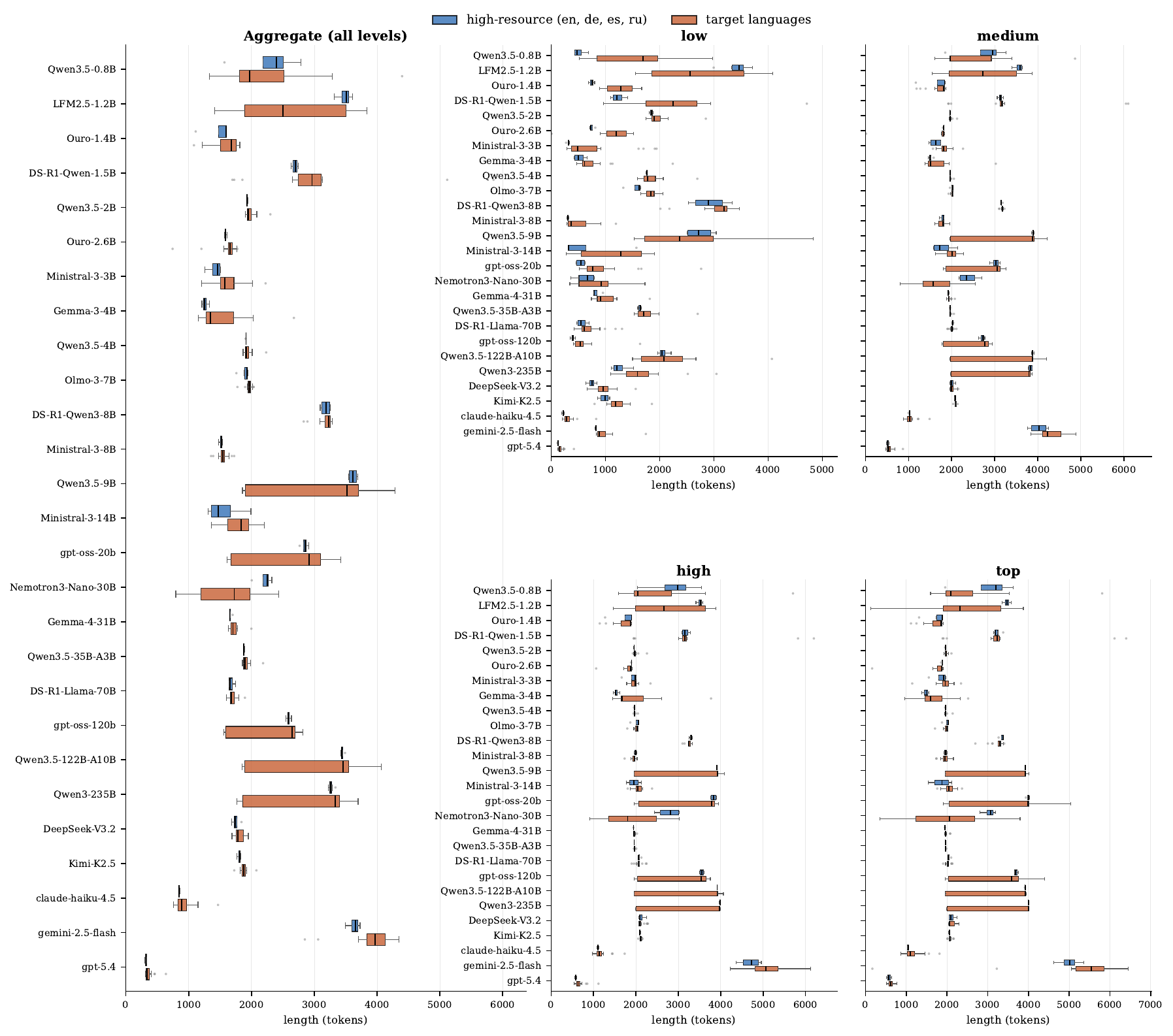}
    \caption{Distribution of models' reasoning and answers length for \texttt{base} prompting across languages types.}
    \label{fig:models_lengths_distribution}
\end{figure}

We report here the distributions of reasoning traces and answers length per language (Figure~\ref{fig:langs_lengths_distribution}) and per model (Figure~\ref{fig:models_lengths_distribution}) for aggregated metrics and per each level. We used \texttt{Qwen3-4B} base model for tokenization.

Overall, reasoning lengths do not differ substantially across languages types. For underrepresented languages, models often generate slightly longer outputs for the \texttt{lower} level, but then for others---even shorter outputs, likely due to failing to reach correct solutions. Notably, the top-performing models exhibit nearly equivalent reasoning lengths and variation across both high-resource and underrepresented languages, while producing substantially more concise outputs overall.

\clearpage

\subsection{Per-Prompting-Ablations Results}
\label{app:results_ablation_prompts}
This section reports the full per-language and per-prompt breakdown of the aggregated results in Table~\ref{tab:prompt-compare}. For every model we report four prompt designs:
\begin{itemize}
    \itemsep0em
    \item \textbf{Base} — the problem is presented and solved entirely in the target language.
    \item \textbf{EnCoT} — the problem is given in the target language but the system prompt instructs the model to perform its chain-of-thought in English as the best performing high-resource language.
    \item \textbf{Backtr.} — the problem is machine-translated back to the corresponding high-resource language and solved there.
\end{itemize}

We compare three prompting strategies exclusively on open-weight models. Backtranslation into a high-resource language yields only marginal improvements for the LFM model and does not improve performance for the remaining systems.

For several mid-sized models, including \texttt{Gemma-3-4b} and \texttt{Nemotron3-Nano-30B}, as well as the larger \texttt{gpt-oss-120b}, EN-CoT prompting improves performance for some languages. However, these gains may be partly from the additional instruction to reason step-by-step rather than from the language switch itself. Overall, no substantial stable improvements in average performance are observed with other than \texttt{base} prompting strategies.

\begin{table*}[t]
\centering\scriptsize
\setlength{\tabcolsep}{2.0pt}\renewcommand{\arraystretch}{1.08}
\resizebox{\textwidth}{!}{%
\begin{tabular}{ll|cccc|c|cccccccccccccccccc|c}
\toprule
\multicolumn{2}{l|}{} & \multicolumn{4}{c}{\textbf{High-resource}} & \multicolumn{1}{c|}{} & \multicolumn{18}{c}{\textbf{Our target languages}} & \multicolumn{1}{c}{} \\
\cmidrule(lr){3-6}\cmidrule(lr){8-25}
\textbf{Model} & \textbf{Prompt} & \rotatebox{68}{\texttt{en}} & \rotatebox{68}{\texttt{de}} & \rotatebox{68}{\texttt{ru}} & \rotatebox{68}{\texttt{es}} & \rotatebox{68}{\textbf{Avg}$_{\textrm{HR}}$} & \rotatebox{68}{\texttt{hi}} & \rotatebox{68}{\texttt{tr}} & \rotatebox{68}{\texttt{pl}} & \rotatebox{68}{\texttt{uk}} & \rotatebox{68}{\texttt{uz}} & \rotatebox{68}{\texttt{or}} & \rotatebox{68}{\texttt{am}} & \rotatebox{68}{\texttt{el}} & \rotatebox{68}{\texttt{kk}} & \rotatebox{68}{\texttt{cs}} & \rotatebox{68}{\texttt{he}} & \rotatebox{68}{\texttt{sr}} & \rotatebox{68}{\texttt{tt}} & \rotatebox{68}{\texttt{sk}} & \rotatebox{68}{\texttt{ca}} & \rotatebox{68}{\texttt{cv}} & \rotatebox{68}{\texttt{hsb}} & \rotatebox{68}{\texttt{dsb}} & \rotatebox{68}{\textbf{Avg}$_{\textrm{TL}}$} \\
\midrule
\multicolumn{26}{c}{\textit{Open-weight --- small ($\leq$4B)}}\\
\midrule
\multirow{3}{*}{\texttt{\scriptsize Qwen3.5-0.8B}} & {\scriptsize Base} & \textbf{4.7} & \textbf{3.1} & 0.7 & \textbf{3.7} & \cellcolor{black!6}3.1 & 0.4 & \textbf{0.4} & \textbf{0.5} & 1.4 & 0.1 & \textbf{1.2} & \textbf{0.1} & 1.7 & 0.4 & \textbf{1.8} & \textbf{2.4} & \textbf{0.5} & \textbf{0.7} & 1.3 & 0.3 & 0.3 & \textbf{0.4} & \textbf{0.8} & \cellcolor{black!6}\textbf{0.8} \\
 & {\scriptsize EnCoT} & 3.7 & 2.4 & \textbf{3.0} & 3.6 & \cellcolor{black!6}\textbf{3.2} & \textbf{0.9} & 0.2 & 0.3 & \textbf{2.4} & \textbf{0.2} & 0.4 & 0.0 & \textbf{1.9} & \textbf{0.5} & 1.7 & 0.9 & 0.2 & 0.6 & \textbf{2.0} & 0.2 & \textbf{0.4} & 0.3 & 0.5 & \cellcolor{black!6}0.7 \\
\bottomrule
\end{tabular}}
\caption{DW-Acc metric for prompting techniques comparison. \textbf{Avg}$_{\textrm{HR}}$ and \textbf{Avg}$_{\textrm{TL}}$ (shaded) are the macro-averages over the four high-resource and the 18 \pmath target languages. For each model and each column, the \textbf{best-performing prompting technique} is shown in \textbf{bold}. \textsc{BT} is defined only for the target languages.}
\label{tab:prompt-compare}
\end{table*}

\clearpage
\newpage

\section{Human Assessments Results per All Levels}
\label{app:humaneval_extended}

Here, we present the detailed results of an additional human evaluation of model answers and reasoning traces across languages. The evaluation includes five models of varying sizes: \texttt{gemma-3-4B}, \texttt{Ministral-3-8B}, \texttt{Nemotron3-30B}, \texttt{gpt-oss-120B}, and \texttt{DeepSeek-V3.2}. We evaluate two high-resource languages---English and Russian---as well as $11$ languages from \pmath: Hindi, Turkish, Polish, Ukrainian, Odia, Greek, Kazakh, Czech, Tatar, Chuvash, Lower Sorbian. The annotation was limited to this subset of languages due to annotators time constraints prior to submission; however, it will be expanded in future stages of this research. Still, we are covering the vast majority of the languages from all our experiments.

For each language, we selected randomly $12$ ($10\%$) samples per each level per each model. Meaning, each annotator had to go through $48$ distinct tasks across $5$ models.

\begin{figure}[h!]
    \centering
    \includegraphics[width=\linewidth]{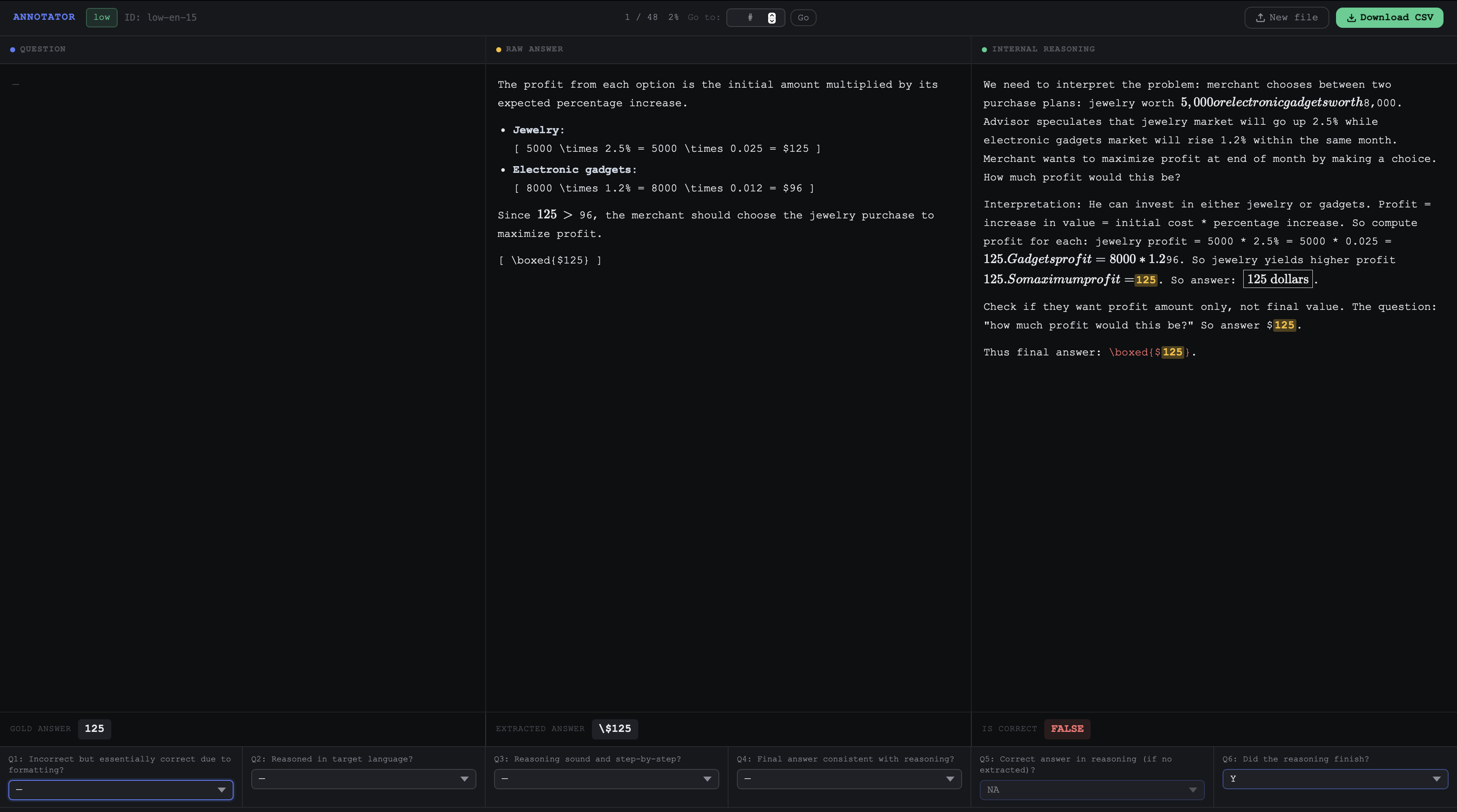}
    \caption{Interface used for the human assessment of models' answers and reasoning part.}
    \label{fig:humaneval_interface}
\end{figure}

The annotation interface used for this evaluation is shown in Figure~\ref{fig:humaneval_interface}. The evaluation criteria are detailed below:
\paragraph{Q1} For incorrect predictions: was the answer essentially correct but mismarked due to formatting or added explanation? (Y/N/NA)

\paragraph{Q2} Reasoned in target language? (Y/N/Partial)

\paragraph{Q3} Reasoning sound and step-by-step? (Y/N/Partial)

\paragraph{Q4} Final answer consistent with reasoning? (Y/N/NA)

\paragraph{Q5} If no final answer was extracted, does the correct answer appear in the reasoning? (Y/N/NA)

\paragraph{Q6} Did the reasoning finish? (Y/N)

Table~\ref{tab:human-eval-appendix} presents extended aggregated human evaluation results per all levels. Appendix~\ref{app:reasoning_traces_examples} illustrates more examples of both successful and problematic models behaviors for one target language.

\begin{table}[th!]
\centering
\resizebox{\textwidth}{!}{%
\begin{tabular}{ll cc cc cc cc cc cc}
\toprule
& & \multicolumn{2}{c}{Q1} & \multicolumn{2}{c}{Q2} & \multicolumn{2}{c}{Q3} & \multicolumn{2}{c}{Q4} & \multicolumn{2}{c}{Q5} & \multicolumn{2}{c}{Q6} \\
\cmidrule(lr){3-4}\cmidrule(lr){5-6}\cmidrule(lr){7-8}\cmidrule(lr){9-10}\cmidrule(lr){11-12}\cmidrule(lr){13-14}
Model & Level & H & U & H & U & H & U & H & U & H & U & H & U \\
\midrule
\multirow{4}{*}{gemma-3-4B}
  & low    & 0 & 1 & 100 & 94 & 100 & 73 & 100 & 82 & 91 & 23 & 100 & 91 \\
  & medium & 0 & 6 & 100 & 84 & 96 & 42 & 78 & 57 & 0 & 4 & 75 & 71 \\
  & high   & 0 & 2 & 96 & 89 & 92 & 38 & 88 & 54 & 0 & 2 & 67 & 61 \\
  & top    & 0 & 0 & 100 & 89 & 88 & 42 & 62 & 67 & 0 & 0 & 67 & 70 \\
\midrule
\multirow{4}{*}{Ministral-3-8B}
  & low    & 0 & 4 & 75 & 51 & 100 & 73 & 100 & 85 & 0 & 14 & 100 & 95 \\
  & medium & 8 & 8 & 8 & 23 & 100 & 67 & 100 & 55 & 0 & 8 & 8 & 55 \\
  & high   & 0 & 1 & 8 & 15 & 100 & 64 & 100 & 51 & 0 & 5 & 17 & 46 \\
  & top    & 0 & 0 & 0 & 8 & 100 & 55 & 100 & 24 & 0 & 0 & 0 & 28 \\
\midrule
\multirow{4}{*}{Nemotron3\_30B}
  & low    & 0 & 1 & 62 & 20 & 67 & 35 & 100 & 57 & 20 & 3 & 67 & 71 \\
  & medium & 0 & 2 & 29 & 5 & 38 & 20 & 100 & 35 & 0 & 4 & 25 & 49 \\
  & high   & 0 & 0 & 33 & 5 & 33 & 22 & 100 & 39 & 0 & 3 & 8 & 53 \\
  & top    & 0 & 0 & 42 & 6 & 42 & 27 & 100 & 55 & 0 & 0 & 8 & 49 \\
\midrule
\multirow{4}{*}{gpt-oss-120b}
  & low    & 50 & 46 & 12 & 22 & 100 & 89 & 100 & 98 & 100 & 17 & 100 & 100 \\
  & medium & 33 & 21 & 0 & 10 & 100 & 75 & 100 & 76 & 0 & 19 & 75 & 67 \\
  & high   & 14 & 6 & 0 & 3 & 96 & 61 & 100 & 75 & 20 & 12 & 67 & 47 \\
  & top    & 9 & 8 & 17 & 6 & 75 & 56 & 100 & 70 & 12 & 6 & 33 & 40 \\
\midrule
\multirow{4}{*}{DeepSeek-V3.2}
  & low    & 0 & 4 & 100 & 58 & 100 & 88 & 100 & 95 & 100 & 62 & 100 & 88 \\
  & medium & 10 & 6 & 100 & 32 & 100 & 50 & 100 & 50 & 0 & 13 & 25 & 22 \\
  & high   & 0 & 2 & 100 & 23 & 100 & 43 & 100 & 31 & 9 & 5 & 8 & 6 \\
  & top    & 0 & 1 & 100 & 24 & 100 & 43 & 100 & 4 & 0 & 1 & 0 & 1 \\
\bottomrule
\end{tabular}%
}
\caption{Human evaluation per difficulty level. Cells are \% ``Yes'' among applicable annotations (Partial = 0.5 for Q2--Q3; NA/blank excluded). H: high-resource (RU); U: underrepresented (10 PluraMath languages).}
\label{tab:human-eval-appendix}
\end{table}

\clearpage
\newpage

\subsection{\pmath Answers and Reasoning Traces Examples}
\label{app:reasoning_traces_examples}
We present examples of model outputs, reasoning traces, and corresponding human annotation analyses. For illustration, we use Ukrainian examples covering all models included in the annotation process: \texttt{gpt-oss-120b}, \texttt{DeepSeek-V3.2}, \texttt{nemotron3\_nano\_omni\_30ba3b}, \texttt{Ministral-3-8B-Reasoning-2512}, and \texttt{gemma-3-4b}.

\begin{figure}[h!]
    \centering
    \includegraphics[width=\linewidth]{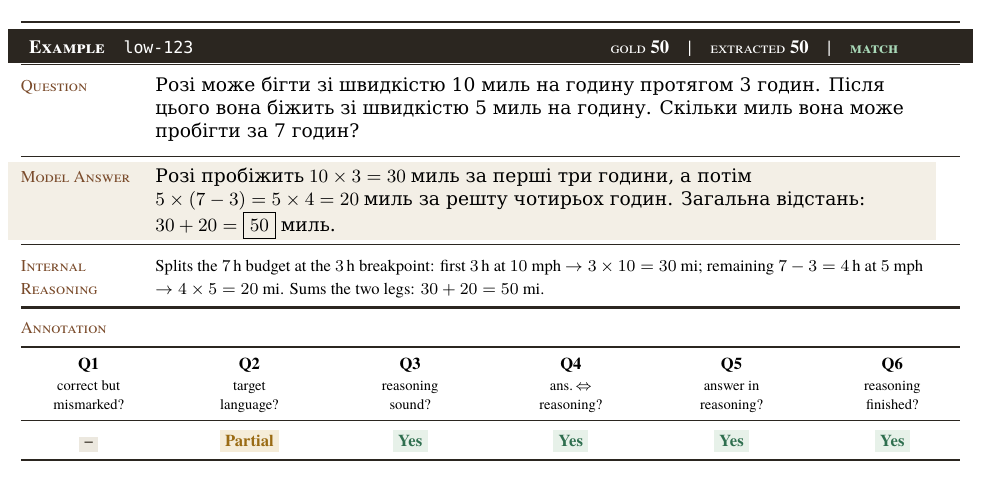}
    \caption{Example of a correct answer and reasoning from \texttt{gpt-oss-120b}.}
    \label{fig:uk_example_reasoning_low_123}
\end{figure}

\begin{figure}[h!]
    \centering
    \includegraphics[width=\linewidth]{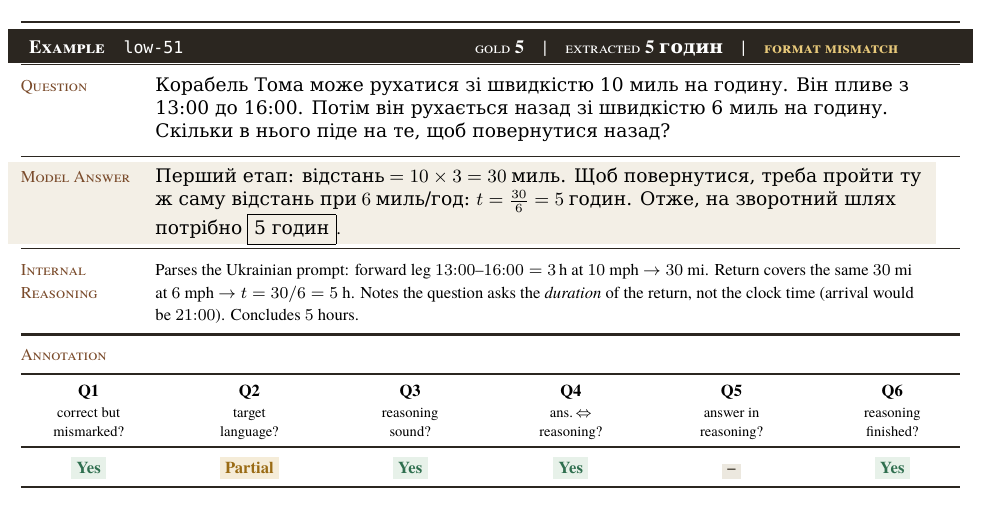}
    \caption{Example of a correct reasoning and semantically correct but mismatched answer from \texttt{gpt-oss-120b}.}
    \label{fig:uk_example_reasoning_low_51}
\end{figure}

Figure~\ref{fig:uk_example_reasoning_low_51} presents an example of a correct prediction and reasoning trace from \texttt{gpt-oss-120b}; however, the model performs its reasoning in English. Figure~\ref{fig:uk_example_reasoning_low_123} shows a case where the predicted answer was semantically correct but marked incorrect due to formatting differences preventing an exact match. Overall, \texttt{gpt-oss-120b} predominantly reasoned in English across all evaluated languages.

\begin{figure}[h!]
    \centering
    \includegraphics[width=\linewidth]{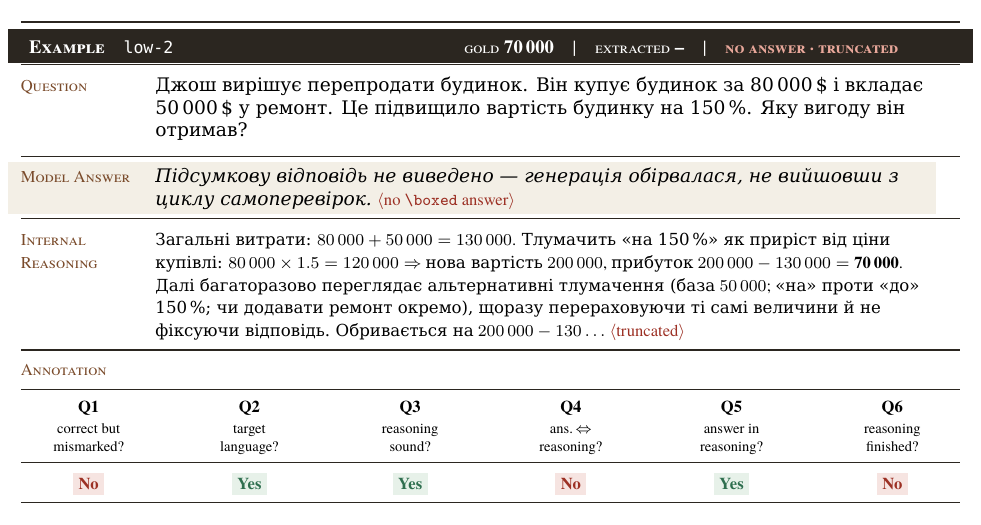}
    \caption{\texttt{DeepSeek-V3.2} typically reasoned in the target language, but quite often did not finish correct answer generation due to repetitive self-checking loops.}
    \label{fig:uk_example_reasoning_low_2}
\end{figure}

\texttt{DeepSeek-V3.2} typically reasoned in the target language, as illustrated by the Ukrainian example in Figure~\ref{fig:uk_example_reasoning_low_2}. However, the model frequently produced excessively long reasoning traces with repetitive self-verification loops. In the shown example, the model arrived at the correct answer but continued generating additional verification steps and failed to complete the final response properly.

\begin{figure}[h!]
    \centering
    \includegraphics[width=\linewidth]{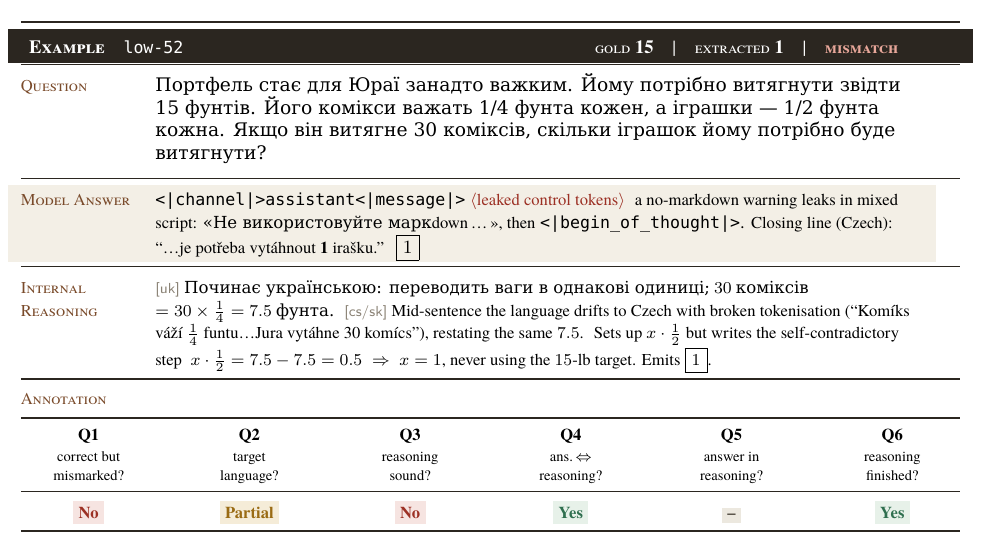}
    \caption{Example of a mixed languages reasoning and not so fluent answer from \texttt{nemotron3\_nano\_omni\_30ba3b}.}
    \label{fig:uk_example_reasoning_low_52}
\end{figure}

\begin{figure}[h!]
    \centering
    \includegraphics[width=\linewidth]{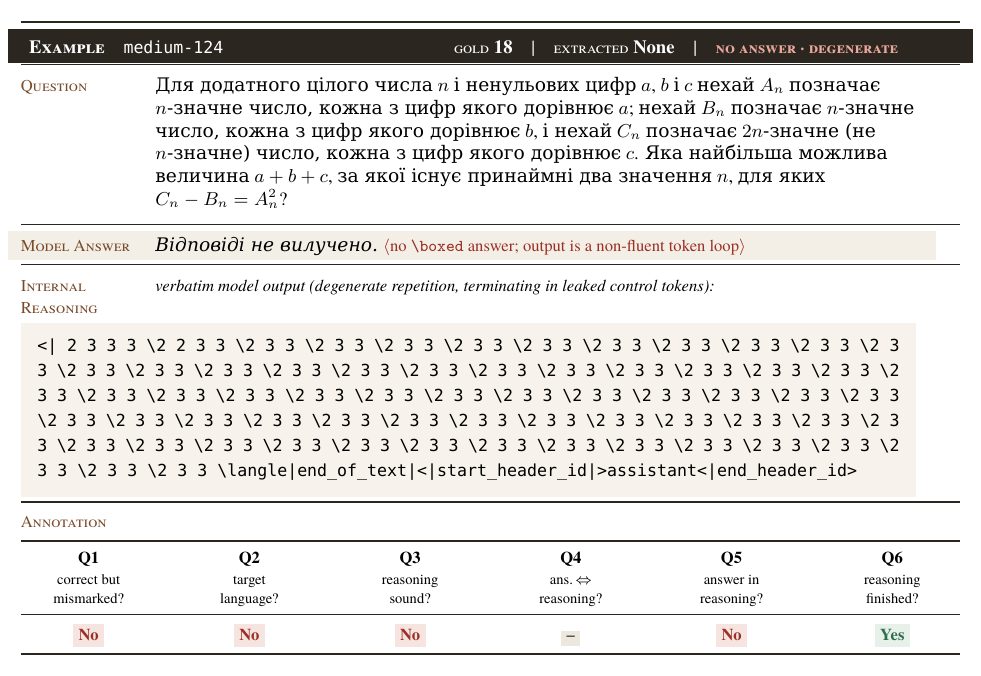}
    \caption{\texttt{Nemotron3\_nano\_omni\_30ba3b} could also produce absolutely non-fluent reasoning with even leaked control tokens.}
    \label{fig:uk_example_reasoning_medium-124}
\end{figure}

\texttt{Nemotron3\_nano\_omni\_30ba3b} frequently produced non-fluent reasoning traces (Figure~\ref{fig:uk_example_reasoning_medium-124}), often mixing the target language with other languages (Figure~\ref{fig:uk_example_reasoning_low_52}). The model also commonly failed to complete its reasoning process, although it still generated a final answer in \$$\backslash$boxed\{\}\$ format, sometimes appearing effectively random.

\begin{figure}[h!]
    \centering
    \includegraphics[width=\linewidth]{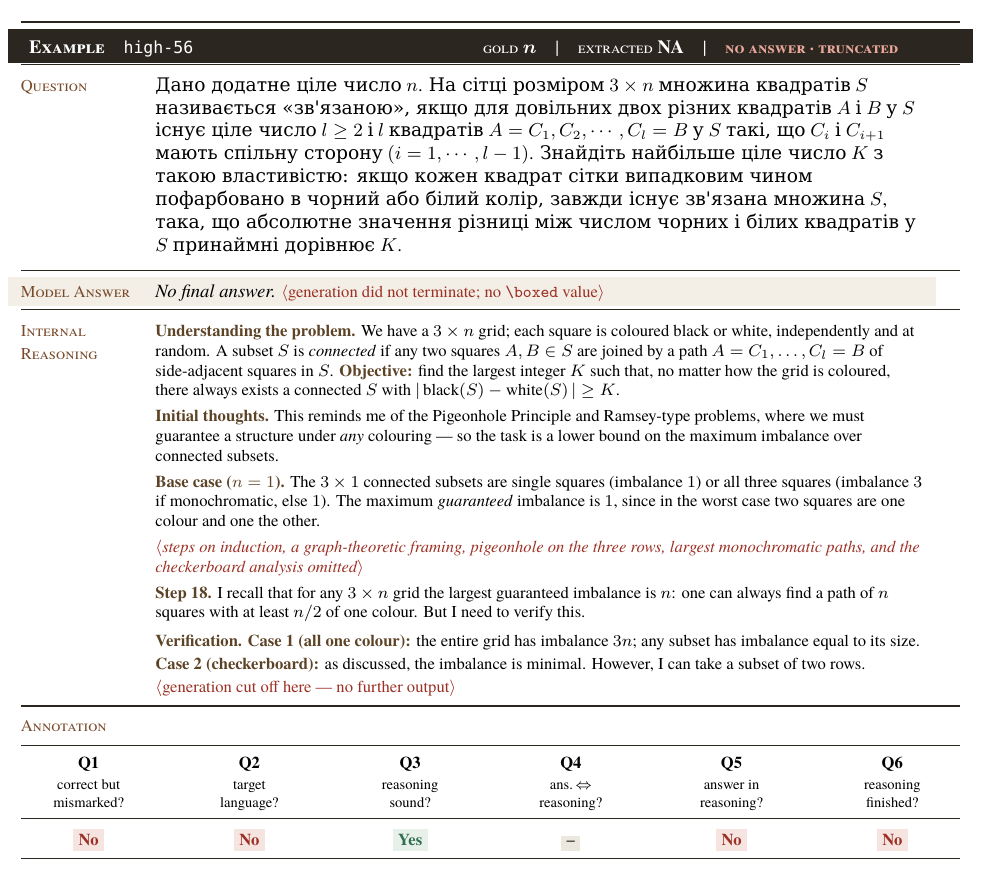}
    \caption{Example of the unfinished reasoning from \texttt{Ministral-3-8B-Reasoning-2512}.}
    \label{fig:uk_example_reasoning_high-56}
\end{figure}

In contrast to \texttt{gpt-oss-120b}, which consistently reasoned in English, and \texttt{DeepSeek-V3.2}, which predominantly reasoned in the target language, \texttt{Ministral-3-8B-Reasoning-2512} exhibited mixed behavior, alternating between target-language and primarily English reasoning. Most incorrect predictions resulted from incomplete reasoning traces as in Figure~\ref{fig:uk_example_reasoning_high-56}.

\begin{figure}
    \centering
    \includegraphics[width=\linewidth]{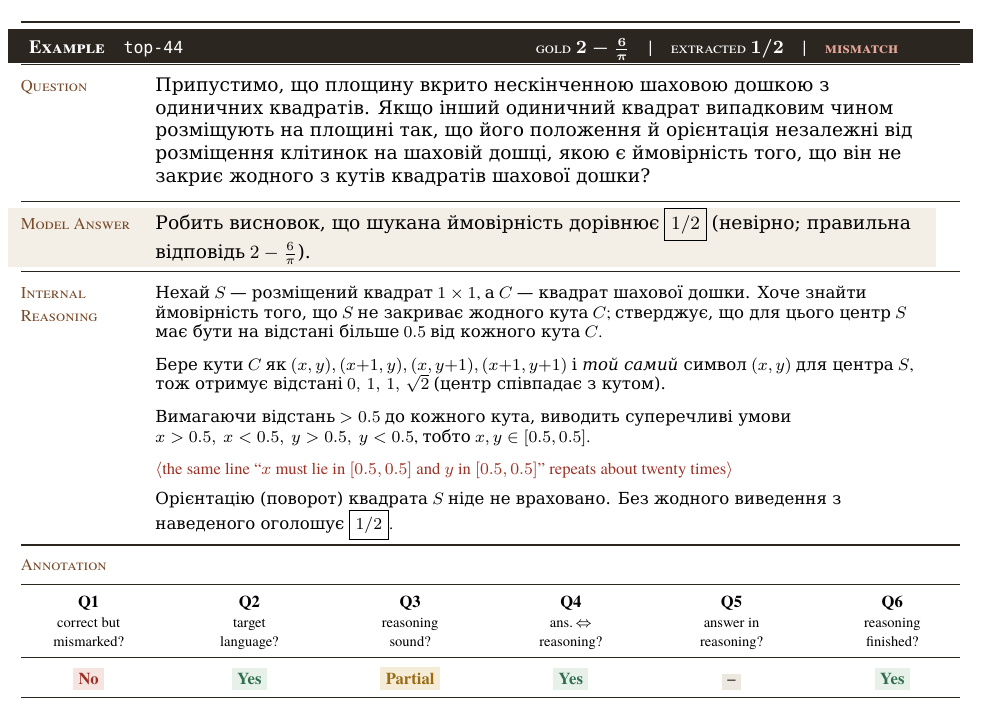}
    \caption{\texttt{Gemma-3-4b} consistently reasoned fluently and in the target language, but there could have been wrong reasoning steps.}
    \label{fig:uk_example_reasoning_top-44}
\end{figure}

Finally, \texttt{gemma-3-4b} consistently reasoned in the target language and, even when producing incorrect answers, generated coherent reasoning traces containing flawed intermediate steps (Figure~\ref{fig:uk_example_reasoning_top-44}).

\clearpage
\newpage

\section{Detailed Translation Quality and Translations Examples Results with Reasoning Models}
\label{app:reasoning_translations_details}

We evaluate a set of reasoning models on machine translation using the FLORES+~\cite{nllb-24} \texttt{dev} split. For the Sorbian languages, we acquire test sets from the authors of \citet{okabe-etal-2025-findings} and sample 300 items from the 4000 segments. For the languages present in FLORES+, we run machine translation both from and into English, whereas for the two Sorbian languages we only run translation from German. We tested a subset of seven opensource tested models: \texttt{Qwen3.5-0.8B} with reasoning and no-think mode, \texttt{Ministral-3-3B} and \texttt{Ministral-3-14B} with reasoning and no-think mode, \texttt{Nemotron-3-Nano-Omni-30B}, \texttt{Qwen3.5-35B-A3B}, \texttt{gpt-oss-120B}, \texttt{Qwen3.5-122B-A10B}, and \texttt{DeepSeek-V3.2}.

We inference models either locally or via API calls. API models are called with \texttt{reasoning\_effort=``medium’’} to match the mathematical reasoning experiments, and limited to a maximum output length of $2048$ tokens. With locally inferenced models, we have more fine-grained control: We use the recommended decoding parameters when specified, set the maximum number of output tokens to $4096$, and use a maximum thinking token budget of $2048$. When models use up the full thinking token budget, we forcedly stop the reasoning trace by inserting “I’ll stop thinking and provide my answer now.”, followed by the appropriate reasoning-end-string.

\begin{table}[h]
\centering
\begin{tabular}{lrrr}
\toprule
\textbf{Parameter} & \textbf{Default} & \textbf{Qwen3.5} & \textbf{Ministral} \\
\midrule
temperature & 0.6 & 1.0 & 0.7 \\
top\_p & 0.95 & 0.95 & 0.95 \\
top\_k& 20 & 20 & --- \\
min\_p & 0 & 0.0 & --- \\
repetition\_penalty & 1.05 & 1.0 & 1.05 \\
presence\_penalty & --- & 1.5 & --- \\
max\_tokens & 4096 & 4096 & 4096 \\
thinking\_token\_budget & 2048 & 2048 & 2048 \\
\bottomrule
\end{tabular}
\caption{Sampling parameters for local models for translation capabilities assessment.}
\end{table}

\begin{figure}[h!]
    \centering
    \includegraphics[width=\linewidth]{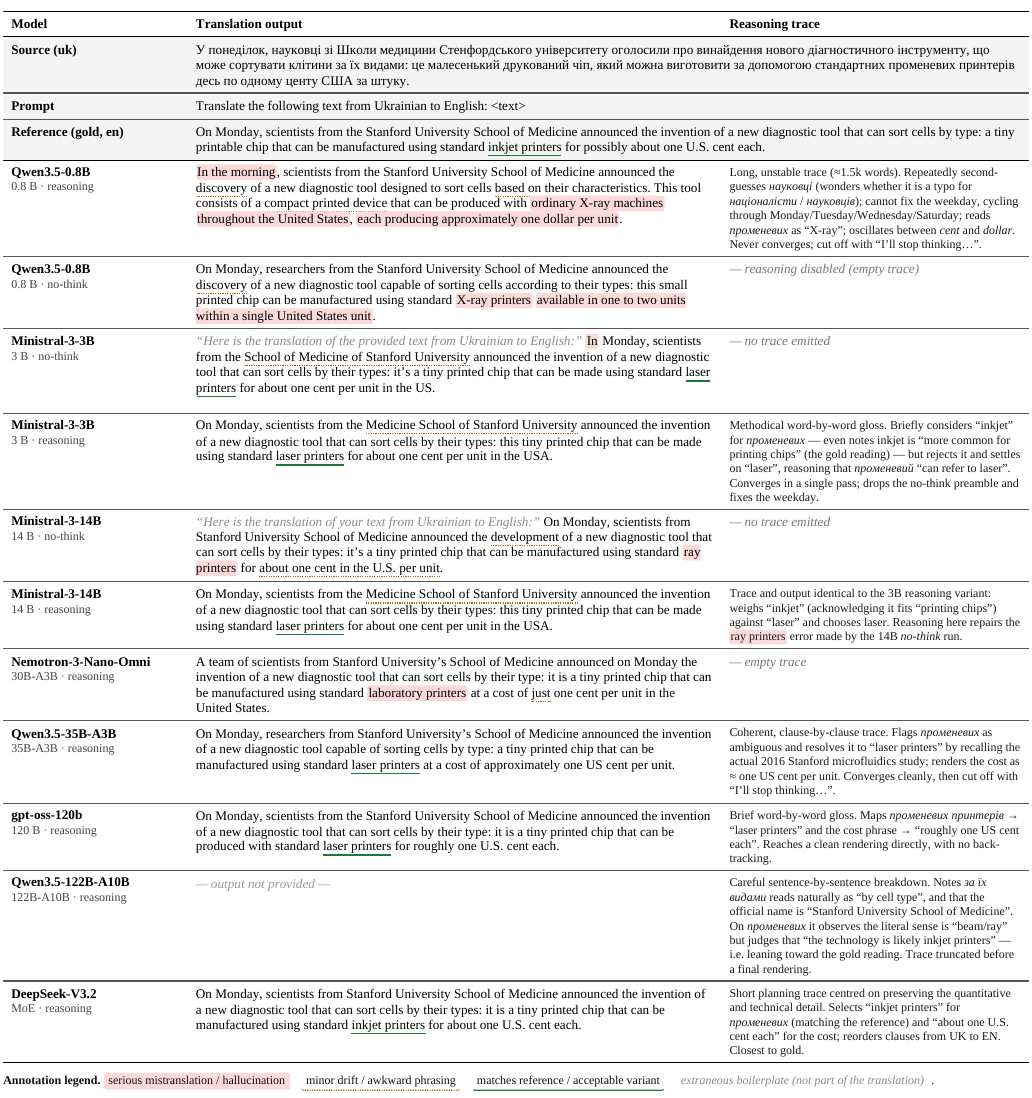}
    \caption{Translation with reasoning models example for the Ukrainian sample. Some models tend to hallucinate additional information or add in the output unnecessary explanation; bigger models translate better, but tend to paraphrase additionally.}
    \label{fig:app-translation-example}
\end{figure}

We evaluate the models’ performance using chrF++, since this metric supports all our target languages---full results in Table~\ref{tab:app-chrf}. Additionally, we calculate the mean length of the MT and thinking outputs (Table~\ref{tab:app-length}), and the percentage of empty outputs (Table~\ref{tab:app-empty}). Explicit examples of models' behavior for the translation task are in Figure~\ref{fig:app-translation-example}.

Our results show rather poor chrF++ scores throughout, though with some variation across language pairs and models. For instance, Olmo consistently adds the most additional information to the machine translation output, leading to long answers and low chrF++ scores. How long the models ``thought'' was more determined by the model than by the language pair, with a weak negative correlation of reasoning length and chrF++. The smaller, locally-inferenced models very frequently used up the full thinking token budget, while GPT-OSS-120B and DeepSeek-V3.2 were more discerning. Translation into the two Sorbian languages achieved some of the best chrF++ scores, but this must be attributed to these segments being shorter on average.

\begin{table*}[t]\centering\scriptsize
\setlength{\tabcolsep}{3pt}\renewcommand{\arraystretch}{1.1}
\definecolor{emptyc}{RGB}{244,242,238}
\resizebox{\textwidth}{!}{%
\begin{tabular}{lcccccccccccc}
\toprule
\textbf{Direction} & \rotatebox{70}{Qwen3.5-0.8B} & \rotatebox{70}{Qwen3.5-0.8B (NT)} & \rotatebox{70}{Ministral-3B} & \rotatebox{70}{Ministral-3B (NT)} & \rotatebox{70}{OLMo-3-7B} & \rotatebox{70}{Ministral-14B} & \rotatebox{70}{Ministral-14B (NT)} & \rotatebox{70}{Nemotron-30B} & \rotatebox{70}{Qwen3.5-35B} & \rotatebox{70}{gpt-oss-120b} & \rotatebox{70}{Qwen3.5-122B} & \rotatebox{70}{DeepSeek-V3.2} \\
\midrule
\multicolumn{13}{l}{\textit{Into English ($X\rightarrow$en)}}\\
Amharic\,$\rightarrow$\,en & \cellcolor[RGB]{232,239,240}\textcolor{black}{1.3} & \cellcolor[RGB]{144,179,183}\textcolor{black}{8.8} & \cellcolor[RGB]{137,175,179}\textcolor{black}{9.3} & \cellcolor[RGB]{185,208,210}\textcolor{black}{5.2} & \cellcolor[RGB]{235,241,242}\textcolor{black}{1.0} & \cellcolor[RGB]{184,207,209}\textcolor{black}{5.3} & \cellcolor[RGB]{222,233,234}\textcolor{black}{2.1} & \cellcolor[RGB]{212,226,227}\textcolor{black}{2.9} & \cellcolor[RGB]{133,172,176}\textcolor{black}{9.7} & \cellcolor[RGB]{111,157,162}\textcolor{white}{\textbf{11.6}} & \cellcolor[RGB]{247,250,250}\textcolor{black}{0.0} & \cellcolor[RGB]{132,171,175}\textcolor{black}{9.8} \\
Catalan\,$\rightarrow$\,en & \cellcolor[RGB]{159,190,193}\textcolor{black}{7.5} & \cellcolor[RGB]{130,170,175}\textcolor{black}{\textbf{9.9}} & \cellcolor[RGB]{139,176,180}\textcolor{black}{9.2} & \cellcolor[RGB]{156,188,191}\textcolor{black}{7.7} & \cellcolor[RGB]{222,233,234}\textcolor{black}{2.1} & \cellcolor[RGB]{133,172,176}\textcolor{black}{9.7} & \cellcolor[RGB]{171,198,201}\textcolor{black}{6.4} & \cellcolor[RGB]{146,181,185}\textcolor{black}{8.6} & \cellcolor[RGB]{143,179,183}\textcolor{black}{8.9} & \cellcolor[RGB]{142,178,182}\textcolor{black}{8.9} & \cellcolor[RGB]{228,237,238}\textcolor{black}{1.6} & \cellcolor[RGB]{140,176,180}\textcolor{black}{9.1} \\
\bottomrule
\end{tabular}}
\caption{\textsc{chrF}\texttt{++} translation quality per model and translation direction. Rows are translation directions grouped into $X\rightarrow$en, en$\rightarrow X$, and German-pivot (Sorbian) directions; columns are models (\textsc{nt}\,=\,non-thinking variant). Cell shading encodes \textsc{chrF}\texttt{++} (pale\,$\rightarrow$\,deep teal); the \textbf{best} model per direction is in \textbf{bold}. Higher is better.}
\label{tab:app-chrf}
\end{table*}

\begin{table*}[t]\centering\scriptsize
\setlength{\tabcolsep}{3pt}\renewcommand{\arraystretch}{1.05}
\resizebox{\textwidth}{!}{%
\begin{tabular}{lrrrrrrrrrrrr}
\toprule
\textbf{Direction} & \rotatebox{70}{Qwen3.5-0.8B} & \rotatebox{70}{Qwen3.5-0.8B (NT)} & \rotatebox{70}{Ministral-3B} & \rotatebox{70}{Ministral-3B (NT)} & \rotatebox{70}{OLMo-3-7B} & \rotatebox{70}{Ministral-14B} & \rotatebox{70}{Ministral-14B (NT)} & \rotatebox{70}{Nemotron-30B} & \rotatebox{70}{Qwen3.5-35B} & \rotatebox{70}{gpt-oss-120b} & \rotatebox{70}{Qwen3.5-122B} & \rotatebox{70}{DeepSeek-V3.2} \\
\midrule
\multicolumn{13}{l}{\textit{Into English ($X\rightarrow$en)}}\\
Amharic\,$\rightarrow$\,en & 6{,}464 & 154 & 7{,}953 & 345 & 8{,}031 & 10{,}747 & 747 & 561 & 7{,}129 & 1{,}254 & 2{,}723 & 1{,}172 \\
Catalan\,$\rightarrow$\,en & 8{,}299 & 134 & 3{,}829 & 189 & 7{,}893 & 8{,}661 & 195 & 138 & 9{,}308 & 867 & 3{,}950 & 931 \\
Chuvash\,$\rightarrow$\,en & 7{,}350 & 152 & 6{,}904 & 283 & 9{,}616 & 10{,}477 & 676 & 249 & 7{,}821 & 2{,}540 & 2{,}927 & 1{,}289 \\
Czech\,$\rightarrow$\,en & 8{,}090 & 132 & 4{,}117 & 189 & 8{,}684 & 8{,}895 & 179 & 134 & 9{,}041 & 822 & 3{,}909 & 1{,}129 \\
Dutch\,$\rightarrow$\,en & 8{,}473 & 132 & 3{,}731 & 180 & 7{,}616 & 8{,}587 & 195 & 133 & 9{,}252 & 761 & 4{,}035 & 920 \\
Greek\,$\rightarrow$\,en & 7{,}779 & 140 & 4{,}718 & 193 & 8{,}592 & 8{,}998 & 199 & 141 & 9{,}294 & 879 & 3{,}743 & 1{,}305 \\
Hebrew\,$\rightarrow$\,en & 7{,}626 & 136 & 4{,}625 & 192 & 8{,}802 & 7{,}864 & 192 & 133 & 8{,}774 & 882 & 3{,}612 & 1{,}303 \\
Hindi\,$\rightarrow$\,en & 7{,}666 & 156 & 4{,}516 & 192 & 7{,}928 & 8{,}329 & 195 & 135 & 8{,}554 & 903 & 3{,}393 & 1{,}192 \\
Kazakh\,$\rightarrow$\,en & 7{,}643 & 144 & 5{,}073 & 207 & 9{,}698 & 9{,}230 & 209 & 147 & 8{,}948 & 1{,}068 & 3{,}596 & 1{,}098 \\
Odia\,$\rightarrow$\,en & 6{,}908 & 152 & 7{,}363 & 278 & 6{,}268 & 7{,}970 & 471 & 195 & 7{,}400 & 1{,}004 & 2{,}873 & 992 \\
Polish\,$\rightarrow$\,en & 8{,}284 & 135 & 3{,}788 & 189 & 8{,}012 & 8{,}622 & 175 & 133 & 9{,}500 & 821 & 3{,}999 & 1{,}158 \\
Romanian\,$\rightarrow$\,en & 8{,}317 & 135 & 3{,}970 & 185 & 7{,}776 & 8{,}533 & 182 & 136 & 9{,}282 & 815 & 3{,}972 & 948 \\
Serbian\,$\rightarrow$\,en & 7{,}866 & 135 & 4{,}144 & 208 & 8{,}844 & 7{,}932 & 234 & 142 & 9{,}084 & 893 & 3{,}742 & 1{,}204 \\
Slovak\,$\rightarrow$\,en & 8{,}094 & 135 & 4{,}515 & 197 & 9{,}318 & 8{,}749 & 194 & 136 & 9{,}191 & 869 & 3{,}898 & 1{,}081 \\
Tatar\,$\rightarrow$\,en & 7{,}533 & 147 & 5{,}350 & 208 & 9{,}979 & 8{,}660 & 224 & 196 & 8{,}716 & 1{,}203 & 3{,}456 & 1{,}166 \\
Ukrainian\,$\rightarrow$\,en & 7{,}983 & 135 & 4{,}153 & 198 & 7{,}701 & 8{,}335 & 184 & 137 & 9{,}206 & 843 & 3{,}848 & 1{,}065 \\
Uzbek\,$\rightarrow$\,en & 8{,}144 & 150 & 5{,}485 & 216 & 10{,}612 & 9{,}082 & 261 & 159 & 9{,}197 & 1{,}119 & 3{,}814 & 1{,}149 \\
\midrule
\multicolumn{13}{l}{\textit{Out of English (en$\rightarrow X$)}}\\
en\,$\rightarrow$\,Amharic & 7{,}232 & 129 & 5{,}236 & 501 & 5{,}432 & 6{,}601 & 245 & 490 & 5{,}902 & 950 & 2{,}236 & 1{,}107 \\
en\,$\rightarrow$\,Catalan & 7{,}944 & 146 & 4{,}627 & 157 & 10{,}745 & 10{,}113 & 141 & 272 & 9{,}251 & 966 & 3{,}895 & 915 \\
en\,$\rightarrow$\,Chuvash & 8{,}314 & 188 & 5{,}275 & 515 & 10{,}219 & 11{,}049 & 748 & 1{,}324 & 7{,}665 & 1{,}840 & 2{,}907 & 1{,}098 \\
en\,$\rightarrow$\,Czech & 8{,}218 & 143 & 5{,}691 & 149 & 11{,}523 & 10{,}875 & 140 & 233 & 8{,}904 & 939 & 3{,}519 & 1{,}054 \\
en\,$\rightarrow$\,Dutch & 8{,}294 & 146 & 5{,}442 & 154 & 10{,}666 & 10{,}479 & 145 & 181 & 9{,}488 & 952 & 3{,}896 & 942 \\
en\,$\rightarrow$\,Greek & 8{,}061 & 155 & 9{,}162 & 266 & 8{,}808 & 11{,}709 & 228 & 283 & 8{,}565 & 926 & 3{,}647 & 994 \\
en\,$\rightarrow$\,Hebrew & 7{,}733 & 108 & 8{,}005 & 224 & 9{,}337 & 11{,}144 & 128 & 917 & 8{,}102 & 783 & 3{,}141 & 964 \\
en\,$\rightarrow$\,Hindi & 7{,}362 & 134 & 10{,}013 & 403 & 8{,}267 & 11{,}764 & 348 & 614 & 8{,}119 & 678 & 3{,}156 & 989 \\
en\,$\rightarrow$\,Kazakh & 8{,}204 & 156 & 6{,}231 & 365 & 9{,}297 & 10{,}570 & 231 & 1{,}058 & 7{,}910 & 993 & 3{,}143 & 960 \\
en\,$\rightarrow$\,Odia & 7{,}683 & 138 & 7{,}084 & 411 & 5{,}084 & 8{,}848 & 407 & 271 & 6{,}668 & 949 & 2{,}465 & 957 \\
en\,$\rightarrow$\,Polish & 8{,}227 & 139 & 5{,}965 & 178 & 11{,}413 & 10{,}896 & 144 & 305 & 9{,}168 & 904 & 3{,}693 & 962 \\
en\,$\rightarrow$\,Romanian & 7{,}987 & 141 & 5{,}030 & 207 & 10{,}793 & 9{,}956 & 142 & 362 & 9{,}259 & 874 & 3{,}874 & 913 \\
en\,$\rightarrow$\,Serbian & 8{,}002 & 160 & 6{,}192 & 154 & 10{,}906 & 11{,}343 & 139 & 1{,}258 & 8{,}737 & 1{,}028 & 3{,}319 & 1{,}081 \\
en\,$\rightarrow$\,Slovak & 8{,}260 & 131 & 5{,}523 & 163 & 12{,}054 & 10{,}578 & 162 & 327 & 8{,}876 & 966 & 3{,}498 & 923 \\
en\,$\rightarrow$\,Tatar & 7{,}978 & 145 & 6{,}313 & 412 & 9{,}641 & 11{,}155 & 366 & 1{,}471 & 7{,}839 & 1{,}210 & 3{,}016 & 988 \\
en\,$\rightarrow$\,Ukrainian & 7{,}829 & 136 & 4{,}983 & 174 & 9{,}548 & 9{,}312 & 135 & 342 & 8{,}757 & 852 & 3{,}446 & 906 \\
en\,$\rightarrow$\,Uzbek & 8{,}619 & 206 & 6{,}594 & 659 & 10{,}150 & 10{,}285 & 220 & 1{,}000 & 8{,}704 & 1{,}134 & 3{,}352 & 994 \\
\midrule
\multicolumn{13}{l}{\textit{German pivot}}\\
de\,$\rightarrow$\,L.\,Sorbian & 9{,}115 & 67 & 4{,}519 & 211 & 12{,}161 & 12{,}288 & 257 & 858 & 8{,}288 & 1{,}717 & 3{,}404 & 966 \\
de\,$\rightarrow$\,U.\,Sorbian & 9{,}170 & 80 & 4{,}795 & 246 & 12{,}349 & 12{,}994 & 245 & 975 & 8{,}353 & 1{,}881 & 3{,}394 & 1{,}002 \\
\bottomrule
\end{tabular}}
\caption{Mean generation length (tokens) per model and translation direction, summing the answer/translation output and the reasoning (``thinking) trace. Rows are translation directions grouped into $X\rightarrow$en, en$\rightarrow X$, and German-pivot (Sorbian) directions; columns are models (\textsc{nt}\,=\,non-thinking variant). Larger values indicate longer total generations.}
\label{tab:app-length}
\end{table*}

\begin{table*}[t]\centering\scriptsize
\setlength{\tabcolsep}{3pt}\renewcommand{\arraystretch}{1.05}
\resizebox{\textwidth}{!}{%
\begin{tabular}{lrrrrrrrrrrrr}
\toprule
\textbf{Direction} & \rotatebox{70}{Qwen3.5-0.8B} & \rotatebox{70}{Qwen3.5-0.8B (NT)} & \rotatebox{70}{Ministral-3B} & \rotatebox{70}{Ministral-3B (NT)} & \rotatebox{70}{OLMo-3-7B} & \rotatebox{70}{Ministral-14B} & \rotatebox{70}{Ministral-14B (NT)} & \rotatebox{70}{Nemotron-30B} & \rotatebox{70}{Qwen3.5-35B} & \rotatebox{70}{gpt-oss-120b} & \rotatebox{70}{Qwen3.5-122B} & \rotatebox{70}{DeepSeek-V3.2} \\
\midrule
\multicolumn{13}{l}{\textit{Into English ($X\rightarrow$en)}}\\
Amharic\,$\rightarrow$\,en & 0\,/\,0 & 0\,/\,100 & 21\,/\,0 & 0\,/\,100 & 0\,/\,1.5 & 61\,/\,0 & 0\,/\,100 & 0\,/\,100 & 5.5\,/\,0 & 7.2\,/\,0 & 99\,/\,0 & 1.0\,/\,0 \\
Catalan\,$\rightarrow$\,en & 0\,/\,0 & 0\,/\,100 & 0.1\,/\,0 & 0\,/\,100 & 0\,/\,0 & 10\,/\,0 & 0\,/\,100 & 0\,/\,100 & 1.5\,/\,0 & 0\,/\,0 & 90\,/\,0 & 0\,/\,0 \\
Chuvash\,$\rightarrow$\,en & 0\,/\,0 & 0\,/\,100 & 0.9\,/\,0 & 0\,/\,100 & 0\,/\,0.4 & 20\,/\,0 & 0\,/\,100 & 0\,/\,100 & 5.3\,/\,0 & 53\,/\,0 & 100\,/\,0 & 1.2\,/\,0 \\
Czech\,$\rightarrow$\,en & 0\,/\,0 & 0\,/\,100 & 0.1\,/\,0 & 0\,/\,100 & 0\,/\,0 & 10\,/\,0 & 0\,/\,100 & 0\,/\,100 & 1.1\,/\,0 & 0.1\,/\,0 & 93\,/\,0 & 0.9\,/\,0 \\
Dutch\,$\rightarrow$\,en & 0\,/\,0 & 0\,/\,100 & 0\,/\,0 & 0\,/\,100 & 0\,/\,0 & 9.0\,/\,0 & 0\,/\,100 & 0\,/\,100 & 0.8\,/\,0 & 0\,/\,0 & 92\,/\,0 & 0\,/\,0 \\
Greek\,$\rightarrow$\,en & 0\,/\,0 & 0\,/\,100 & 0.3\,/\,0 & 0\,/\,100 & 0\,/\,0.1 & 11\,/\,0 & 0\,/\,100 & 0\,/\,100 & 2.2\,/\,0 & 0\,/\,0 & 99\,/\,63 & 1.7\,/\,0 \\
Hebrew\,$\rightarrow$\,en & 0\,/\,0 & 0\,/\,100 & 0.5\,/\,0 & 0\,/\,100 & 0\,/\,0.4 & 7.8\,/\,0 & 0\,/\,100 & 0\,/\,100 & 2.1\,/\,0 & 0\,/\,0 & 93\,/\,0 & 1.7\,/\,0 \\
Hindi\,$\rightarrow$\,en & 0\,/\,0 & 0\,/\,100 & 0.3\,/\,0 & 0\,/\,100 & 0\,/\,0 & 7.9\,/\,0 & 0\,/\,100 & 0\,/\,100 & 3.1\,/\,0 & 0\,/\,0 & 99\,/\,0 & 1.1\,/\,0 \\
Kazakh\,$\rightarrow$\,en & 0\,/\,0 & 0\,/\,100 & 0.6\,/\,0 & 0\,/\,100 & 0\,/\,0.4 & 10\,/\,0 & 0\,/\,100 & 0\,/\,100 & 2.7\,/\,0 & 0\,/\,0 & 98\,/\,0 & 0.6\,/\,0 \\
Odia\,$\rightarrow$\,en & 0.5\,/\,0.2 & 0\,/\,100 & 39\,/\,0 & 0\,/\,100 & 0\,/\,1.9 & 69\,/\,0 & 0\,/\,100 & 0\,/\,100 & 4.7\,/\,0 & 0\,/\,0 & 99\,/\,0 & 3.2\,/\,3.2 \\
Polish\,$\rightarrow$\,en & 0\,/\,0 & 0\,/\,100 & 0.2\,/\,0 & 0\,/\,100 & 0\,/\,0 & 7.4\,/\,0 & 0\,/\,100 & 0\,/\,100 & 1.5\,/\,0 & 0\,/\,0 & 96\,/\,0 & 1.0\,/\,0 \\
Romanian\,$\rightarrow$\,en & 0\,/\,0 & 0\,/\,100 & 0\,/\,0 & 0\,/\,100 & 0\,/\,0 & 9.1\,/\,0 & 0\,/\,100 & 0\,/\,100 & 1.6\,/\,0 & 0\,/\,0 & 90\,/\,0 & 0\,/\,0 \\
Serbian\,$\rightarrow$\,en & 0\,/\,0 & 0\,/\,100 & 0\,/\,0 & 0\,/\,100 & 0\,/\,0 & 7.0\,/\,0 & 0\,/\,100 & 0\,/\,100 & 2.8\,/\,0 & 0\,/\,0 & 92\,/\,0 & 1.1\,/\,0 \\
Slovak\,$\rightarrow$\,en & 0\,/\,0 & 0\,/\,100 & 0.1\,/\,0 & 0\,/\,100 & 0\,/\,0 & 9.0\,/\,0 & 0\,/\,100 & 0\,/\,100 & 1.5\,/\,0 & 0\,/\,0 & 93\,/\,0 & 0.7\,/\,0 \\
Tatar\,$\rightarrow$\,en & 0\,/\,0 & 0\,/\,100 & 0.2\,/\,0 & 0\,/\,100 & 0\,/\,0.8 & 9.6\,/\,0 & 0\,/\,100 & 0\,/\,100 & 3.1\,/\,0 & 0\,/\,0 & 98\,/\,0 & 0.8\,/\,0 \\
Ukrainian\,$\rightarrow$\,en & 0\,/\,0 & 0\,/\,100 & 0.3\,/\,0 & 0\,/\,100 & 0\,/\,0 & 8.0\,/\,0 & 0\,/\,100 & 60\,/\,100 & 2.0\,/\,0 & 0.3\,/\,0 & 95\,/\,0 & 0.6\,/\,0 \\
Uzbek\,$\rightarrow$\,en & 0\,/\,0 & 0\,/\,100 & 0.4\,/\,0 & 0\,/\,100 & 0\,/\,0.1 & 9.4\,/\,0 & 0\,/\,100 & 0\,/\,100 & 2.2\,/\,0 & 76\,/\,76 & 97\,/\,0 & 0.9\,/\,0 \\
\midrule
\multicolumn{13}{l}{\textit{Out of English (en$\rightarrow X$)}}\\
en\,$\rightarrow$\,Amharic & 0.1\,/\,0 & 0\,/\,100 & 42\,/\,0 & 0\,/\,100 & 0.2\,/\,5.0 & 97\,/\,0 & 0\,/\,100 & 0.1\,/\,100 & 4.3\,/\,0 & 22\,/\,0 & 100\,/\,0 & 1.3\,/\,0 \\
en\,$\rightarrow$\,Catalan & 0\,/\,0 & 0\,/\,100 & 0.7\,/\,0 & 0\,/\,100 & 0\,/\,0.3 & 12\,/\,0 & 0\,/\,100 & 0\,/\,100 & 4.8\,/\,0 & 0.1\,/\,0 & 99\,/\,0 & 0\,/\,0 \\
en\,$\rightarrow$\,Chuvash & 0.1\,/\,0 & 0\,/\,100 & 5.0\,/\,0 & 0\,/\,100 & 0\,/\,6.8 & 78\,/\,0 & 0\,/\,100 & 0\,/\,100 & 3.4\,/\,0 & 40\,/\,0 & 100\,/\,0 & 0.7\,/\,0 \\
en\,$\rightarrow$\,Czech & 0.1\,/\,0 & 0\,/\,100 & 4.4\,/\,0 & 0\,/\,100 & 0.1\,/\,5.9 & 25\,/\,0 & 0\,/\,100 & 0\,/\,100 & 4.3\,/\,0 & 0\,/\,0 & 100\,/\,0 & 0.7\,/\,0 \\
en\,$\rightarrow$\,Dutch & 0\,/\,0 & 0\,/\,100 & 2.2\,/\,0 & 0\,/\,100 & 0\,/\,0.4 & 14\,/\,0 & 0\,/\,100 & 0\,/\,100 & 3.6\,/\,0 & 0\,/\,0 & 99\,/\,0 & 0.2\,/\,0 \\
en\,$\rightarrow$\,Greek & 0\,/\,0 & 0\,/\,100 & 29\,/\,0 & 0\,/\,100 & 0\,/\,9.0 & 46\,/\,0 & 0\,/\,100 & 0\,/\,100 & 4.3\,/\,0 & 0.4\,/\,0 & 100\,/\,0 & 0.3\,/\,0 \\
en\,$\rightarrow$\,Hebrew & 0\,/\,0 & 0\,/\,100 & 20\,/\,0 & 0\,/\,100 & 0.3\,/\,8.6 & 44\,/\,0 & 0\,/\,100 & 0\,/\,100 & 3.7\,/\,0 & 0.1\,/\,0 & 100\,/\,0 & 0.4\,/\,0 \\
en\,$\rightarrow$\,Hindi & 0\,/\,0 & 0\,/\,100 & 32\,/\,0 & 0\,/\,100 & 0.1\,/\,5.3 & 48\,/\,0 & 0\,/\,100 & 0\,/\,100 & 3.9\,/\,0 & 0\,/\,0 & 100\,/\,0 & 0.4\,/\,0 \\
en\,$\rightarrow$\,Kazakh & 0\,/\,0 & 0\,/\,100 & 7.9\,/\,0 & 0\,/\,100 & 0.2\,/\,7.2 & 33\,/\,0 & 0\,/\,100 & 0\,/\,100 & 3.6\,/\,0 & 0\,/\,0 & 100\,/\,0 & 0.2\,/\,0 \\
en\,$\rightarrow$\,Odia & 0.2\,/\,0 & 0\,/\,100 & 17\,/\,0 & 0\,/\,100 & 0\,/\,6.4 & 81\,/\,0 & 0\,/\,100 & 0\,/\,100 & 4.6\,/\,0 & 0.6\,/\,0 & 100\,/\,0 & 76\,/\,76 \\
en\,$\rightarrow$\,Polish & 0\,/\,0 & 0\,/\,100 & 5.5\,/\,0 & 0\,/\,100 & 0\,/\,2.8 & 21\,/\,0 & 0\,/\,100 & 0\,/\,100 & 3.6\,/\,0 & 0\,/\,0 & 100\,/\,0 & 0.2\,/\,0 \\
en\,$\rightarrow$\,Romanian & 0\,/\,0 & 0\,/\,100 & 1.2\,/\,0 & 0\,/\,100 & 0\,/\,0.6 & 14\,/\,0 & 0\,/\,100 & 0\,/\,100 & 4.0\,/\,0 & 0\,/\,0 & 99\,/\,0 & 0\,/\,0 \\
en\,$\rightarrow$\,Serbian & 0\,/\,0 & 0\,/\,100 & 4.5\,/\,0 & 0\,/\,100 & 0\,/\,6.9 & 36\,/\,0 & 0\,/\,100 & 0\,/\,100 & 5.6\,/\,0 & 0.1\,/\,0 & 100\,/\,0 & 0.8\,/\,0 \\
en\,$\rightarrow$\,Slovak & 0.2\,/\,0 & 0\,/\,100 & 3.6\,/\,0 & 0\,/\,100 & 0\,/\,7.5 & 24\,/\,0 & 0\,/\,100 & 0\,/\,100 & 4.6\,/\,0 & 0.3\,/\,0 & 100\,/\,0 & 0.1\,/\,0 \\
en\,$\rightarrow$\,Tatar & 0\,/\,0 & 0\,/\,100 & 12\,/\,0 & 0\,/\,100 & 0\,/\,6.4 & 48\,/\,0 & 0\,/\,100 & 0\,/\,100 & 4.4\,/\,0 & 0.1\,/\,0 & 100\,/\,0 & 0.3\,/\,0 \\
en\,$\rightarrow$\,Ukrainian & 0\,/\,0 & 0\,/\,100 & 1.4\,/\,0 & 0\,/\,100 & 0\,/\,1.7 & 12\,/\,0 & 0\,/\,100 & 59\,/\,100 & 3.6\,/\,0 & 0.1\,/\,0 & 100\,/\,0 & 64\,/\,64 \\
en\,$\rightarrow$\,Uzbek & 0.2\,/\,0 & 0\,/\,100 & 12\,/\,0 & 0\,/\,100 & 0\,/\,6.4 & 20\,/\,0 & 0\,/\,100 & 0\,/\,100 & 3.9\,/\,0 & 0.2\,/\,0 & 100\,/\,31 & 0.2\,/\,0 \\
\midrule
\multicolumn{13}{l}{\textit{German pivot}}\\
de\,$\rightarrow$\,L.\,Sorbian & 0\,/\,0 & 0\,/\,100 & 1.0\,/\,0 & 0\,/\,100 & 0\,/\,6.3 & 46\,/\,0 & 0\,/\,100 & 0\,/\,100 & 2.0\,/\,0 & 2.0\,/\,0 & 99\,/\,0 & 0\,/\,0 \\
de\,$\rightarrow$\,U.\,Sorbian & 0\,/\,0 & 0\,/\,100 & 1.0\,/\,0 & 0\,/\,100 & 1.0\,/\,8.0 & 61\,/\,0 & 0\,/\,100 & 0\,/\,100 & 4.3\,/\,0 & 1.7\,/\,0 & 100\,/\,0 & 0\,/\,0 \\
\bottomrule
\end{tabular}}
\caption{Empty-output rate (\%) per model and translation direction, reported as \emph{answer\,/\,thinking}: the left value is the share of empty answer/translation outputs, the right value the share of empty reasoning traces. Rows are translation directions grouped into $X\rightarrow$en, en$\rightarrow X$, and German-pivot directions; columns are models (\textsc{nt}\,=\,non-thinking variant, which by construction emits no reasoning, hence 100\% empty thinking). Lower is better.}
\label{tab:app-empty}
\end{table*}

\clearpage
\newpage

\section{Discovered Problems in PolyMath Dataset}
\label{app:polymath_problems}


While annotating PolyMath \citep{wang2025polymath} and assessing the
associated human reasoning traces, our annotators identified a number of
errors in the original release. We group them into (i)~incorrect reference
answers in the English source and (ii)~translation and localization errors in
the multilingual subsets. We report all instances below and will contribute the
corresponding corrections as a pull request to the original dataset.

\subsection{Incorrect Reference Answers}
\label{app:polymath-answers}

We found three items in the English split whose gold answers are incorrect.
The corrected values follow from the problem statements and were verified by
our annotators. These errors propagated in other languages of PolyMath benchmark.

\paragraph{\texttt{low-en-94}.}
\emph{Problem.} \textit{In a neighborhood, the number of rabbits pets is twelve less than the combined number of pet dogs and cats. If there are two cats for every dog, and the number of dogs is 60, how many pets in total are in the neighborhood?}\\
Original answer: $348$.\quad Corrected answer: $195$.

\paragraph{\texttt{high-en-122}.}
\emph{Problem.} \textit{Let $a$, $b$, and $c$ be positive real numbers satisfying $ab+bc+ca=abc$. Determine the minimum value of $$a^abc + b^bca + c^cab.$$}\\
Original answer: $729$.\quad Corrected answer: $3^{28}$.

\paragraph{\texttt{top-en-118}.}
\emph{Problem.} \textit{Consider the surface $S$ of a cube side length $s$. Let $P$ be one of the vertices of the cube, and $D\subset S$ the collection of points on $S$ that are at most $\sqrt{2} \cdot s$ away from $P$, where distance is measured along the surface. Divide the area of $D$ by the area of $S$, leaving the answer in its exact form.}\\
Original answer: $\dfrac{\pi+3\sqrt{3}-3}{6}$.\quad
Corrected answer: $\dfrac{2\pi-3+3\sqrt{3}}{12}$.

\subsection{Translation and Localization Errors}
\label{app:polymath-translation}

Beyond the answer errors, we observed recurring problems in the translated
subsets. We label each instance with one or more of the following error types:
\textbf{(U)}~untranslated source-language content;
\textbf{(L)}~\LaTeX{}/formatting error;
\textbf{(T)}~terminology error or inconsistency;
\textbf{(M)}~mistranslation, including factual distortion;
\textbf{(I)}~information loss relative to the source;
\textbf{(C)}~calque or unidiomatic phrasing;
\textbf{(S)}~symbol/notation placement; and
\textbf{(X)}~cross-lingual misalignment between subsets.

\paragraph{German (de).}
\begin{itemize}\setlength{\itemsep}{1pt}
  \item \texttt{high-de-79}~(U): the otherwise German statement still
        contains untranslated English text.
  \item \texttt{high-de-81}~(L): the predicate ``gut'' is typeset in math mode
        (\verb|$gut$|) instead of as emphasized text.
  \item \texttt{medium-de-12}~(T): inconsistent terminology within a single
        item---``Kisten'' vs.\ ``Boxen'' for the containers and ``Kugel'' vs.\
        ``Ball'' for the balls.
  \item \texttt{medium-de-39}~(L): missing commas in the conditional clause.
  \item \texttt{low-de-17}~(M): ``35 Wochen pro Woche'' (\emph{35 weeks per
        week}) is nonsensical; the source specifies 35 \emph{hours} per week,
        so the item is factually distorted.
  \item \texttt{low-de-26}~(C): unidiomatic ``3 Paar Shorts''; natural German
        would not use ``Paar'' here (cf.\ ``3 Shorts'').
  \item \texttt{low-de-110}~(M): the localized problem is incoherent in both
        the German and the Czech versions and does not yield a solvable task.
\end{itemize}

\paragraph{Russian (ru).}
\texttt{low-ru-87}~(I): relative to the English source, the final question
omits information, leaving the requested quantity ambiguous. The English item
asks for an \emph{annual} salary, and the gold answer ($9360$) is the annual
figure; the Russian rendering drops the word ``annual'' (\foreignlanguage{russian}{``годовая''}), so a
solver cannot tell whether a monthly or an annual amount is wanted.

\begin{quote}\small
\textbf{English (source).} A company pays each of its employees \$600 in a
month. The company has a policy of increasing the salaries of each of its
employees by 10\% of the initial salary every year for those who have stayed
in the company for five years. If Sylvie just clocked 5 years in the company
last December, what is her \underline{annual} salary after three more years of
service? \emph{(Gold answer: $9360$.)}

\smallskip
\textbf{Russian (as released).} \foreignlanguage{russian}{Компания платит каждому из своих сотрудников
600~\$ в месяц. В компании действует политика ежегодного повышения заработной
платы на 10~\% от начальной для каждого сотрудника, проработавшего в компании
пять лет. Если Сильви отработала в компании 5 лет в прошлом декабре, сколько
составит её заработная плата спустя ещё три года работы?}

\smallskip
\textbf{Explanation.} ``\ldots how much will her \underline{salary} be after another
three years of work?'' --- the qualifier ``annual'' present in the source is
absent, and the monthly rate (\$600) is also given, so the target unit is
under\-specified.
\end{quote}

\paragraph{Spanish (es).}
The Spanish subset shows several recurring issues:
\begin{itemize}\setlength{\itemsep}{1pt}
  \item (M)~Mistranslations, e.g.\ ``intersecta'' for ``interseca'', and
        ``secesi\'on'' (\emph{secession}) for ``sucesi\'on'' (\emph{sequence}).
  \item (T)~Incorrect terminology, e.g.\ ``casco convexo'' instead of
        ``envoltura convexa'' for \emph{convex hull}.
  \item (T)~Context-unaware term choices, e.g.\ ``suministros'' for
        \emph{supplies}, which is wrong in context.
  \item (C)~English calques, e.g.\ ``se est\'a poniendo demasiado pesada'' for
        \emph{is getting too heavy}, a poor choice here.
  \item (S)~Currency/unit symbols placed before rather than after the numeric
        amount.
  \item (M)~Nonsensical translations, e.g.\ \texttt{low-es-110}.
  \item (X)~\texttt{high-es-46}: the Spanish and Vietnamese entries are
        swapped in the release; the content occupying the Spanish slot is in
        fact Vietnamese, whose \LaTeX{} formatting has additionally been
        altered.
\end{itemize}
Finally, the Spanish subset predominantly uses Latin American Spanish rather than Peninsular Spanish.

\newpage
\clearpage

\section{LLMs Usage in This Work}
Finally, we clarify the role of LLMs in this work---where they were used and where they were \textit{not}. Neural machine translation systems were used only to generate initial translation drafts for the target languages. All subsequent quality assessment and refinement were performed exclusively through \textit{human} correction, without automated post-editing systems.

Claude was actively used to assist with code development, while ChatGPT was used for language editing and polishing.

All research ideas, experimental design, analyses, and findings were developed \textit{solely} by the authors.

\end{document}